\documentclass[runningheads]{llncs}

\usepackage[final]{eccv}

\usepackage{eccvabbrv}

\usepackage{graphicx}
\usepackage{booktabs}
\usepackage{makecell}
\usepackage{amsfonts}       %
\usepackage{nicefrac}       %
\usepackage{microtype}      %
\usepackage{lipsum}
\usepackage{fancyhdr}       %
\usepackage{bm}
\usepackage{amsmath}
\usepackage{float}
\usepackage{bbding}
\usepackage{comment}

\usepackage[accsupp]{axessibility}  %

\excludecomment{hidden}

\usepackage{hyperref}

\usepackage{orcidlink}

\usepackage{amssymb}
\usepackage{amsmath}
\usepackage{bbm}
\usepackage{float}
\usepackage{scalerel}
\usepackage{pgfplots}
\usepackage{ulem}
\usepackage{multirow}
\usepackage{adjustbox}
\usepackage{colortbl}
\usepackage{fancyhdr}
\usepackage{silence}
\usepackage{numprint}
\usepackage{soul}
\usepackage{algorithm}
\usepackage{algpseudocode}
\usepackage{caption}
\usepackage{subcaption}
\usepackage{graphicx}
\usepackage{tikz}
\usetikzlibrary{angles,quotes,3d,math,arrows.meta,calc,positioning,fit,backgrounds,decorations.pathreplacing,calligraphy,shapes,shapes.multipart}

\usepackage{arydshln}
\usepackage{tabularray}
\usepackage{diagbox}

\usepackage{letltxmacro}

\newcommand{\methodname}{LoRAdapter}
\newcommand{\style}{style}
\newcommand{\structure}{structure}

\newlength{\mycw}

\newcommand\rurl[1]{%
  \href{https://#1}{\nolinkurl{#1}}%
}

\definecolor{ourgreen}{RGB}{46, 204, 113}
\definecolor{ourgreenborder}{RGB}{39, 174, 96}
\definecolor{ourblue}{RGB}{52, 152, 219}
\definecolor{ourblueborder}{RGB}{41, 128, 185}
\definecolor{ourorange}{RGB}{230, 126, 34}
\definecolor{ourorangeborder}{RGB}{211, 84, 0}
\definecolor{ourred}{RGB}{231, 76, 60}
\definecolor{ourredborder}{RGB}{192, 57, 43}
\definecolor{ouryellow}{RGB}{241, 196, 15}
\definecolor{ouryellowborder}{RGB}{243, 156, 18}
\definecolor{ourpurple}{RGB}{155, 89, 182}
\definecolor{ourpurpleborder}{RGB}{142, 68, 173}
\definecolor{ourturquoise}{RGB}{26, 188, 156}
\definecolor{ourturquoiseborder}{RGB}{22, 160, 133}
\definecolor{ourturquoise}{RGB}{26, 188, 156}
\definecolor{ourturquoiseborder}{RGB}{22, 160, 133}
\definecolor{ourwhite}{RGB}{236, 240, 241}
\definecolor{ourwhiteborder}{RGB}{189, 195, 199}
\definecolor{ourgray}{RGB}{149, 165, 166}
\definecolor{ourgrayborder}{RGB}{127, 140, 141}

\tikzset{
    text centered,
    anchor=center,
}
\tikzset{
    simple node image/.style n args={0}{%
        rectangle,
        inner sep=0,
        text centered,
        anchor=center,
        align=center,
        node distance=0mm
    }
}

\makeatletter
\def\adl@drawiv#1#2#3{%
        \hskip.5\tabcolsep
        \xleaders#3{#2.5\@tempdimb #1{1}#2.5\@tempdimb}%
                #2\z@ plus1fil minus1fil\relax
        \hskip.5\tabcolsep}
\newcommand{\cdashlinelr}[1]{%
  \noalign{\vskip\aboverulesep
           \global\let\@dashdrawstore\adl@draw
           \global\let\adl@draw\adl@drawiv}
  \cdashline{#1}
  \noalign{\global\let\adl@draw\@dashdrawstore
           \vskip\belowrulesep}}
\makeatother

\begin{document}

\title{\fcolorbox{ourwhiteborder}{white}{\textbf{CTR}\underline{L}}\hspace{-1.2mm}\underline{\textit{or}}\hspace{-1.2mm}\fcolorbox{ourwhiteborder}{white}{\hspace{-5.8mm}\underline{\textit{or}\hspace{-.5mm}A}LT}\hspace{-1.2mm}er: Conditional LoRAdapter for Efficient 0-Shot Control \& Altering of T2I Models} 

\titlerunning{CTRLorALTer: Conditional LoRAdapter}

\author{Nick Stracke\inst{1} \and Stefan Andreas Baumann\inst{1} \and Joshua Susskind \inst{2} \and\\ Miguel Angel Bautista \inst{2} \and Björn Ommer\inst{1}}

\authorrunning{Stracke et al.}

\institute{CompVis @ LMU Munich, MCML\and Apple}

\maketitle

\begin{abstract}

Text-to-image generative models have become a prominent and powerful tool that excels at generating high-resolution realistic images. However, guiding the generative process of these models to take into account detailed forms of conditioning reflecting style and/or structure information remains an open problem. In this paper, we present \textbf{\methodname{}}, an approach that unifies both style and structure conditioning under the same formulation using a novel conditional LoRA block that enables zero-shot control. \methodname{} is an efficient and powerful approach to condition text-to-image diffusion models, which enables fine-grained control conditioning during generation and outperforms recent state-of-the-art approaches.

Project page and code: \rurl{compvis.github.io/LoRAdapter/}

  \keywords{T2I Models \and Diffusion Models \and Controllable Generation}
\end{abstract}

\begin{figure}
    \centering
    \includegraphics[width=0.85\textwidth]{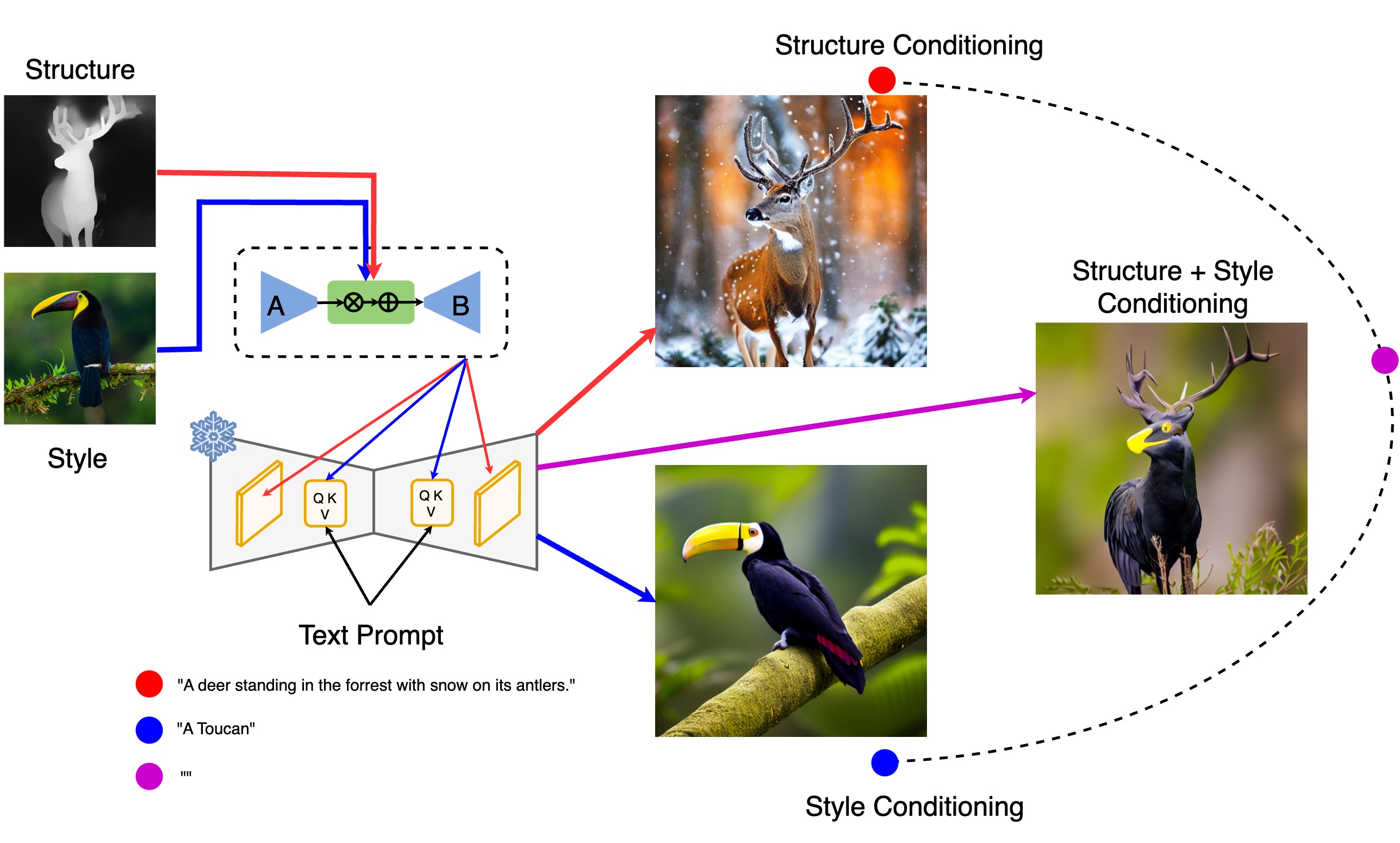}
    \caption{\textbf{\methodname{}} allows structure and style control of the image generation process of text-to-image models in a zero-shot manner. Our approach enables powerful fine-grained and efficient unified control over \textit{both structure and style conditioning} using conditional LoRA blocks. }
    \label{fig:teaser}
\end{figure}

\section{Introduction}

Text-to-image models have become a foundational component of Computer Vision, especially with the introduction of large-scale diffusion models like Stable Diffusion (SD)~\cite{rombach2022high}, DALL-E 2~\cite{ramesh2022hierarchical}, Imagen~\cite{saharia2022photorealistic}, RAPHAEL~\cite{xue2023raphael} and, eDiff-I~\cite{balaji2022ediffi}. These models have empowered users to create realistic images from textual prompts, although crafting effective prompts is a complex task in itself due to the nuanced prompt engineering required~\cite{witteveen2022investigating} to effectively condition the image generation process. To tackle this problem, prior approaches such as SD Image Variations \cite{sdimagevariations}, Stable unCLIP \cite{rombach2022high}, and Composer \cite{huang2023composer}, have attempted to perform conditioning using image prompts to enhance the expressivity of the conditioning mechanism giving users more flexibility during generation. In practice, this family of approaches requires either training from scratch or fully fine-tuning large text-conditioned model weights, incurring high computational costs and requiring large datasets.

To reduce training cost and time, recent approaches like ControlNet \cite{zhang2023adding}, T2I-Adapter \cite{mou2023t2i}, Uni-ControlNet \cite{zhao2023uni} or IP-Adapters \cite{ye2023ip-adapter} have focused on designing adapters that keep the base model frozen and only introduce a limit number of new layers. This not only reduces the number of trainable parameters but also allows training on much smaller datasets because the frozen base model cannot suffer from catastrophic forgetting. These adapters are typically handcrafted to a specific model architecture and conditioning modality. They can be roughly assigned into one of two groups:

\begin{itemize}

\item Adapters for pixel-level fine-grained conditioning over the generated image. This is often referred to as structural or local conditioning since it allows influencing the generated image on a per-pixel level by \eg providing a depth map and effectively setting the layout of the image \cite{zhang2023adding, zhao2023uni, qin2023unicontrol, jiang2023scedit, zavadski2023controlnetxs, mou2023t2i}

\item Adapters for style or global conditioning perform tasks such as image-to-image translation similar to unCLIP \cite{rombach2022high, huang2023composer, xu2022versatile, Li_2023_CVPR}. Here, the focus is not on fine-grained control but instead on generating an image similar to the prompt image, where similarity is defined by sharing the same style or semantics. 
 
\end{itemize}

Empirically, adapters always favor one type of conditioning over the other. For example, both Uni-ControlNet \cite{zhao2023uni} and T2I-Adapter \cite{mou2023t2i} have attempted \style{} conditioning but their performance is not comparable to leading \style{} approaches \cite{ye2023ip-adapter}. Conversely, \style{} adapters require ControlNet or T2I-Adapter for \structure{} conditioning. However, ControlNet is a large adapter that copies the entire encoder of Stable Diffusion's U-Net which significantly increases compute and thus inference time. Our goal is to find a unified approach that is capable of both \structure{} and \style{} conditioning, providing complete control over the generated image. Formulating such a unified approach for conditioning on global controls like \textit{style} and on local controls like \textit{structure}, in an \textit{efficient} and \textit{generic} manner remains a key open problem.

To address this challenging problem, we introduce \methodname{} (see Fig. \ref{fig:teaser}), a unified approach for incorporating images to condition both the style and structure of the generated images. \methodname{} takes inspiration from Low-Rank-Adaptations (LoRAs) \cite{hu2021lora}, which have shown remarkable performance for various tasks in diffusion models, such as distillation or learning specific concepts \cite{ruiz2023hyperdreambooth, luo2023lcm}. In particular, LoRAs low-rank property naturally regularizes the conditioning while offering unparalleled flexibility as any layer can be adapted in an architecture-agnostic way. Thus far, however, LoRAs have been implemented as fixed adaptations, being independent of the input. This means layer adaptation cannot change at test time for a given conditioning and, therefore, cannot be used for zero-shot generalization.

\methodname{} is a novel approach to adding conditional information to LoRAs, enabling zero-shot generalization and making them applicable for both \structure{} and \style{} and possibly many other conditioning types. \methodname{} is compact and efficient, \eg, optimizing 16M parameters vs the 22M of IP-Adapters \cite{ye2023ip-adapter} or 361M of ControlNet \cite{zhang2023adding}, while outperforming recent adapter approaches \cite{ye2023ip-adapter} and even approaches that train models from scratch (see Tab. \ref{tab:main_style}). \methodname{} trains LoRA \cite{hu2021lora} modules for different blocks across the base network and is efficient both during training and inference. Our contributions are summarized as follows:

\begin{itemize}
    \item 
    We propose \methodname{}, a generic approach to train conditional LoRAs that is agnostic to model architecture and conditioning modality.
    \item
    We implement \methodname{} for Stable Diffusion, offering a unified conditioning mechanism for both style and structure, enabling zero-shot conditioning.
    \item 
    We show the effectiveness of this approach, outperforming dedicated adapters for either style and structure on various metrics.
    
\end{itemize}

\section{Related Work}

We now review existing adapter approaches for \style{} and \structure{} and discuss the current role of LoRAs in diffusion models. 

\subsection{Structure Adapters}
Structure adapters are specifically designed to operate on various types of local conditioning modalities such as Depth, HED \cite{hed}, Canny Edges \cite{canny}, Scribbles, or key poses \cite{cao2017realtime}. The goal of these adapters is to spatially align the generated image with the provided condition. One of the most prominent approaches to tackle this problem is ControlNet \cite{zhang2023adding} for Stable Diffusion. ControlNet creates a copy of the U-Net's encoder and combines the skip connections additively with the original skip connections, which then get fed into the original decoder. While obtaining great performance, ControlNet adds a lot of computational overhead as the forward pass includes a second encoder which can increase inference speed by up to 50\%. Another consequence of the large capacity is that ControlNet tends to interpret the \structure{} map directly. This can be desirable when sampling images without other conditioning via the text prompt or other adapters, but can also lead to entanglement of structure and style.

Following a similar idea, T2I-Adapter \cite{mou2023t2i} also modifies skip connections while using a much smaller encoder network. To extend conditioning to multiple modalities, Uni-ControlNet \cite{zhao2023uni} was proposed as an alternative solution to having to train multiple separate and modality-specific ControlNets. Uni-ControlNet trains a single model for multiple conditions by concatenating the structure maps along the channel dimension and using the concatenated tensor as input. Uni-ControlNet uses SPADE \cite{park2019semantic} to combine skip-connections instead of the simple addition used by ControlNet. SCEdit \cite{jiang2023scedit} proposed a similar but more efficient solution for conditioning on multiple modalities by integrating a small tuner network (SC-Tuner) between skip-connections which removes the need for a separate encoder network akin to ControlNet.

\subsection{Style Adapters}
Style adapters are used as an alternative to unCLIP \cite{rombach2022high} like models, which are directly trained to invert CLIP image embeddings. The main advantage of using a \style{} adapter as opposed to an unCLIP model is that the original text conditioning is preserved, enabling additional control over the generated images. Additionally, all recently released large open-source diffusion models only offered text conditioning, making an adapter approach necessary \cite{podell2023sdxl, pernias2024wrstchen}.

IP-Adapter \cite{ye2023ip-adapter} adds a decoupled cross-attention layer that operates on four image tokens instead of text tokens. The resulting activations of this decoupled cross-attention layer are scaled and added to the original cross-attention activations. This results in a lightweight adapter that integrates well with potential text conditioning and generates images faithful to the image prompt.
In contrast, SeeCoder \cite{xu2023prompt} trains an encoder with 2D spatial embedding to convert image tokens to text tokens and completely replaces the standard text tokens with their tokens.

Structure adapters have also attempted to include style conditioning. Uni-ControlNet \cite{zhao2023uni} includes a separate mapping network to convert a pooled CLIP image token to four CLIP text tokens, which they concatenate to the original text embeddings. ControlNet-shuffle \cite{zhang2023adding} shuffles an RGB image and uses that as conditioning similar to its \structure{} conditioning. However, the performance of these approaches does not compare favorably to dedicated style adapters.

\subsection{LoRAs in Diffusion Models}
While LoRAs were initially introduced as an efficient alternative to fine-tuning large language models \cite{hu2021lora}, they were quickly adopted for diffusion models \cite{ryudiffusionlora} as a relatively efficient and cheap option to adjust their generation capabilities. Combining LoRAs with methods such as Dreambooth \cite{ruiz2023dreambooth} performs on par with a full fine-tuning approach while only introducing a fraction of trainable parameters. LoRAs are commonly trained on a small set of images with the goal of capturing a specific style instead of a subject. ZipLoRA \cite{shah2023ziplora} proposes a merging algorithm to combine multiple LoRAs gracefully by minimizing the cosine similarity of LoRAs that adapt the same weight matrix. LyCORIS \cite{lycoris} trained an entire library of specific LoRAs in diffusion models with various benefits. \cite{luo2023lcm} showed that LoRAs are even powerful enough for more complex objectives such as consistency distillation. Finally, Concept Sliders \cite{gandikota2023concept} use LoRAs to train disentangled edit directions to edit images. They scale the LoRA in the negative and positive directions to either increase or decrease a given attribute.

\section{Preliminaries}

\subsection{Diffusion}
Diffusion models represent a category of generative models characterized by a two-stage process: a forward diffusion that incrementally introduces Gaussian noise over a series of $T$ steps, and a reverse denoising process that reconstructs samples from this noise. In text-to-image models, diffusion can be directed by supplementary inputs like text descriptions. The primary training objective, $\boldsymbol{\epsilon}_{\theta}$, aimed at noise prediction, simplifies the variational lower bound as follows:
\begin{equation}
L_{\text{simple}}=\mathbb{E}_{\boldsymbol{x}_{0},\boldsymbol{\epsilon}\sim \mathcal{N}(\mathbf{0}, \mathbf{I}), \boldsymbol{c}, t} \| \boldsymbol{\epsilon}- \boldsymbol{\epsilon}_\theta\big(\underbrace{\sqrt{\bar{\alpha}_t} \boldsymbol{x}_0 + \sqrt{1 - \bar{\alpha}_t}\boldsymbol{\epsilon}}_{\boldsymbol{x}_t}, \boldsymbol{c}, t\big)\|^2,
\vspace{-1mm}
\end{equation}
where $\boldsymbol{x}_{0}$ is the original data influenced by condition $\boldsymbol{c}$, $t$ is the diffusion timestep, and $\boldsymbol{x}_t$ represents the data at step $t$, with $\bar{\alpha}_t$ dictating the noise level. 

During sampling, an initial pure Gaussian latent $\boldsymbol{x}_T$ is iteratively denoised by the diffusion model to obtain a sample $\boldsymbol{x}_0$ from the modeled (conditional) distribution.
Conditional models often employ classifier guidance to fine-tune the balance between accuracy and diversity, with classifier-free guidance \cite{ho2022classifier} serving as a preferred alternative that merges conditional and unconditional noise predictions during training and sampling:
\begin{equation}
\hat{\boldsymbol{\epsilon}}_{\theta}(\boldsymbol{x}_t, \boldsymbol{c}, t) = w\boldsymbol{\epsilon}_{\theta}(\boldsymbol{x}_t, \boldsymbol{c}, t)+(1-w)\boldsymbol{\epsilon}_{\theta}(\boldsymbol{x}_t, t),
\end{equation}
where $w$ modulates the adherence to condition $\boldsymbol{c}$. This technique is pivotal for text-to-image models, enhancing the alignment between generated images and text prompts.

\subsection{Low-Rank Adaptation (LoRA)}
Low-Rank Adaptation (LoRA) \cite{hu2021lora} is a method for finetuning pre-trained foundation models by keeping the original weight matrices frozen and instead adding a new set of low-rank weight matrices. While this method was originally proposed in the context of large language models, it has been recently adopted in the domain of diffusion models \cite{luo2023lcm, gandikota2023concept, cheng2024resadapter}. 

Specifically, the original weight matrix of a layer in the base model $W_0 \in \mathbb{R}^{d \times k} $  is kept frozen and is adapted by a low-rank matrix $W + \Delta W = W + BA$ with $B \in \mathbb{R}^{d \times r}$ and  $A \in \mathbb{R}^{r \times k}$. Crucially, $r$ is chosen following $r \ll \min(d, k)$. The forward pass for the adapted layer is then:
\begin{equation}\label{eq:lora}
    h = W_0 x + \Delta Wx = W_0 x + B A x.
\end{equation}

Typically, $B$ is zero-initialized, so the LoRA does not affect the model initially. This yields a generic approach, as we can freely choose the rank $r$ and the weights according to the given task. This happens on a per-layer basis, \ie, if we adapt $n$ layers, we will have $n$ $A$ and $B$ matrices.

\section{Method}

There are two kinds of approaches for conditioning text-to-image diffusion models: style or global conditioning \cite{ye2023ip-adapter, xu2023prompt}, and structure or local conditioning \cite{zhao2023uni, qin2023unicontrol, mou2023t2i, jiang2023scedit}. These approaches are specifically tailored to a specific conditioning objective and therefore missing a holistic approach that can efficiently learn both \textit{style and structure}. We now describe \methodname{}, an efficient, holistic, and customizable method for introducing both \style{} and \structure{} conditioning into text-to-image diffusion models.

\begin{figure}
    \centering
    \includegraphics[width=0.5\textwidth]{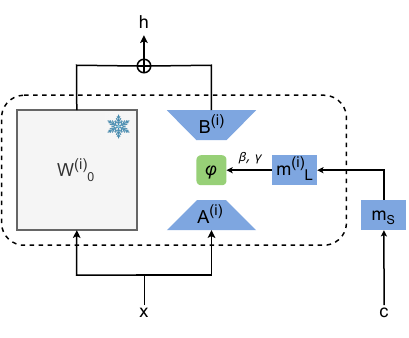}
    \caption{Overview of the proposed conditional LoRA block. The original weight matrix $W^{(i)}_0$ is frozen while all other layers are trained. $\phi$ is an affine transformation that operates on the low-dimensional embedding $A^{(i)}x$ and introduces the conditioning. The local mapper network $m_L^{(i)}$ predicts the scale and shift parameters $\beta, \gamma$ for the affine transformation. Typically, we set $m_L^{(i)}$ to be a small network. If complex transformations are required to map the conditioning $c$, this happens in $m_S$ since it is shared across all adapted layers.}
    \label{fig:general-arch}
\end{figure}

\subsection{Conditional LoRAs}
LoRAs have successfully enabled adaptation of diffusion models across a range of architectures \cite{luo2023lcm, ruiz2023hyperdreambooth, cheng2024resadapter} to steer their generation process. In this setting, LoRAs have been considered as free parameters and optimized on a training set of images exhibiting a specific characteristic.

Instead of learning a single LoRA to apply one specific modification, \eg, changing the style of the generated image to a painting or learning to represent one specific subject \cite{ruiz2023hyperdreambooth, ruiz2023dreambooth}, we propose a LoRA-based conditioning mechanism whose behavior changes based on conditioning provided at inference time, enabling zero-shot generalization. While previous methods depend on specific aspects of the architectures of the models they are applied to \cite{zhang2023adding, ye2023ip-adapter}, using LoRAs for conditioning enables us to efficiently adapt foundation models for conditional tasks in an architecture-agnostic way. Our LoRA-based conditioning mechanism is shown in Fig. \ref{fig:general-arch}.

Starting from the standard LoRA setup as in \cref{eq:lora}, we propose implementing conditioning of the LoRA by applying a transformation $\phi(Ax|m(c))$ to the low-dimensional intermediate embedding in the LoRA that introduces conditional behavior based on the conditioning $c$:
\begin{equation}\label{eq:cond_lora}
    h = W_0 x + \Delta Wx = W_0 x + B \phi(Ax|m(c)).
\end{equation}
This limits the conditioning to apply only to a low-rank subspace of the original vector space the weight matrix operates in. This is crucial for the generalization capabilities of LoRAs, as it introduces regularization and thus limits the adaptation to only focus on relevant aspects instead of spurious correlations from the dataset it was trained on, which results in more efficient training. By introducing our conditional adaptation of the model's behavior in such a low-rank space, our method can also benefit from those advantages.

In our framework for conditional LoRAs, we can theoretically choose $\phi(Ax|m(c))$ to be arbitrary, non-linear transformations, but we find that a simple affine transformation is sufficiently expressive for a wide range of applications:
\begin{equation}
    \phi(Ax|m(c)) = \gamma_\phi(m(c)) \odot Ax + \beta_\phi(m(c)),
\end{equation}
with $\odot$ denoting an elementwise (Hadamard) product and $\gamma$ and $\beta$ referring to the scale and shift factors.

To predict the $\gamma$ and $\beta$ from the conditioning $c$, we utilize a mapping network. This is generally a small neural network that we separate into a shared part $m_S(\cdot)$ and a layer-specific part $m_L^{(i)}(\cdot)$ that is individually learned for each adapted layer. For both parts, we generally use very small neural networks, which is in stark contrast to other works such as ControlNet \cite{zhang2023adding} that uses a full copy of the U-Net's encoder (see Tab. \ref{tab:main_style}). For brevity, we will always refer to both parts simply as $m(\cdot)$.

This general formulation directly enables both local and global conditioning, as it can be applied to the convolutional and attention layers in a standard diffusion model. For local conditioning, the mapping network can supply spatially dependent LoRA modulations. At the same time, it can also be used to supply global conditioning by introducing non-spatially dependent information, such as by adapting the cross-attention projections. Concurrent works \cite{kopiczko2024vera,liu2024dora,zhang2023adalora} also introduced improvements to the basic LoRA architecture. As \methodname{} does not depend on a specific implementation of LoRAs but just on the low-rank intermediate sub-space, these improvements can be directly transferred to our model.

\subsection{A Unified Conditioning Approach}

In this section, we provide implementation details for applying our conditional LoRAs to the layer types commonly used in large-scale diffusion models.

\begin{figure}[htbp]
\centering

\begin{subfigure}[b]{0.45\textwidth}
  \centering
  \includegraphics[width=\textwidth]{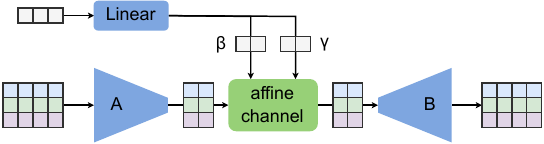} %
  \vspace{0mm} %
  \caption{In attention layers, we apply an affine transformation on the $K$ and $V$ projection matrix per channel for every token.}
  \label{fig:sub1}
\end{subfigure}
\hfill
\begin{subfigure}[b]{0.45\textwidth}
  \centering
  \includegraphics[width=\textwidth]{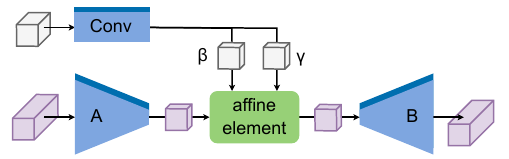} %
  \caption{To add structural conditioning to convolutional layers, we perform an affine transformation for every element similar to \cite{park2019semantic}.}
  \label{fig:sub2}
\end{subfigure}

\caption{Visualization of implementations of our conditional LoRAs for specific layers. }
\label{fig:test}
\end{figure}

\subsubsection{Attention Layers}
Many diffusion models, such as Stable Diffusion $\{1,2,\text{XL}\}$ \cite{rombach2022high, podell2023sdxl} or Imagen \cite{saharia2022photorealistic} include self- and cross-attention layers that operate on image (and text) tokens, with transformer-based diffusion models such as \cite{peebles2022dit} even purely relying on them for spatial computations. The output $h'$ of the attention layers is defined as 
\begin{equation}
    h' = \text{Attention}(Q,K,V) = \text{Softmax}\left(\frac{QK^T}{\sqrt{d}}\right)V
\end{equation}
with the projections $Q=hW_q$, $K=hW_k$ or $K=c_t W_k$, and $V=hW_v$ or $V=c_t W_v$, depending on whether it is self- or cross-attention, and the text tokens $c_t$.
Here, we apply our conditional LoRA on the linear projections $W_k$ and $W_v$ based on Equation \ref{eq:cond_lora}. We show a diagram of conditional LoRAs for attention layers in Fig. \ref{fig:sub1}.  

\subsubsection{Convolutional Layers}

For convolutional layers, which form the main backbone of many diffusion models \cite{rombach2022high, pernias2024wrstchen}, we adapt the convolutional layer
\begin{equation}
    h = K_0 \star x,
\end{equation}
with the 2D convolution kernel $K$ having input channel count ${ch}_{i}$ and output channel count ${ch}_o$,
in a similar manner as linear projections. However, in this setting we work on convolution kernels instead of matrices: we perform one convolution $K_A$ with the same kernel size, stride, and padding as the original convolution $K_0$, which reduces the channel count from ${ch}_i$ to the rank $r$ of the LoRA while keeping all else equal. Then, we apply our LoRA conditioning $\phi(\cdot|m(c))$ in this bottlenecked space and expand to the output channel count ${ch}_o$ using a pointwise convolution $K_B$:
\begin{equation}
    h = K_0 \star x + K_B \star \phi(K_A \star x|m(c)).
\end{equation}
A visual diagram of conditional LoRAs for convolutional layers is shown in Fig. \ref{fig:sub2}.

\section{Experiments}

\subsection{Experimental Setup}
\label{sect:results}

\begin{table}[tb]
    \centering
    \footnotesize
    \caption{Quantitative comparison for style conditioning of our proposed \methodname{} with other methods on the COCO validation set with four samples for every image. The best results are in \textbf{bold} (adapted from \cite{ye2023ip-adapter}).}
    \resizebox{\linewidth}{!}{
    \begin{tabular}{ccccccc}
\toprule
  Style Method & \makecell[c]{Reusable to \\custom models} & \makecell[c]{Native structure\\control} & \makecell[c]{Multimodal \\prompts} & \makecell[c]{Trainable \\parameters} & CLIP-T $\uparrow$ & CLIP-I $\uparrow$\\
\midrule
\emph{Training from scratch} \\
\midrule

Open unCLIP & \XSolidBrush & \XSolidBrush & \XSolidBrush & 893M & \textbf{0.608} &\textbf{0.858} \\

Kandinsky-2-1 & \XSolidBrush & \XSolidBrush & \XSolidBrush & 1229M &0.599 &0.855 \\

Versatile Diffusion & \XSolidBrush & \XSolidBrush & \Checkmark & \textbf{860M}& 0.587 & 0.830\\
\midrule
\emph{ Fine-tunining from text-to-image model } \\
\midrule
SD Image Variations &\XSolidBrush & \XSolidBrush & \XSolidBrush &\textbf{860M} &0.548 &0.760 \\
SD unCLIP &\XSolidBrush & \XSolidBrush & \XSolidBrush & 870M & \textbf{0.584} & \textbf{0.810} \\
\midrule
\emph{Adapters} \\
\midrule

Uni-ControlNet (Global Control) & \Checkmark & \Checkmark & \Checkmark &47M & 0.506 & 0.736 \\

T2I-Adapter (Style) & \Checkmark & \Checkmark & \Checkmark & 39M & 0.485 & 0.648 \\

ControlNet Shuffle & \Checkmark & \Checkmark & \Checkmark & 361M & 0.421 & 0.616 \\

IP-Adapter & \Checkmark & \XSolidBrush & \Checkmark & 22M & 0.588 & 0.828 \\
\textbf{\methodname{} (ours)} & \Checkmark & \Checkmark & \Checkmark & \textbf{16M} & \textbf{0.637} & \textbf{0.831}  \\
\textbf{\methodname{} SDXL (ours)} & \Checkmark & \Checkmark & \Checkmark & \textbf{103M} & \textbf{0.649} & \textbf{0.849}  \\
\bottomrule
\end{tabular}}
\label{tab:main_style}
\end{table}

\subsubsection{Data}
We train \methodname{} on a 40 million samples subset of COYO-700M \cite{kakaobrain2022coyo-700m} that only contains images with a short side of at least 512 pixels. Larger images are downscaled and center-cropped to 512 pixels. For text-conditioning, we use the standard prompts provided by the COYO-700M dataset.

\subsubsection{Training and Implementation Details}
All experiments are based on Stable Diffusion 1.5 \cite{rombach2022high} unless noted otherwise and adapted with a conditional LoRA according to the given task.  We train our adapter for 50,000 steps on 32 A100 GPUs with a global batch size of 256 and a learning rate of 0.0001. We use AdamW \cite{adamw} as an optimizer and drop each conditioning with a probability of 0.05. This includes the original text conditioning. 

\paragraph{For style conditioning,} we experiment with both CLIP ViT-L/14 \cite{radford2021learning} by OpenAI and CLIP ViT-H/14 by OpenClip \cite{ilharco2021openclip} as the image encoder but found that the ViT-H/14 tends to produce more visually appealing results. We still use ViT-L/14 for some ablation experiments as it results in a smaller model. Similar to \cite{ye2023ip-adapter}, we use a single linear layer followed by layer normalization for the shared mapping network $m_S$ as it outperformed much larger mapping networks and showed faster convergence. This maps the pooled CLIP image token $c_{img}$, keeping its original dimensionality $d_{img}$. The local mapping network $m^{(i)}_l$  consists of two separate linear layers that predict $\beta$ and $\gamma$ in the correct dimensionality $r$ of the LoRA embedding. In total, the shared mapping network only has 1M parameters.

\paragraph{For structure conditioning,} the shared mapping network is a series of convolutional layers that spatially align the dimensions of the conditioning images with the dimensions that Stable Diffusion's U-Net expects which sums up to only 1.3M parameters. As the U-Net includes several down- and upsampling blocks, the mapping network outputs feature maps at various resolutions that match the specific U-Net blocks. Each individual LoRA only gets the single feature map that matches the spatial dimensionality of the U-Net block that it operates in. This feature map is then further processed with a single convolutional layer as a local mapping to align the number of channels with the rank $r$ of the LoRA embedding. 
 
\subsubsection{Inference}
We sample with 50 DDIM \cite{ddim} steps and apply classifier-free guidance (CFG) \cite{cfg} with a scale of 7.5 to the prompt and style conditioning. We do not apply any CFG for structure conditioning. As with all LoRAs, we can also influence the strength of our \methodname{} by adjusting the LoRA scale $\lambda$, i.e. $h = W_0 + \lambda BAx$. We set $\lambda = 1$ unless otherwise noted.

\subsection{Quantitative Evaluation}

\begin{table}[tbp]
    \centering
    \caption{Comparison of a smaller and a larger version of our models vs. ControlNet.}
    \begin{tabular}{l c c c c}
        \toprule
        Model & Params & MSE-d $\downarrow$ & FID $\downarrow$ & LPIPS $\downarrow$ \\

        \midrule
        
        ControlNet  & 361M & 17.98 & 17.644 & 0.622  \\
        Uni-ControlNet & 361M &17.56 & \underline{16.453} & 0.601 \\ 
        T2I-Adapter & {\color{white}0}39M & 24.78 & 17.941 & 0.636 \\

        \methodname{}-A (Ours) & {\color{white}0}\textbf{17M}  & \underline{17.27} & 16.849  & \underline{0.599}\\

        \methodname{}-B (Ours) & {\color{white}0}\underline{32M}  & \textbf{15.34} & \textbf{15.670}  & \textbf{0.572}\\

         \bottomrule
    \end{tabular}
    \label{tab:main_struct}
\end{table}

To have a fair comparison with previous approaches, we follow previous literature on adapters \cite{ye2023ip-adapter, qin2023unicontrol, mou2023t2i} and evaluate our model on the validation set of COCO2017 \cite{coco} which contains 5,000 images. As every image comes with multiple prompts, we randomly select one prompt and use that for sampling.

\subsubsection{Style}
Following \cite{ye2023ip-adapter}, we sample four images for a given image and text prompt pair. To measure the effectiveness of the new style conditioning, we use CLIP ViT-L/14 to obtain the image embeddings of the generated and the prompt image and compute the cosine similarity between them (CLIP-I). In addition, we quantify how well the generated image still follows the text prompt by calculating the CLIP score \cite{hessel2021clipscore} between the text prompt and the generated image (CLIP-T). In Tab. \ref{tab:main_style} we show results across different approaches that either train models from scratch or fine-tune existing models, as well as compare with recent state-of-the-art adapters.

\setlength{\mycw}{0.07\textwidth}

\begin{table}[htb]
\centering
\begin{tabular}{p{\mycw} p{\mycw} p{\mycw} p{\mycw} p{\mycw} p{\mycw} p{\mycw} p{\mycw} p{\mycw} p{\mycw} p{\mycw} p{\mycw} }

    \multicolumn{12}{c}{Image Conditioning} \\
    \includegraphics[width=\linewidth]{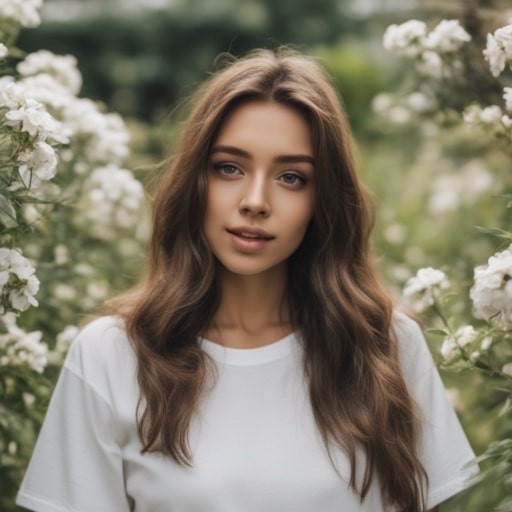} &
    \includegraphics[width=\linewidth]{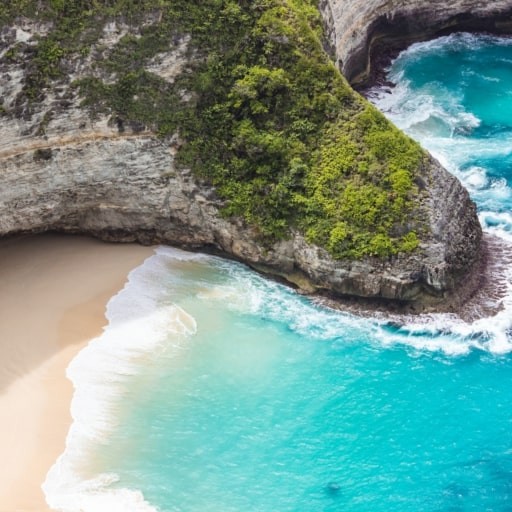} &
    \includegraphics[width=\linewidth]{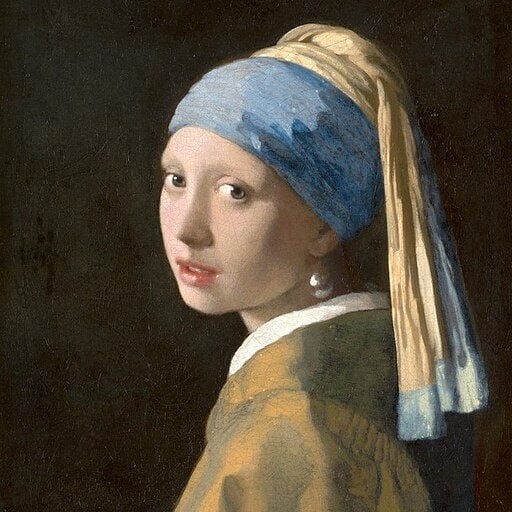} &
    \includegraphics[width=\linewidth]{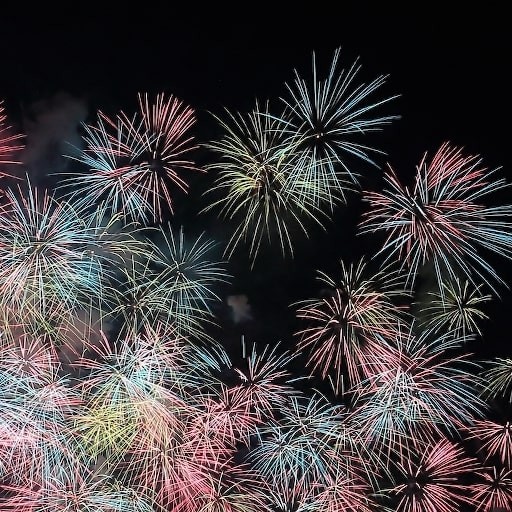} &
    \includegraphics[width=\linewidth]{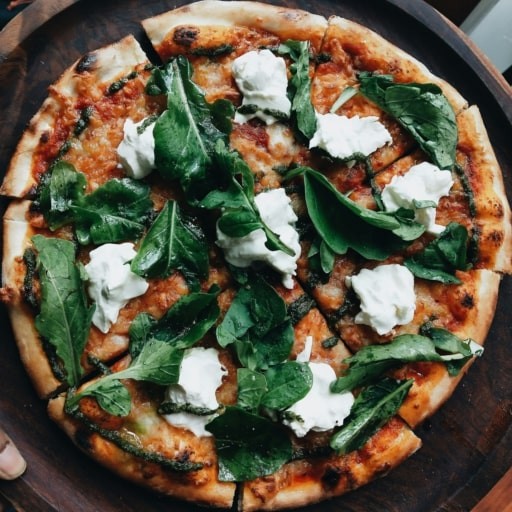} &
    \includegraphics[width=\linewidth]{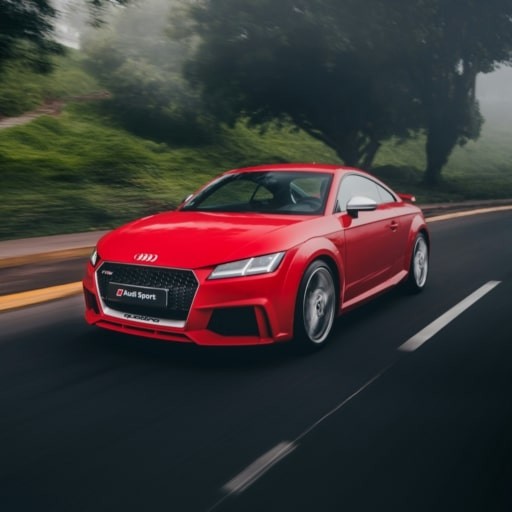} &
    \includegraphics[width=\linewidth]{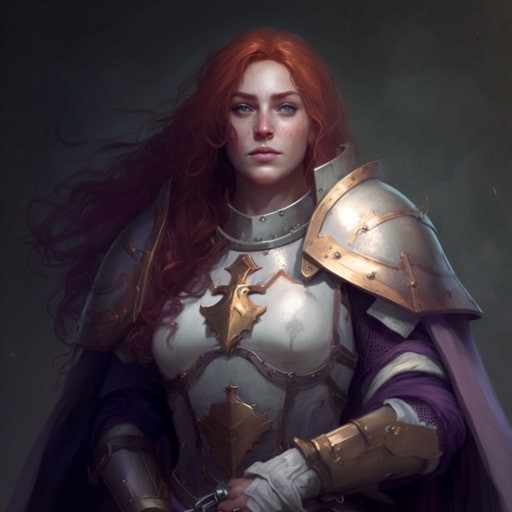} &
    \includegraphics[width=\linewidth]{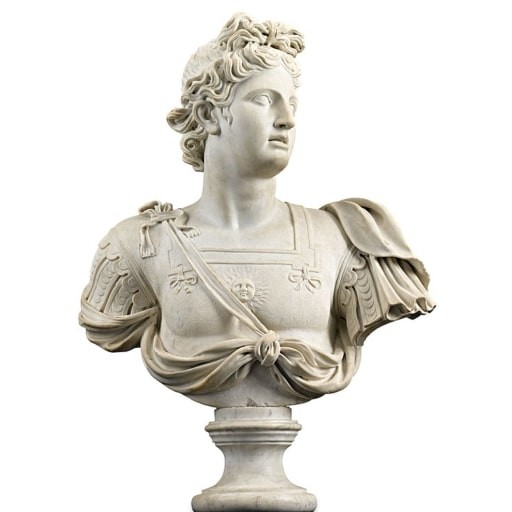} &
    \includegraphics[width=\linewidth]{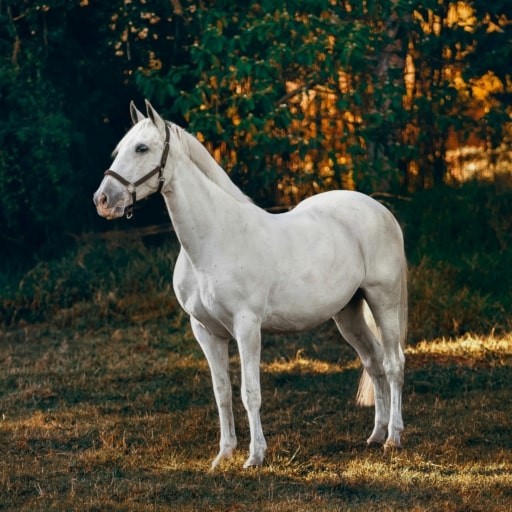} &
    \includegraphics[width=\linewidth]{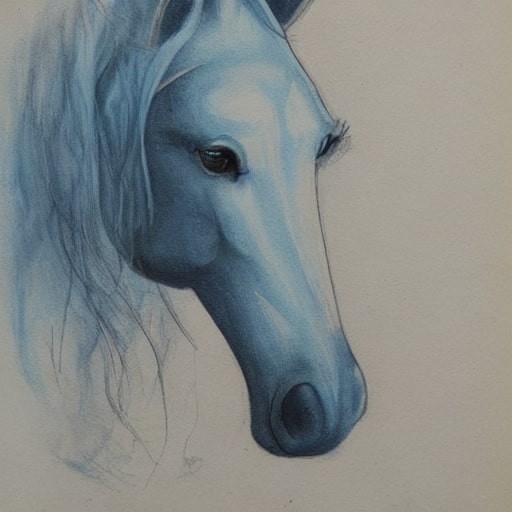} &
    \includegraphics[width=\linewidth]{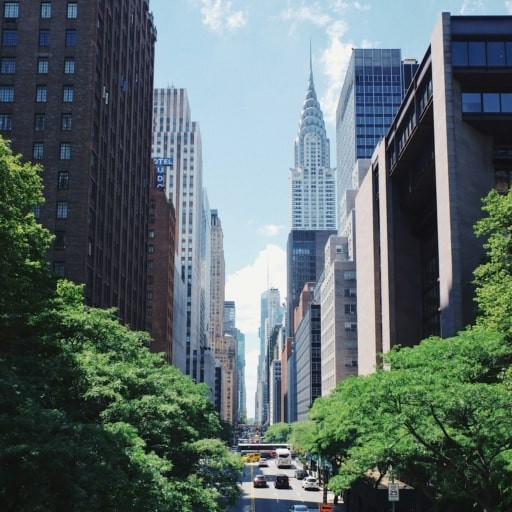} &
    \includegraphics[width=\linewidth]{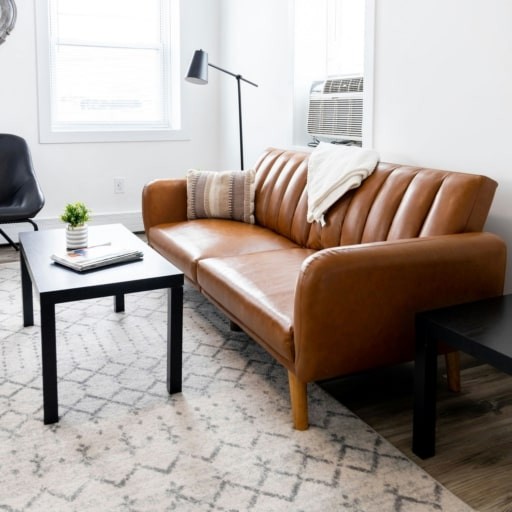} \\

    \multicolumn{12}{c}{ControlNet Shuffle} \\
    
    \includegraphics[width=\linewidth]{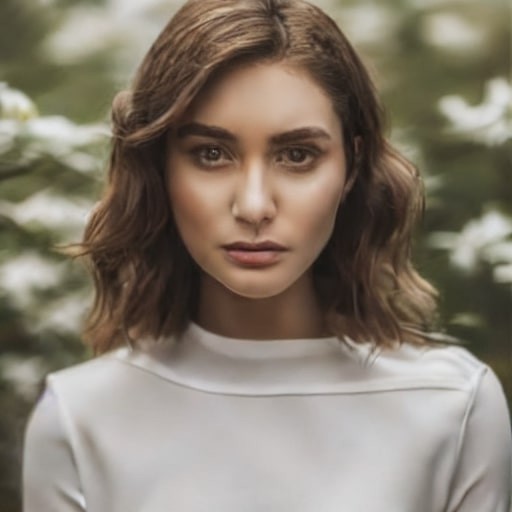}  &
    \includegraphics[width=\linewidth]{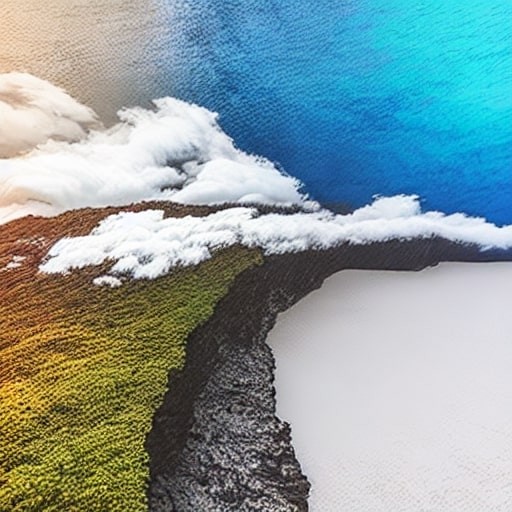}  &
    \includegraphics[width=\linewidth]{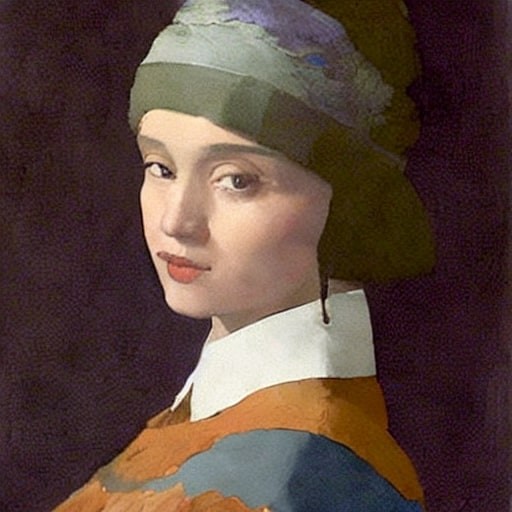}  &
    \includegraphics[width=\linewidth]{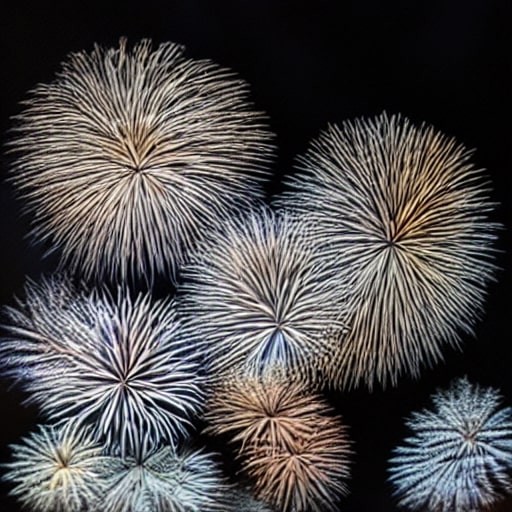}  &
    \includegraphics[width=\linewidth]{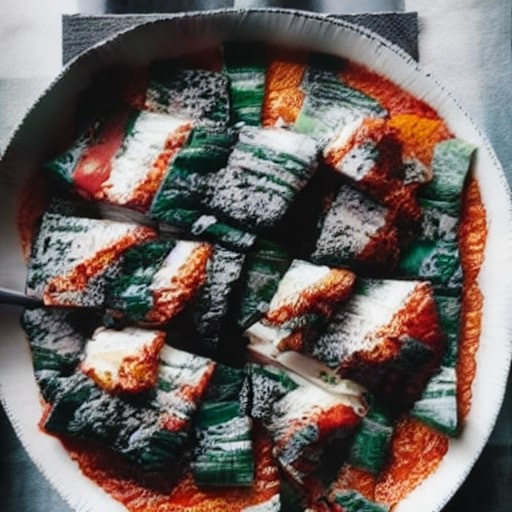}  &
    \includegraphics[width=\linewidth]{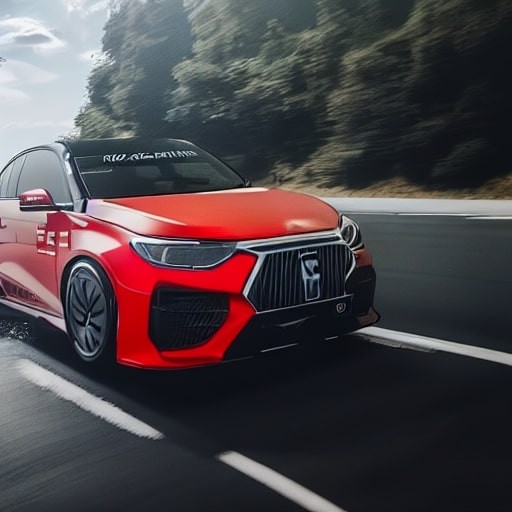}  &
    \includegraphics[width=\linewidth]{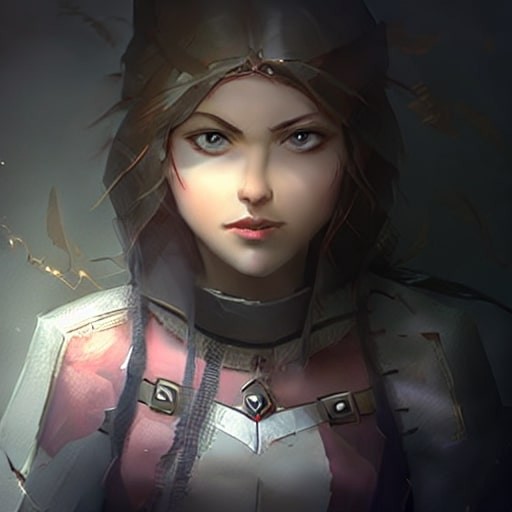}  &
    \includegraphics[width=\linewidth]{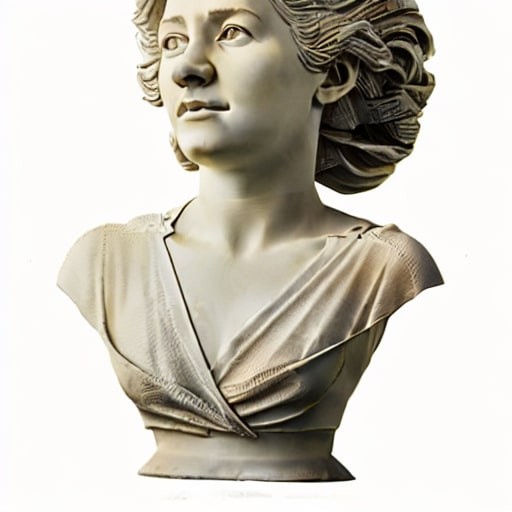}  &
    \includegraphics[width=\linewidth]{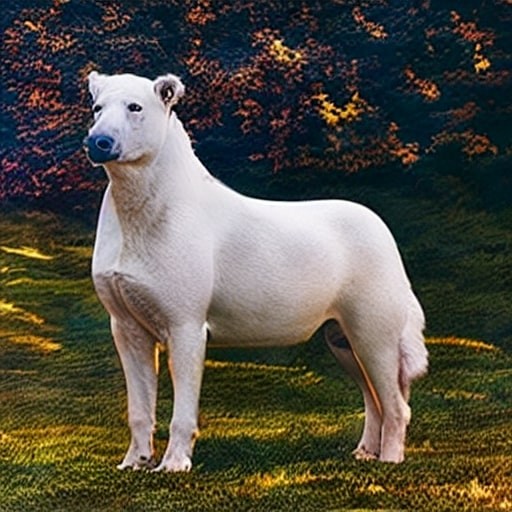}  &
    \includegraphics[width=\linewidth]{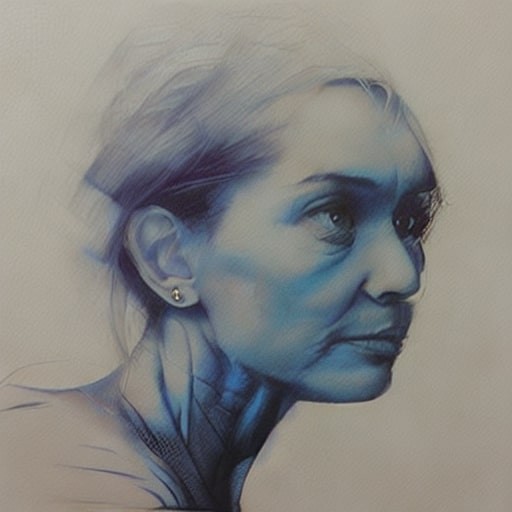}  &
    \includegraphics[width=\linewidth]{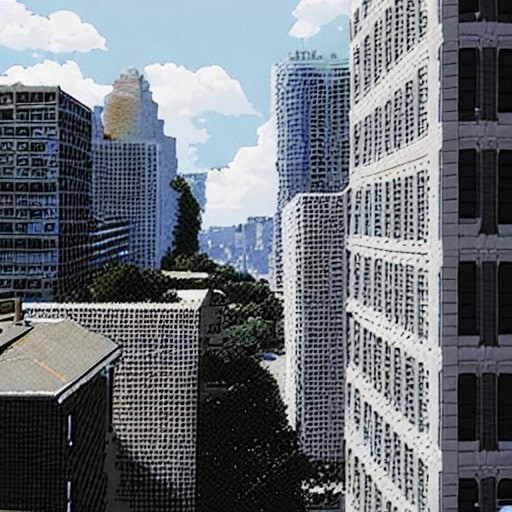}  &
    \includegraphics[width=\linewidth]{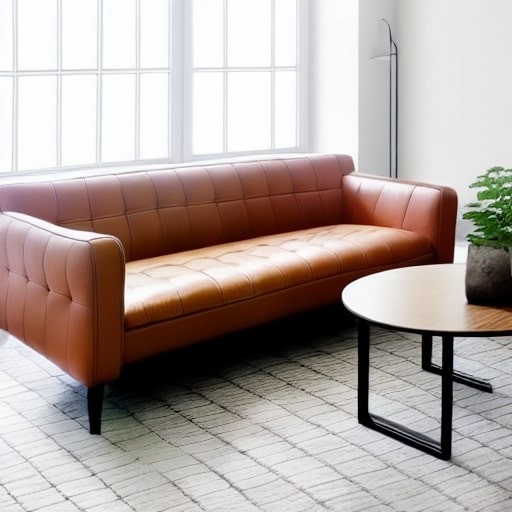}   \\

    \multicolumn{12}{c}{IP Adapter} \\
    
    \includegraphics[width=\linewidth]{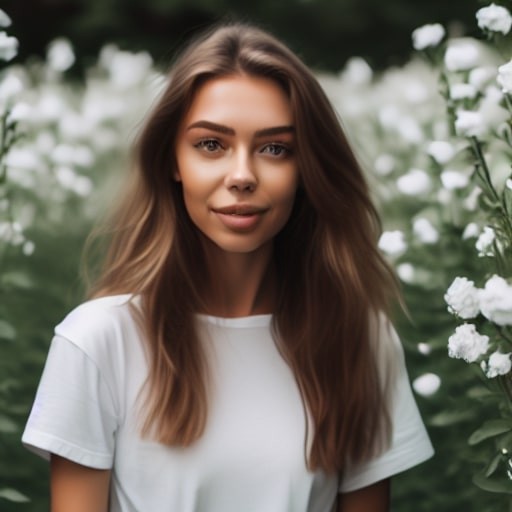}  &
    \includegraphics[width=\linewidth]{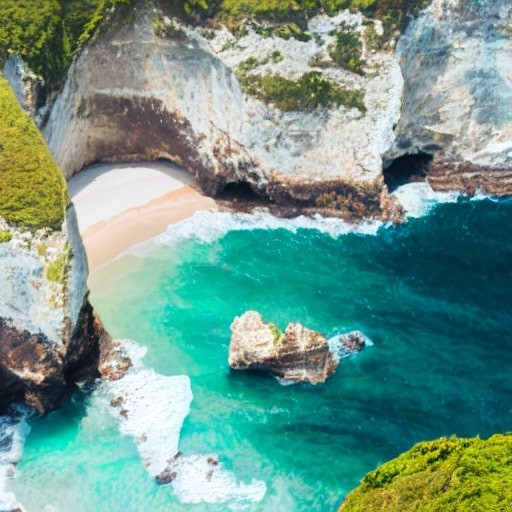}  &
    \includegraphics[width=\linewidth]{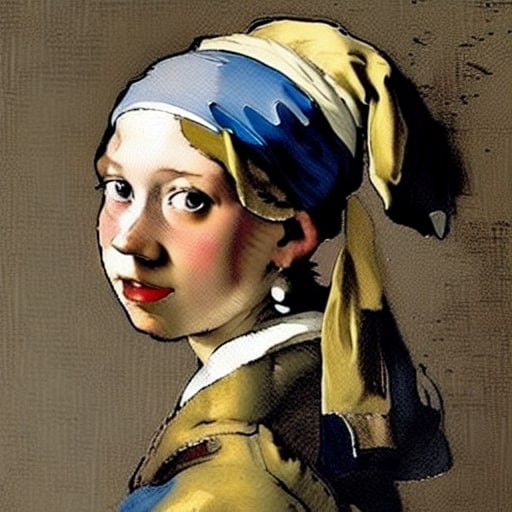}  &
    \includegraphics[width=\linewidth]{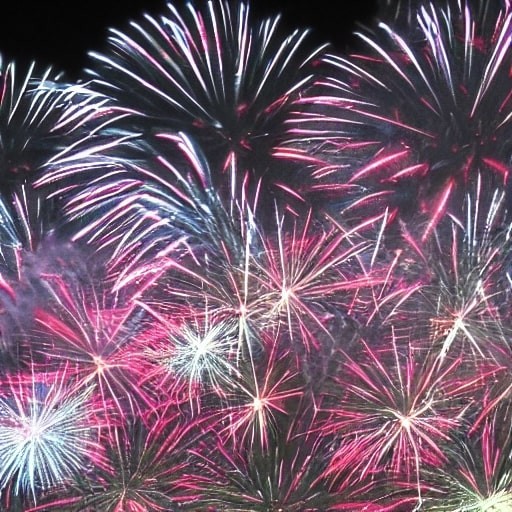}  &
    \includegraphics[width=\linewidth]{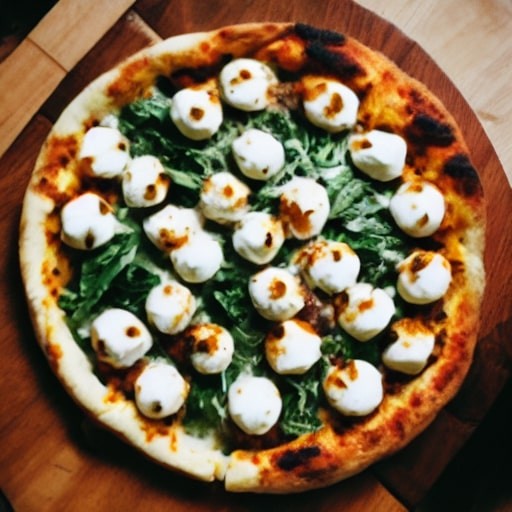}  &
    \includegraphics[width=\linewidth]{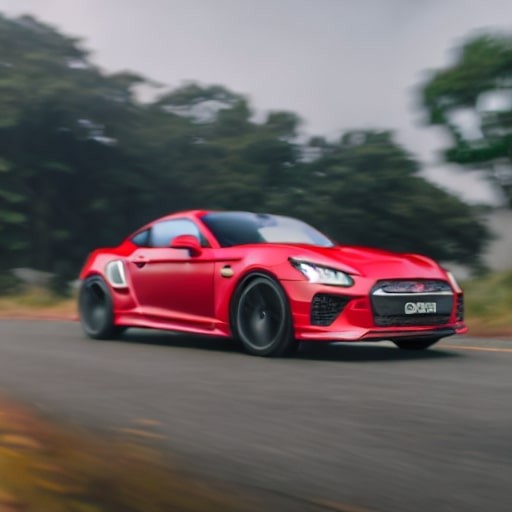}  &
    \includegraphics[width=\linewidth]{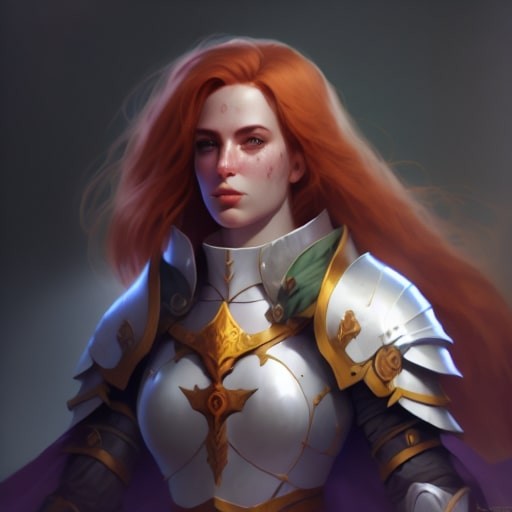}  &
    \includegraphics[width=\linewidth]{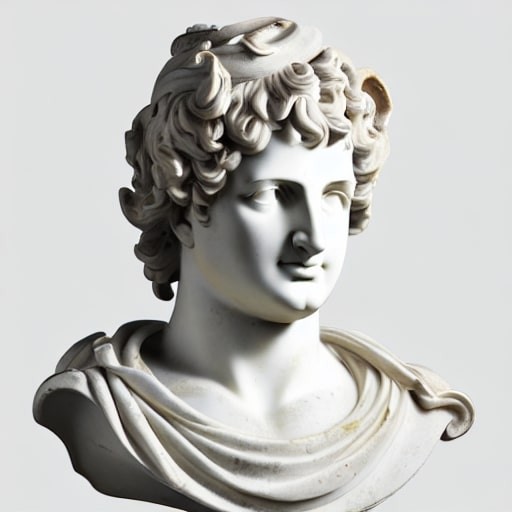}  &
    \includegraphics[width=\linewidth]{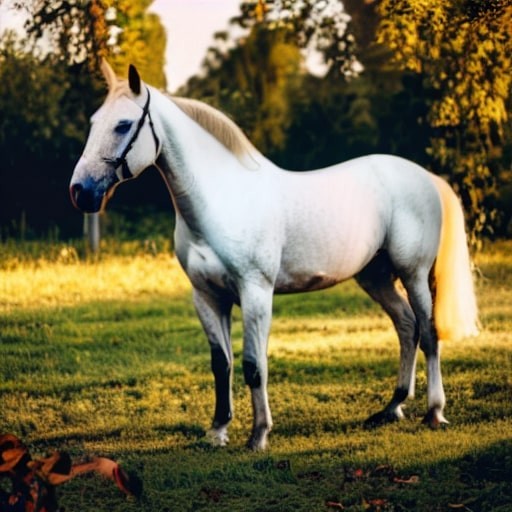}  &
    \includegraphics[width=\linewidth]{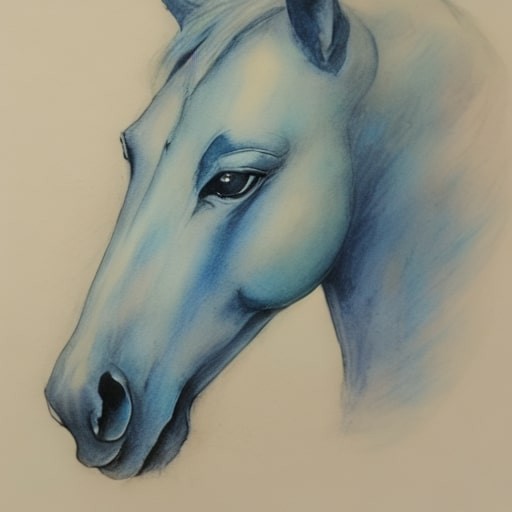}  &
    \includegraphics[width=\linewidth]{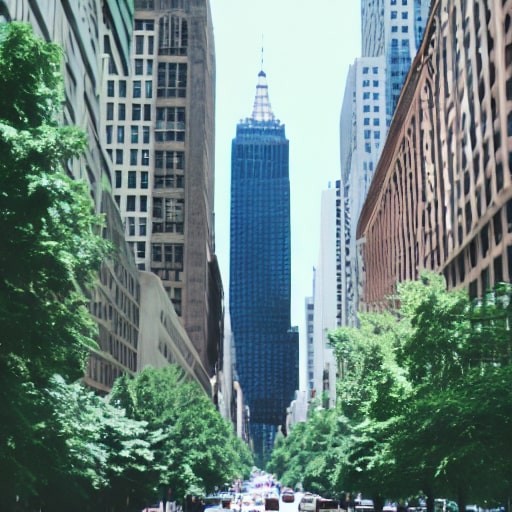}  &
    \includegraphics[width=\linewidth]{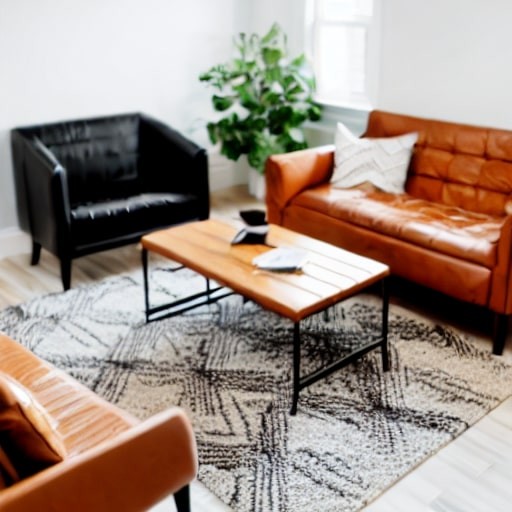}   \\

    \multicolumn{12}{c}{Ours} \\
        
    \includegraphics[width=\linewidth]{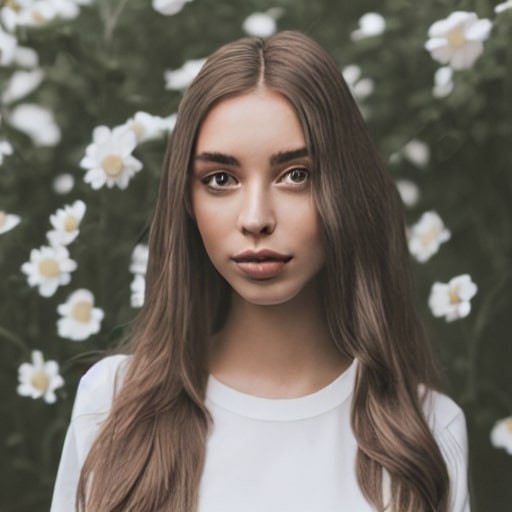} &
    \includegraphics[width=\linewidth]{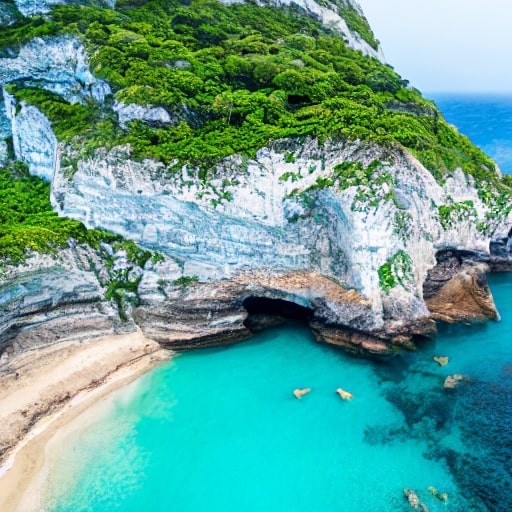} &
    \includegraphics[width=\linewidth]{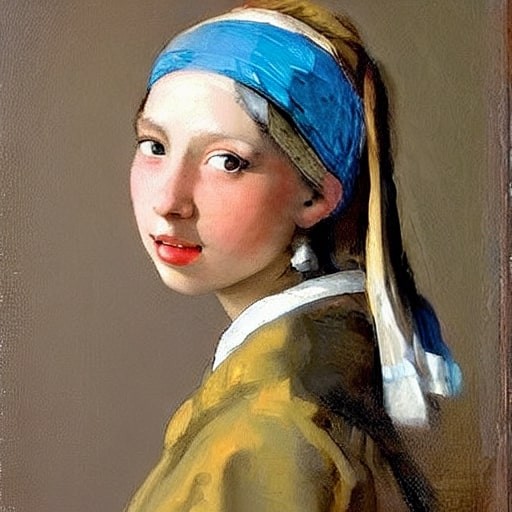}&
    \includegraphics[width=\linewidth]{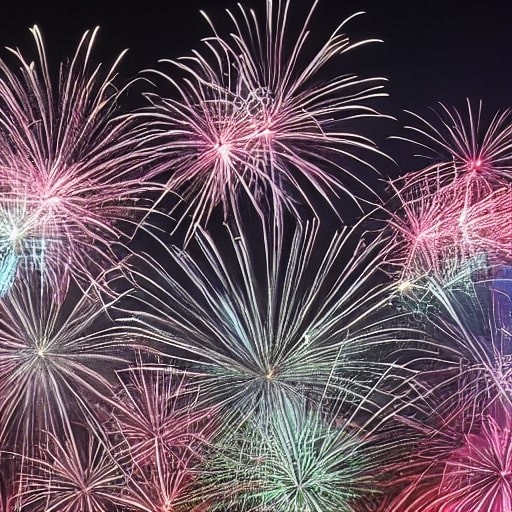} &
    \includegraphics[width=\linewidth]{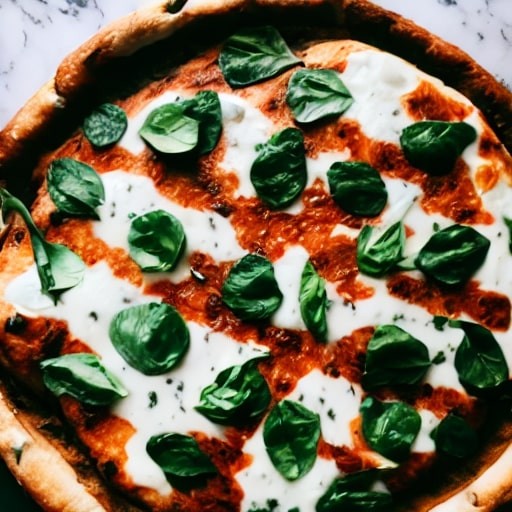} &
    \includegraphics[width=\linewidth]{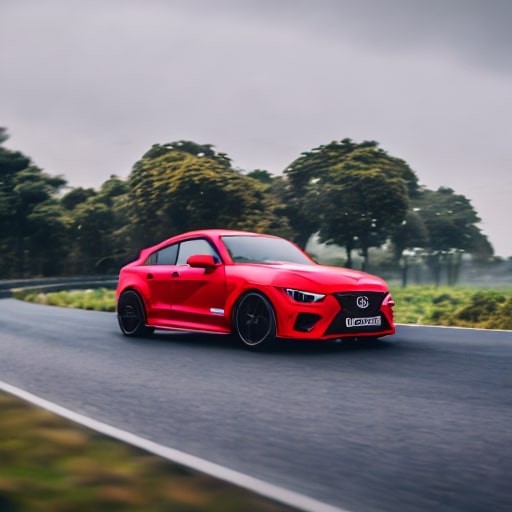}&
    \includegraphics[width=\linewidth]{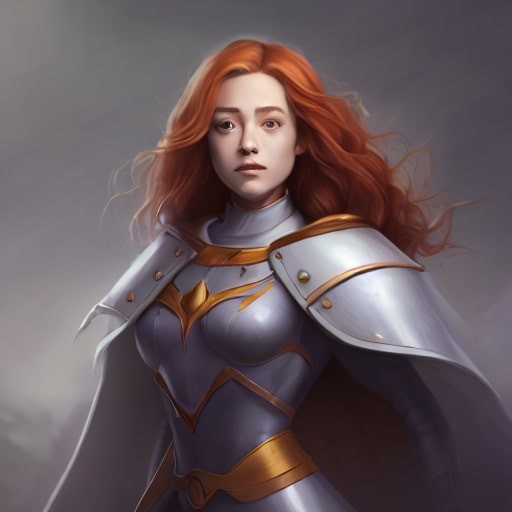} &
    \includegraphics[width=\linewidth]{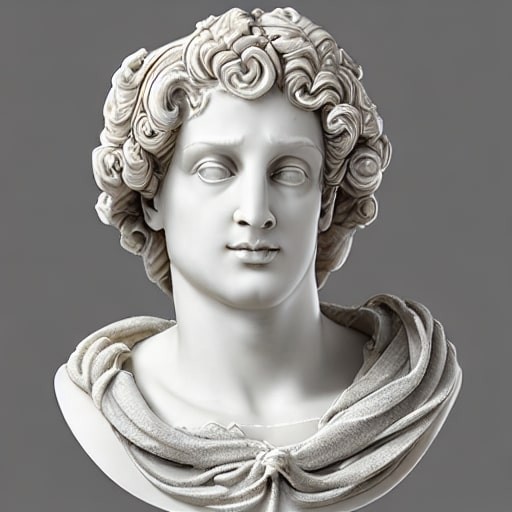} &
    \includegraphics[width=\linewidth]{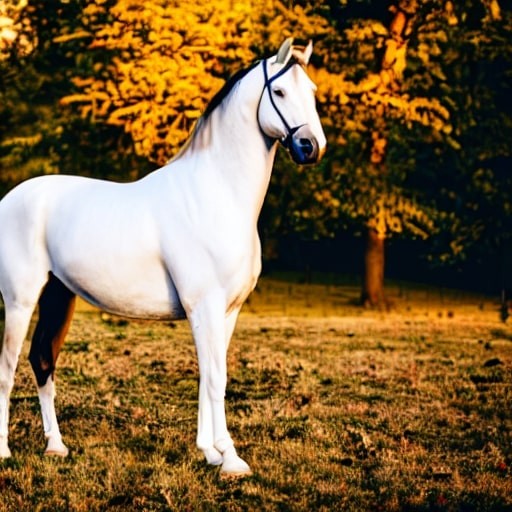}&
    \includegraphics[width=\linewidth]{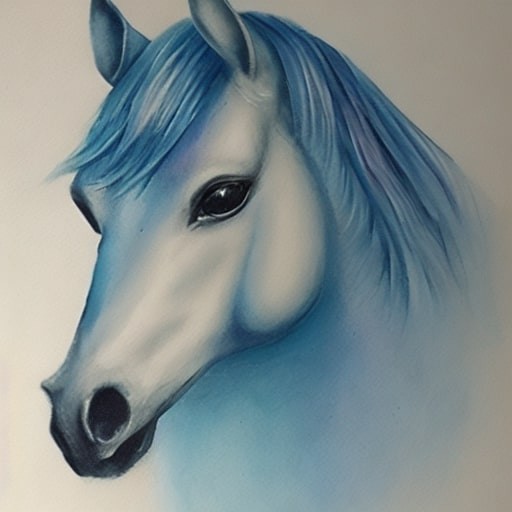} &
    \includegraphics[width=\linewidth]{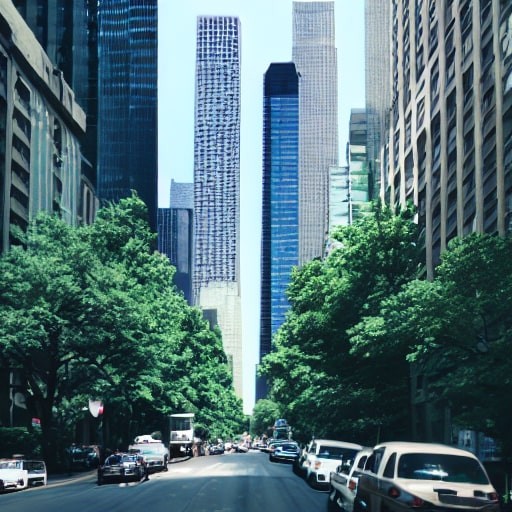} &
    \includegraphics[width=\linewidth]{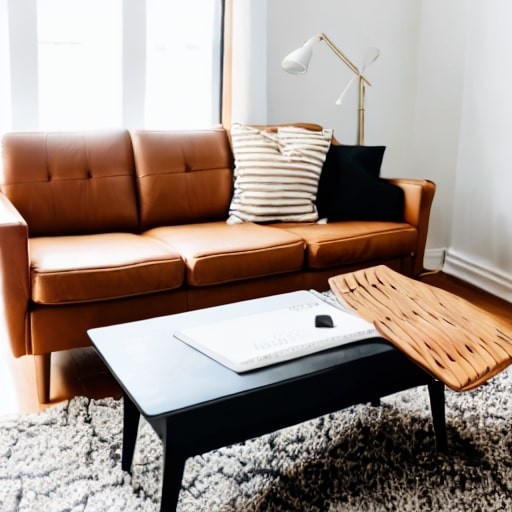} \\

    \multicolumn{12}{c}{Ours (SDXL)} \\
        
    \includegraphics[width=\linewidth]{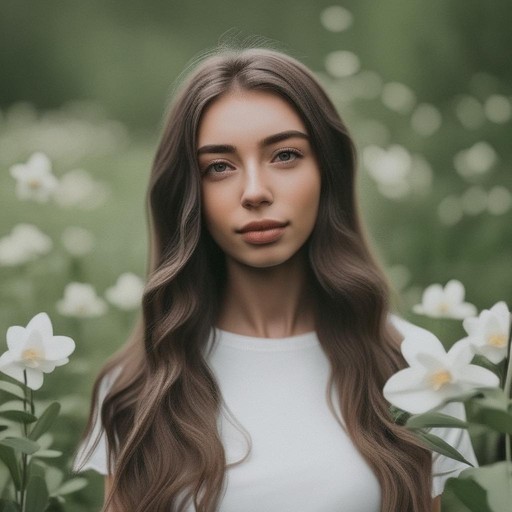} &
    \includegraphics[width=\linewidth]{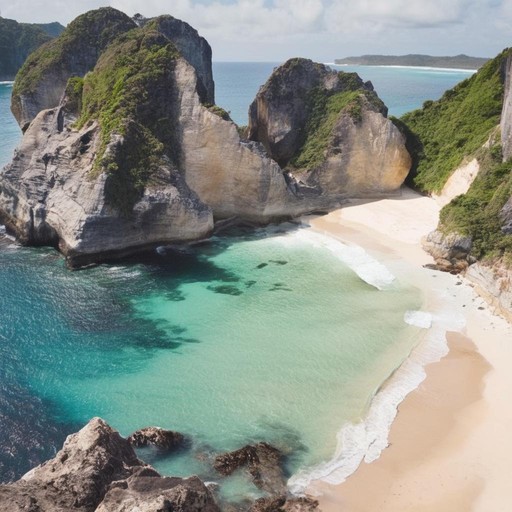} &
    \includegraphics[width=\linewidth]{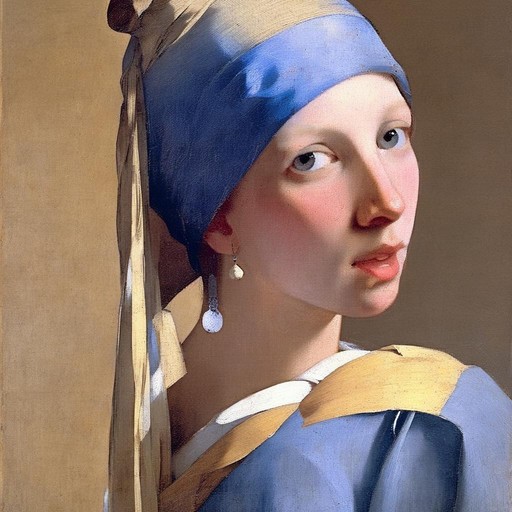}&
    \includegraphics[width=\linewidth]{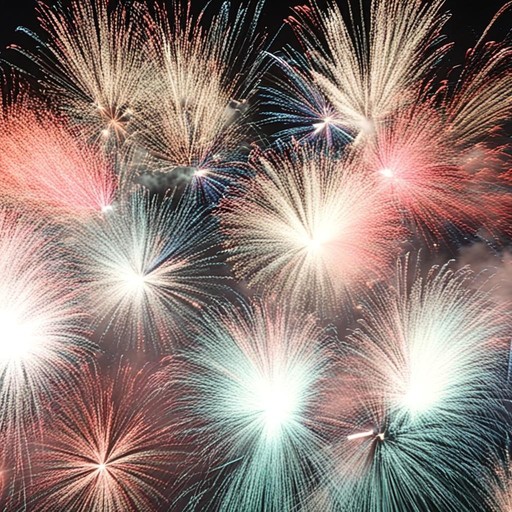} &
    \includegraphics[width=\linewidth]{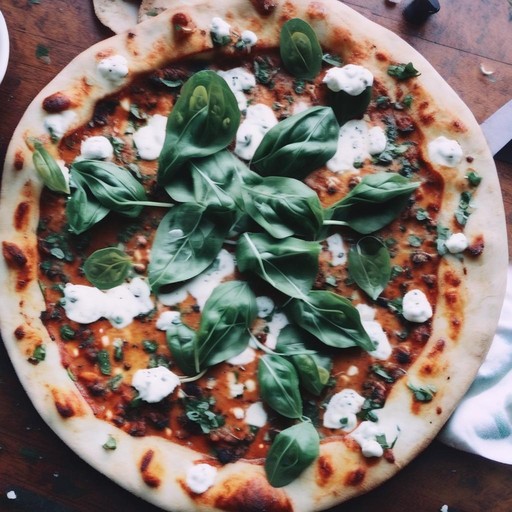} &
    \includegraphics[width=\linewidth]{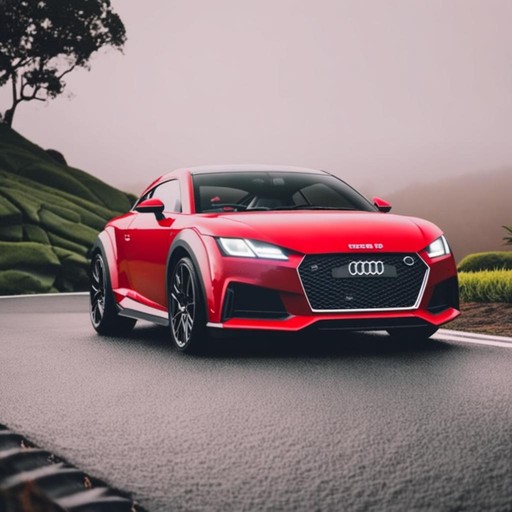} &
    \includegraphics[width=\linewidth]{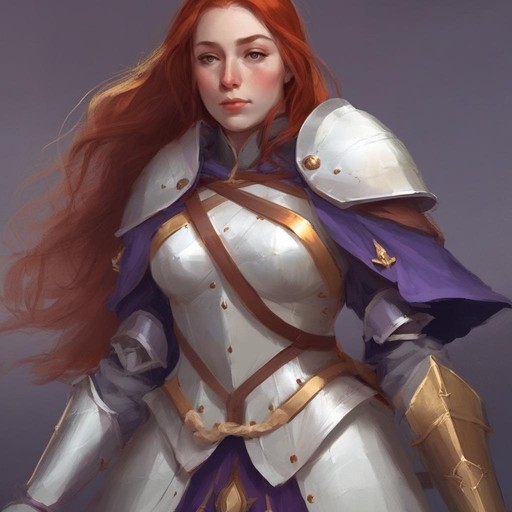} &
    \includegraphics[width=\linewidth]{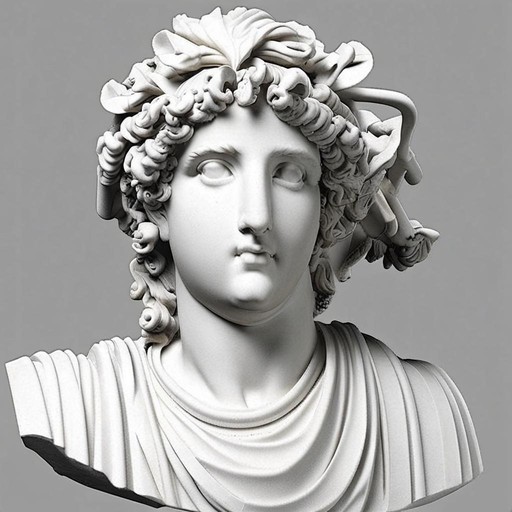}&
    \includegraphics[width=\linewidth]{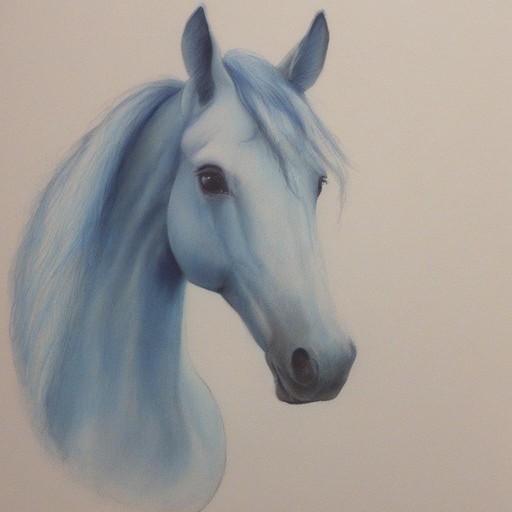} &
    \includegraphics[width=\linewidth]{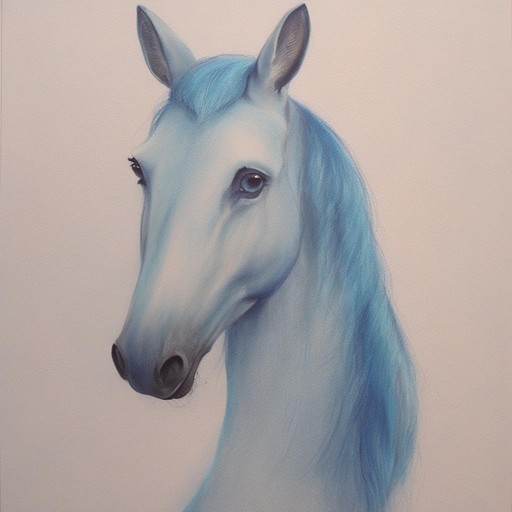} &
    \includegraphics[width=\linewidth]{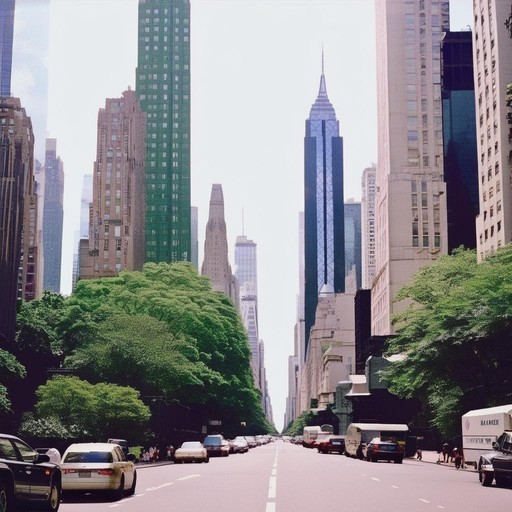} & 
    \includegraphics[width=\linewidth]{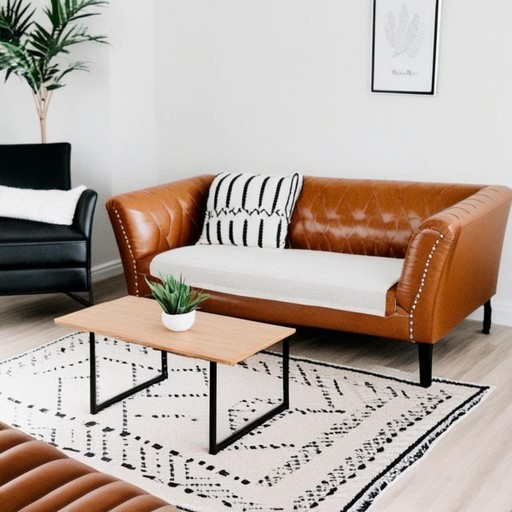} \\

\end{tabular}

\captionof{figure}{Samples from our method with style conditioning compared against other methods. We used an empty prompt and only conditioned on the image. We generally perform on par with IP-Adapter and outperform it on some samples. Note that the third image from the left is less degraded, and the third image from the right captures the mane of the horse better.}

\label{fig:style_main}
\end{table}

\methodname{} obtains state-of-the-art performance on CLIP-I and CLIP-T scores across all adapters, while also being more efficient (approx $27\%$ reduction in the number of trainable parameters compared with IP-Adapter \cite{ye2023ip-adapter}). We attribute this boost in performance to our conditional LoRA approach which can efficiently capture detailed \style{} conditioning in a more effective manner than previous adapter approaches. Interestingly, we find that \methodname{} even outperforms methods that train large models from scratch like unCLIP on CLIP-T score, which further supports our unified conditioning approach.

\subsubsection{Structure}
For evaluating structure control we follow the settings in Uni-ControlNet \cite{zhao2023uni} and sample a single image per condition. To evaluate how closely the generated image is following the provided structure map, we measure the cycle consistency between the predicted structure map of the original image $c = \text{encoder}(x)$ and the predicted structure map of the generated image  $\hat{c} = \text{encoder}(\hat{x})$. The exact metric used to measure the discrepancy depends on the type but $c$ and $\hat{c}$ are always normalized to $[0,1]$ before any metrics computation. The differences in depth maps are measured by the Mean Squared Error between $c$ and $\hat{c}$ (MSE-d). The SSIM (Structural Similarity) is used for HED conditioning. Additionally, we also compute the Learned Perceptual Image Patch Similarity (LPIPS )\cite{zhang2018unreasonable} to implicitly measure the fidelity of the control as well as the Frechet Inception Distance (FID) \cite{heusel2017gans}, which compares the distributions of intermediate features of a pre-trained network between generated and original images.

\setlength{\mycw}{0.10\textwidth}

\begin{table}[tb]
\centering
\scalebox{0.9}{
\begin{tabular}{ p{\mycw} p{\mycw} p{\mycw} p{\mycw} p{\mycw} p{\mycw} p{\mycw} p{\mycw} p{\mycw} }
    \multicolumn{9}{c}{Prompt} \\
    \includegraphics[width=\linewidth]{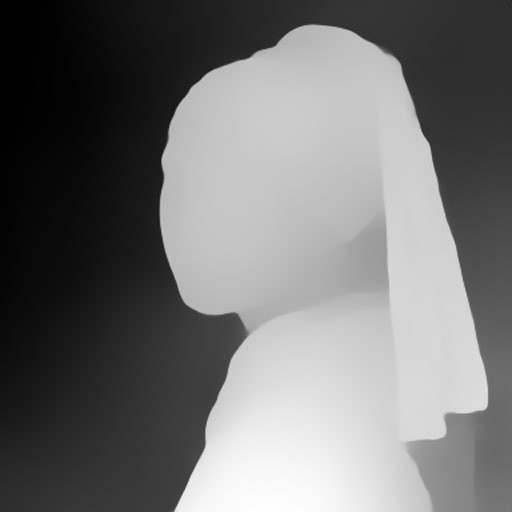} &
    \includegraphics[width=\linewidth]{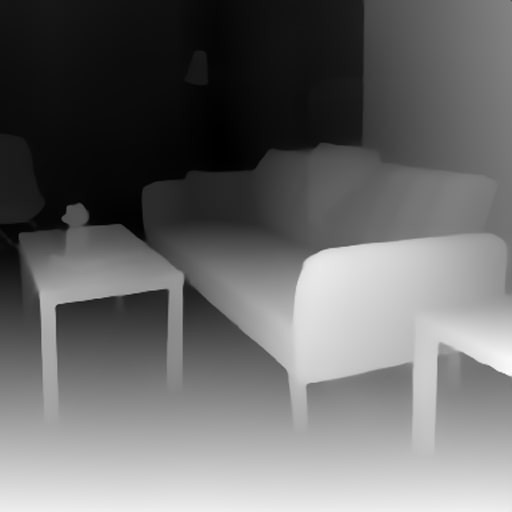} &
    \includegraphics[width=\linewidth]{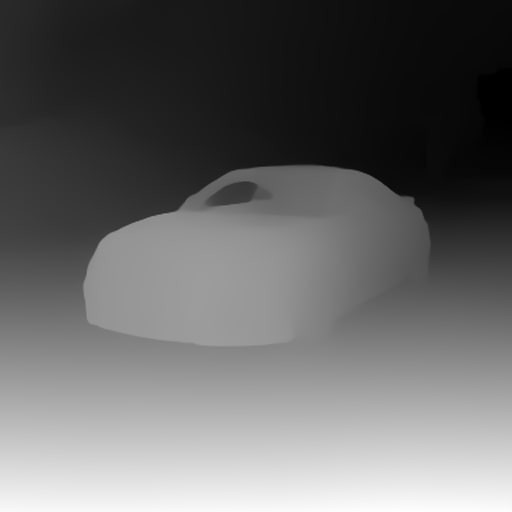} &
    \includegraphics[width=\linewidth]{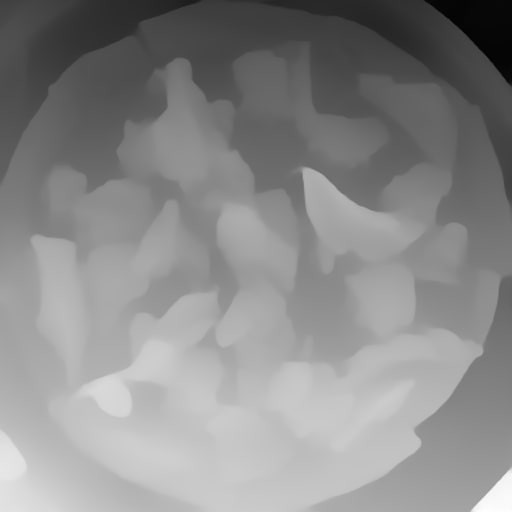} &
    \includegraphics[width=\linewidth]{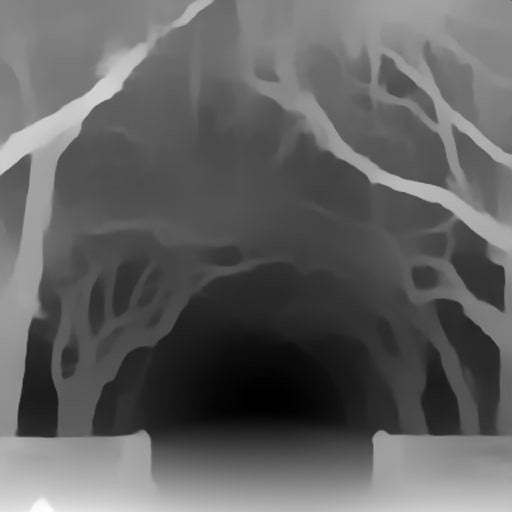} &
    \includegraphics[width=\linewidth]{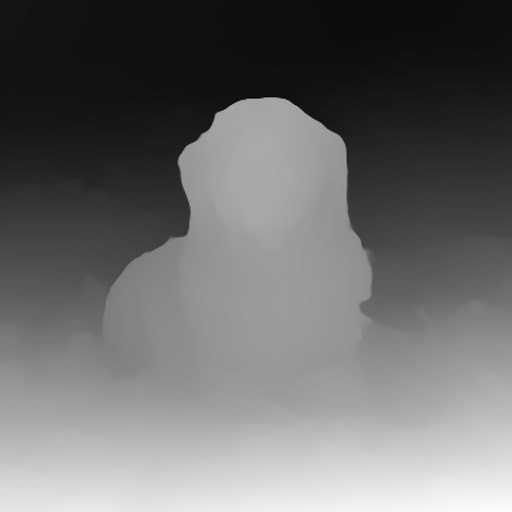} &
    \includegraphics[width=\linewidth]{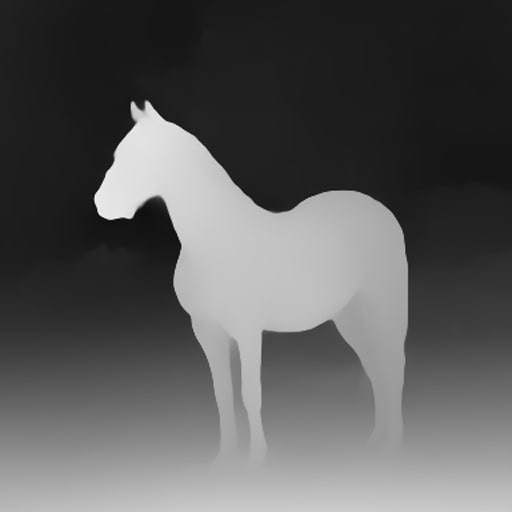} &
    \includegraphics[width=\linewidth]{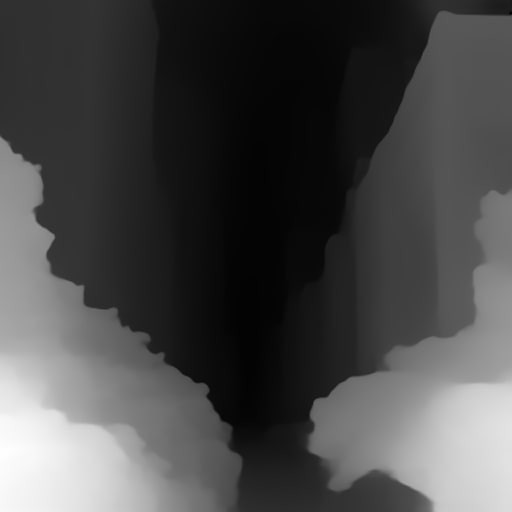} &
    \includegraphics[width=\linewidth]{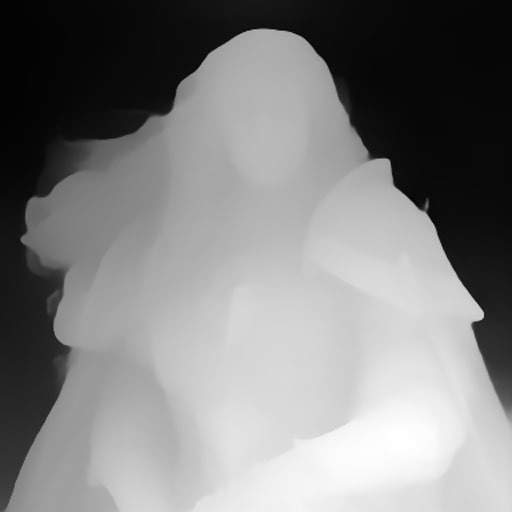} \\

    \multicolumn{9}{c}{ControlNet} \\
    \includegraphics[width=\linewidth]{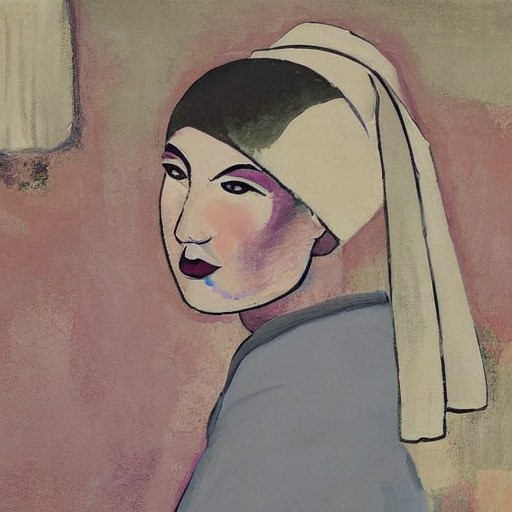}  &
    \includegraphics[width=\linewidth]{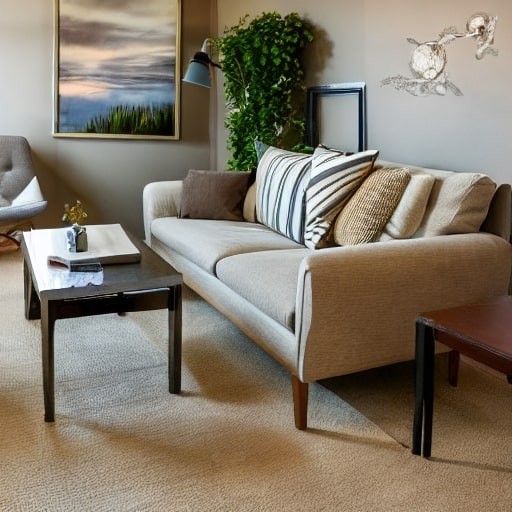}  &
    \includegraphics[width=\linewidth]{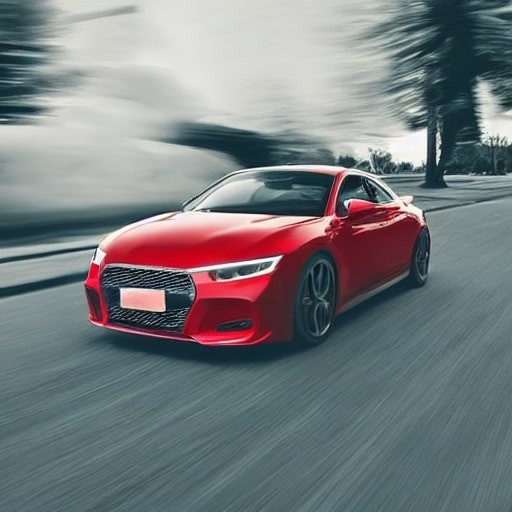}  &
    \includegraphics[width=\linewidth]{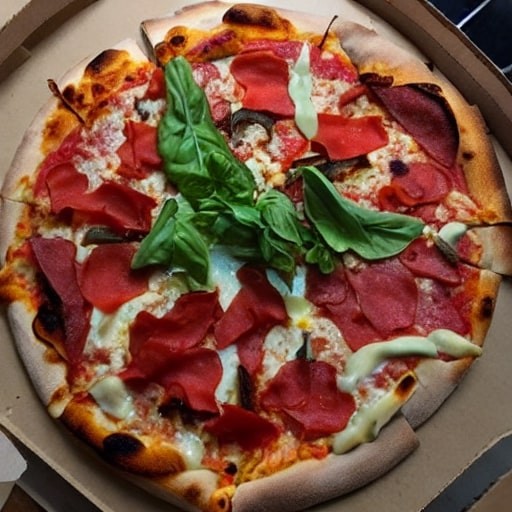}  &
    \includegraphics[width=\linewidth]{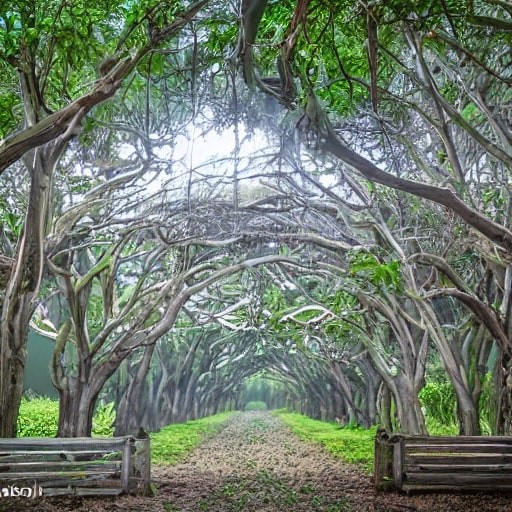}  &
    \includegraphics[width=\linewidth]{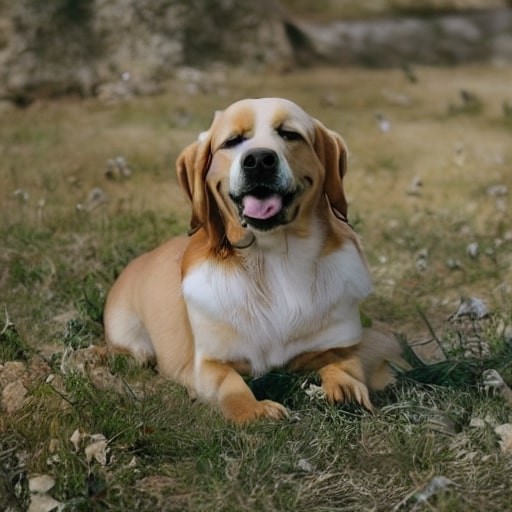}  &
    \includegraphics[width=\linewidth]{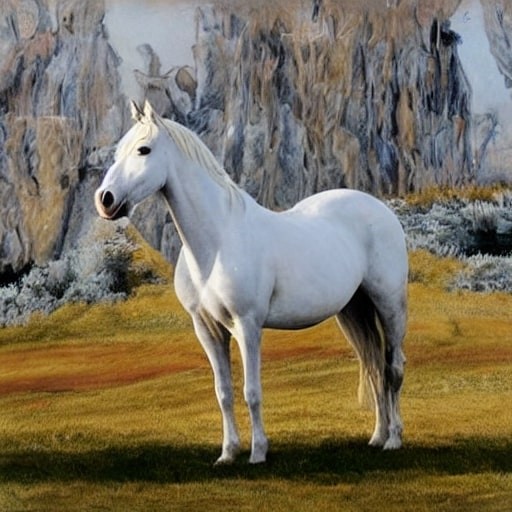}  &
    \includegraphics[width=\linewidth]{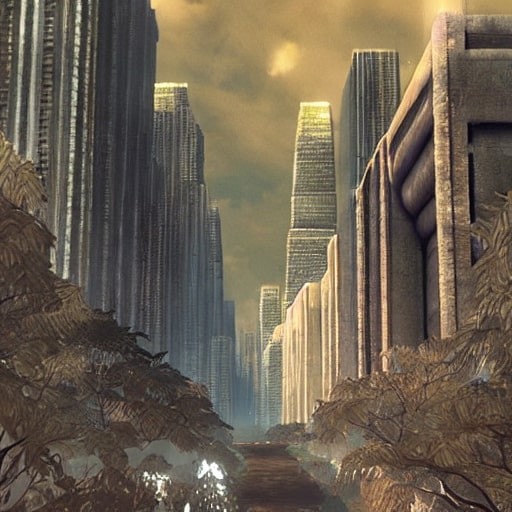}  &
    \includegraphics[width=\linewidth]{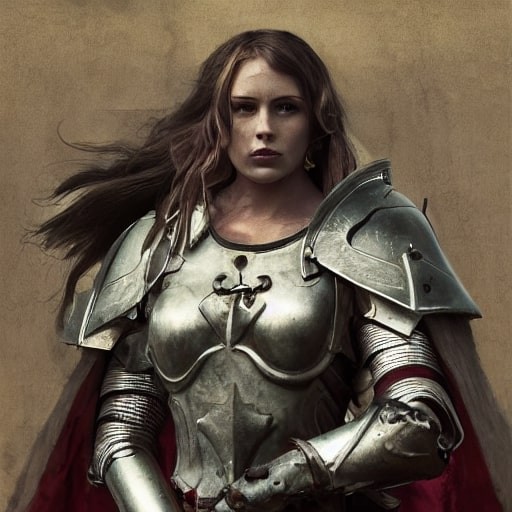}  \\

    \multicolumn{9}{c}{T2I Adapter} \\
    \includegraphics[width=\linewidth]{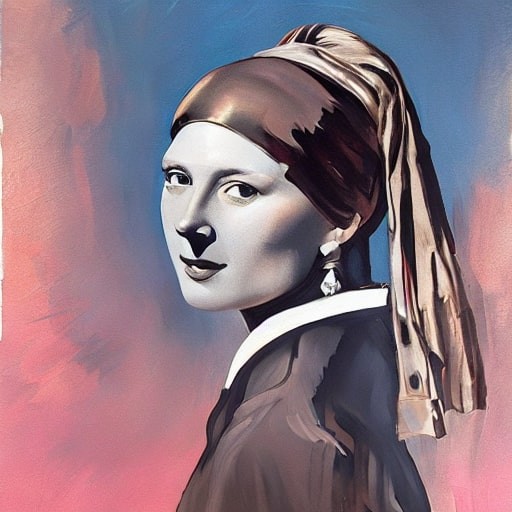}  &
    \includegraphics[width=\linewidth]{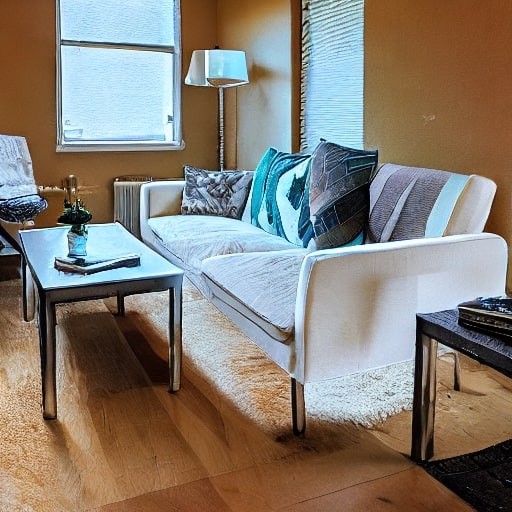}  &
    \includegraphics[width=\linewidth]{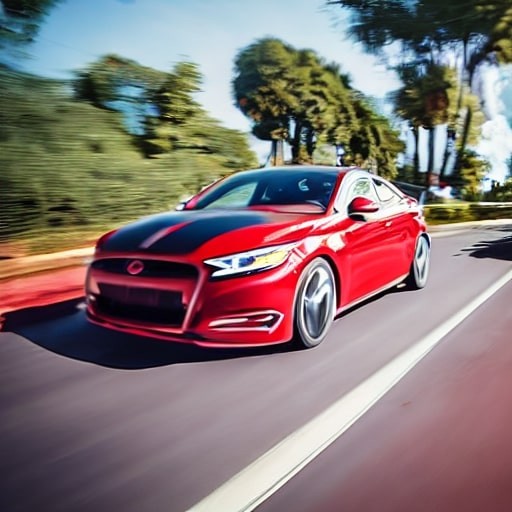}  &
    \includegraphics[width=\linewidth]{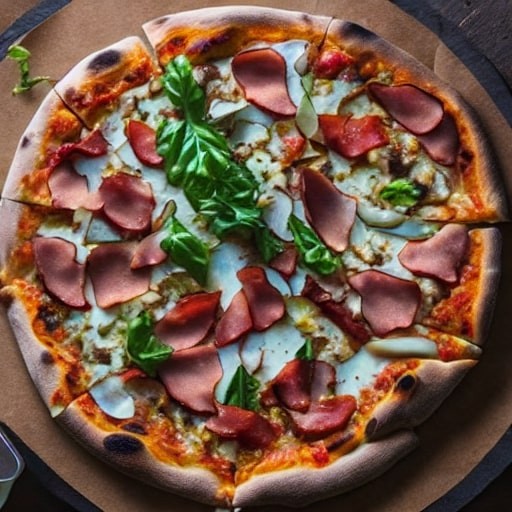}  &
    \includegraphics[width=\linewidth]{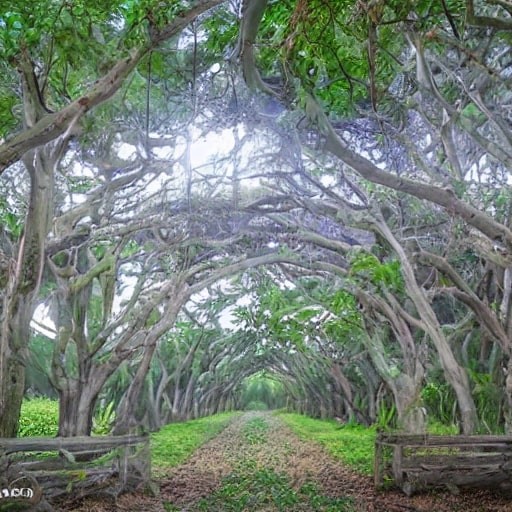}  &
    \includegraphics[width=\linewidth]{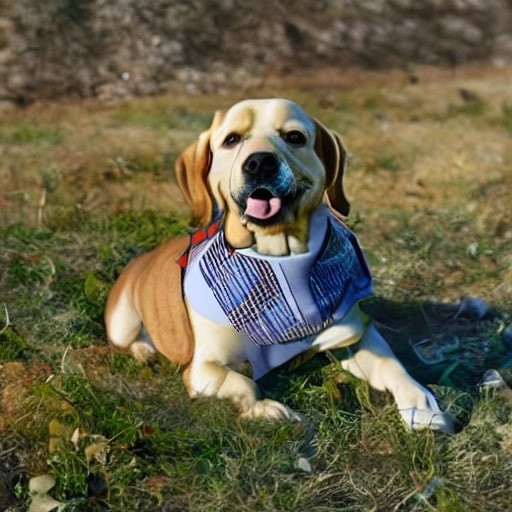}  &
    \includegraphics[width=\linewidth]{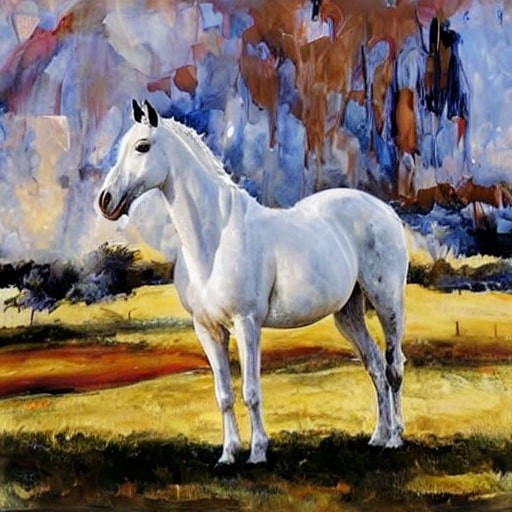}  &
    \includegraphics[width=\linewidth]{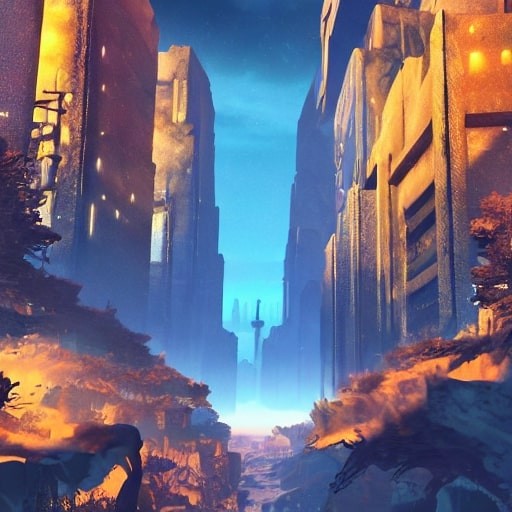}  &
    \includegraphics[width=\linewidth]{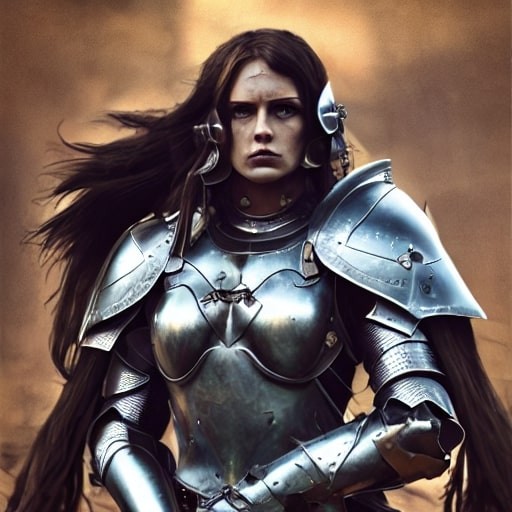}  \\

    \multicolumn{9}{c}{Uni-ControlNet} \\
    \includegraphics[width=\linewidth]{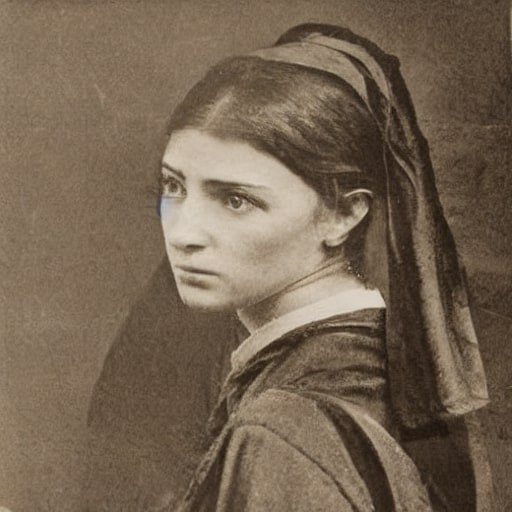}  &
    \includegraphics[width=\linewidth]{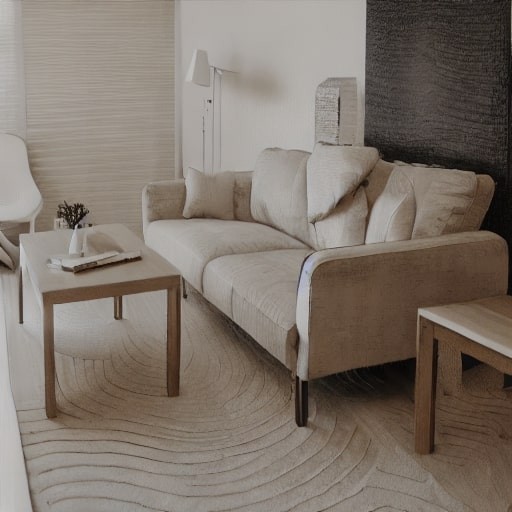}  &
    \includegraphics[width=\linewidth]{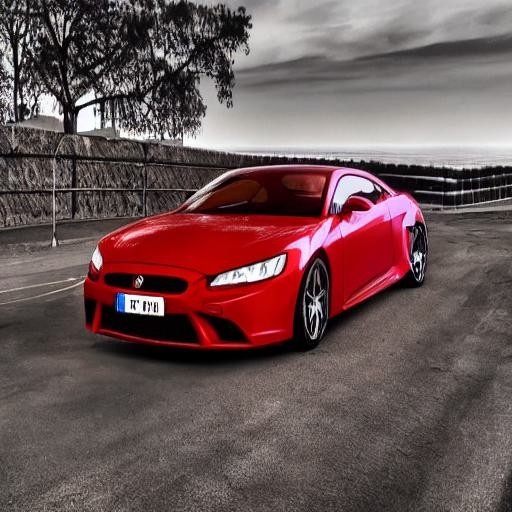}  &
    \includegraphics[width=\linewidth]{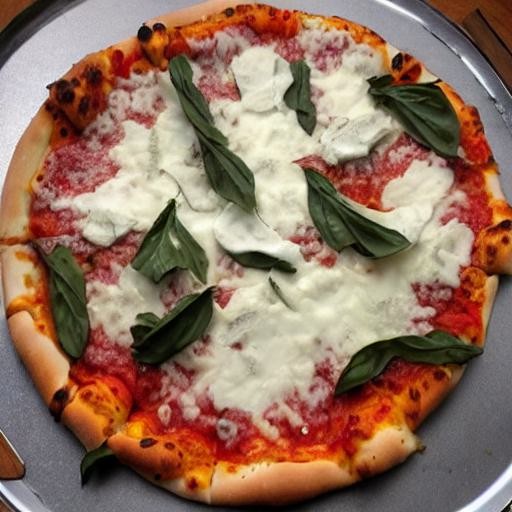}  &
    \includegraphics[width=\linewidth]{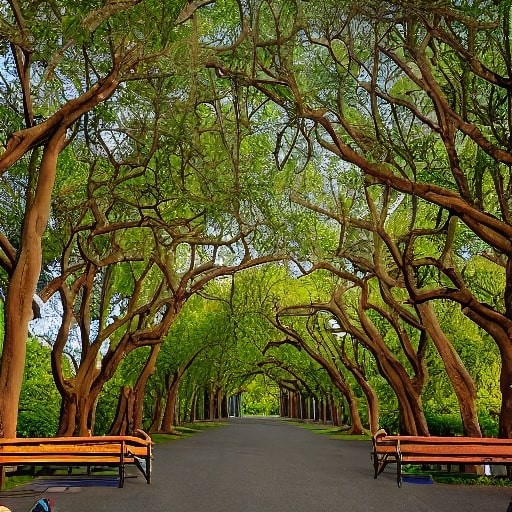}  &
    \includegraphics[width=\linewidth]{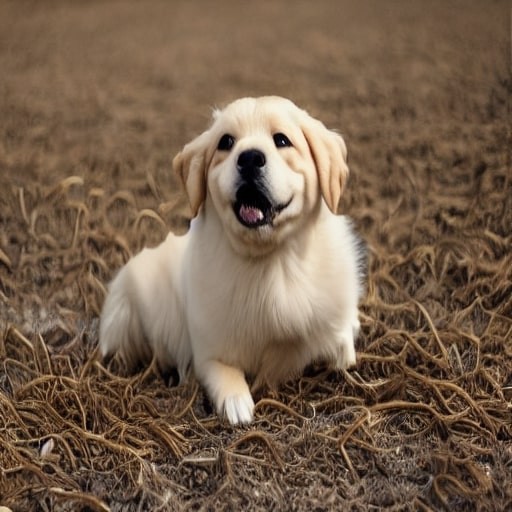}  &
    \includegraphics[width=\linewidth]{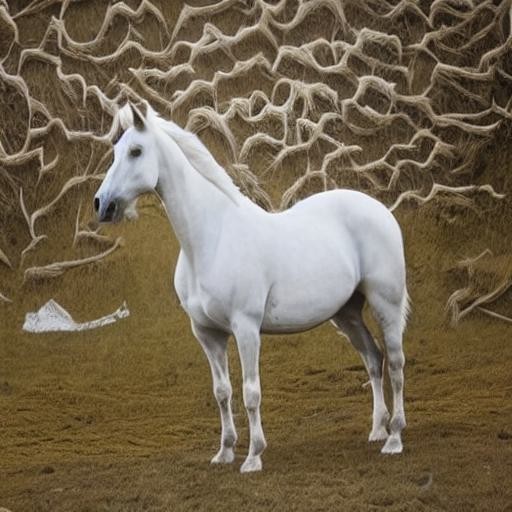}  &
    \includegraphics[width=\linewidth]{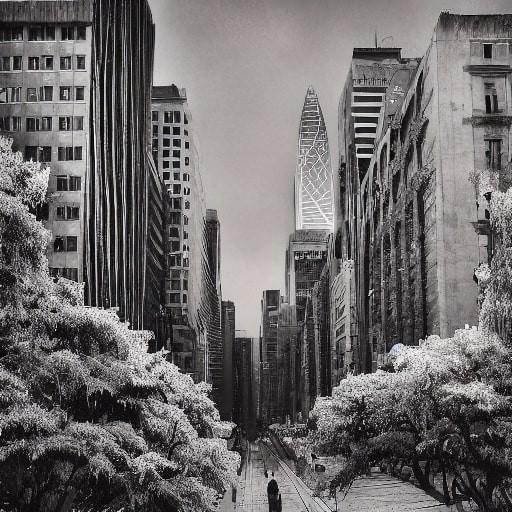}  &
    \includegraphics[width=\linewidth]{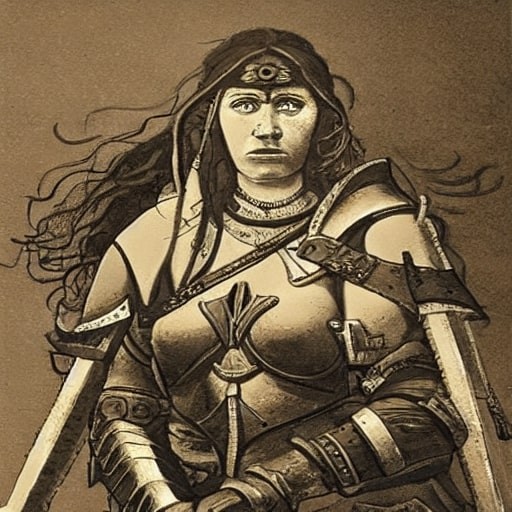}  \\

    \multicolumn{9}{c}{Ours} \\
    \includegraphics[width=\linewidth]{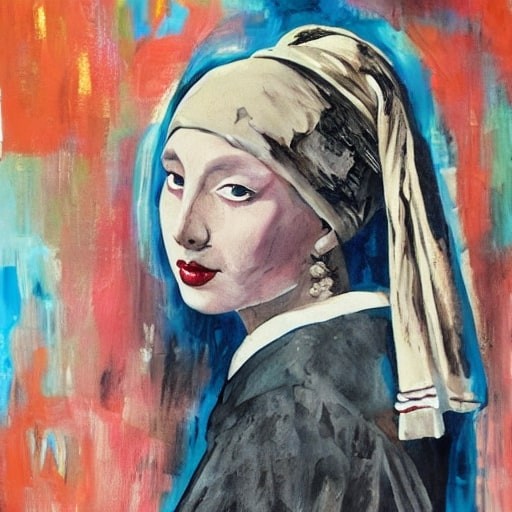} &
    \includegraphics[width=\linewidth]{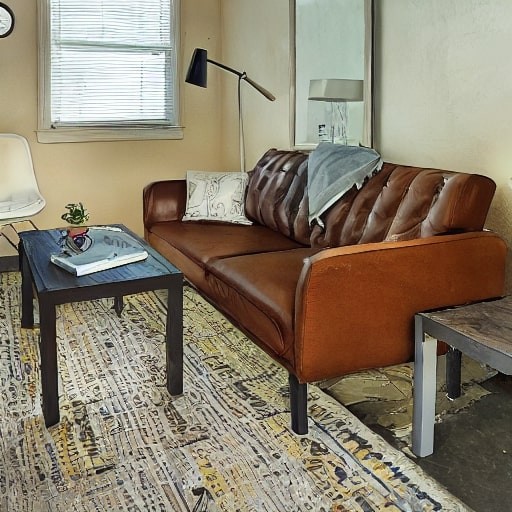} &
    \includegraphics[width=\linewidth]{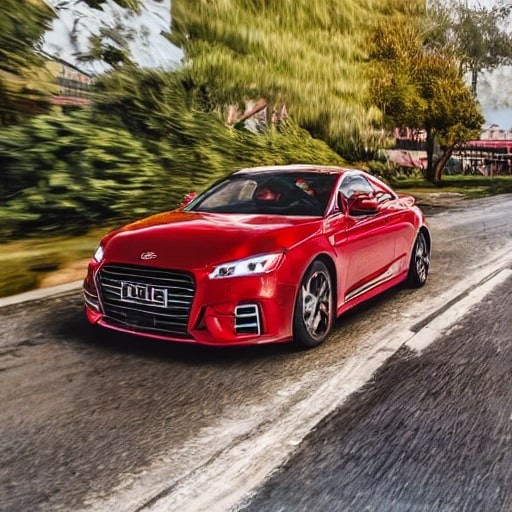} &
    \includegraphics[width=\linewidth]{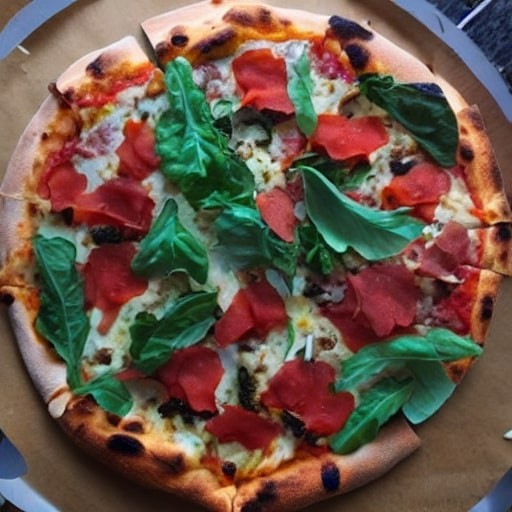} &
    \includegraphics[width=\linewidth]{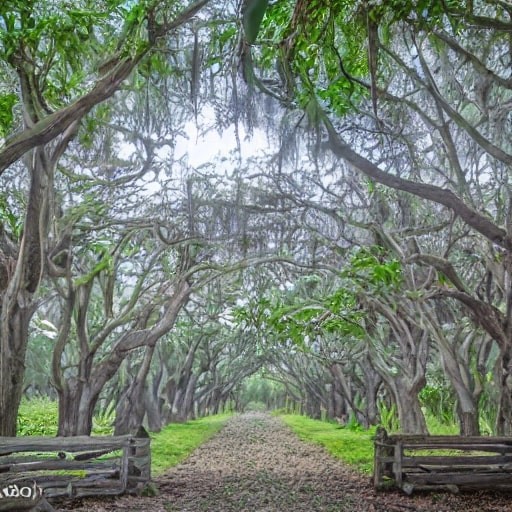} &
    \includegraphics[width=\linewidth]{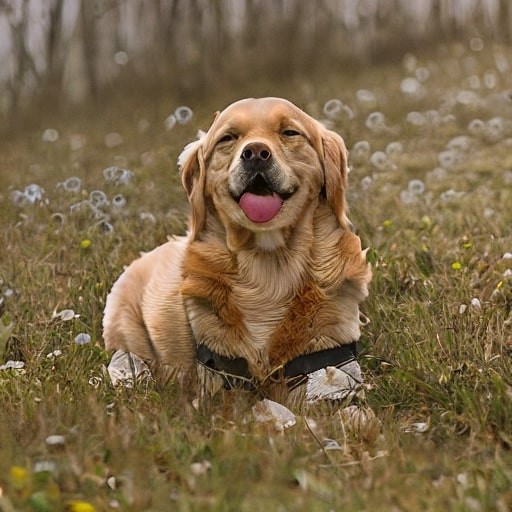} &
    \includegraphics[width=\linewidth]{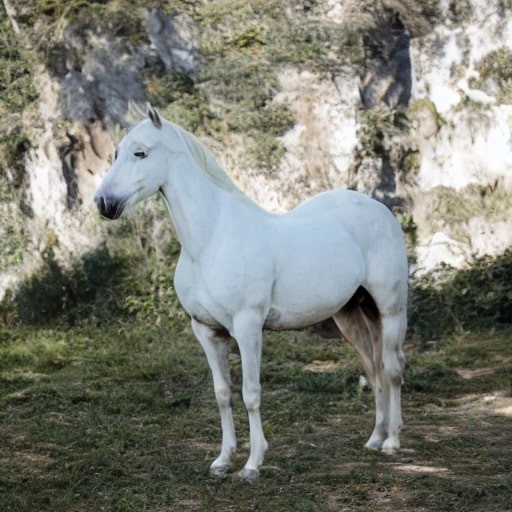} &
    \includegraphics[width=\linewidth]{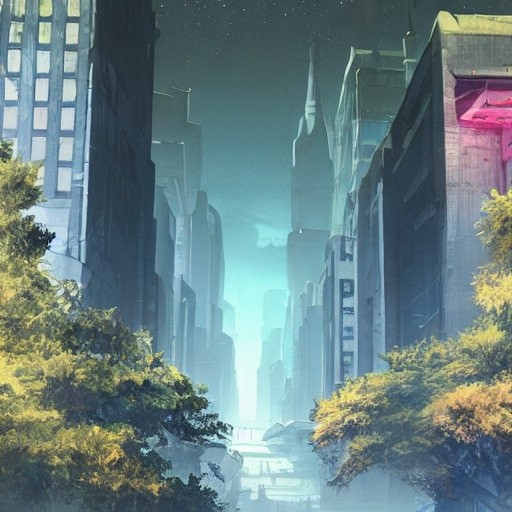} &
    \includegraphics[width=\linewidth]{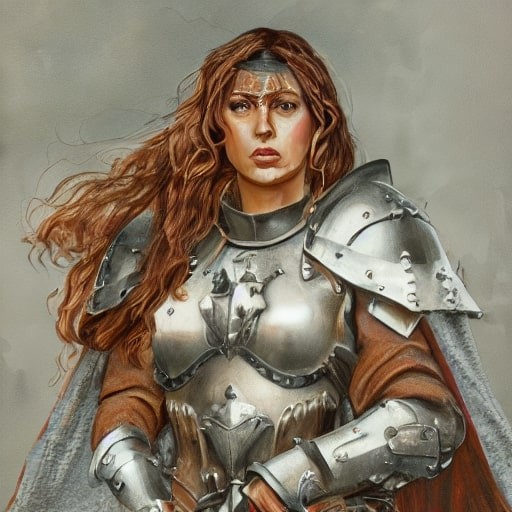} \\
\end{tabular}
}
\captionof{figure}{Samples from our method with structural conditioning compared against other methods. Note that for our method, especially compared with T2I Adapter, the details of the images are substantially more closely aligned with the depth prompt (see, \eg, the lamp in the background of the living room scene and the side table's legs, or the salad on the pizza)}

\label{fig:struct_main}
\end{table}

Table \ref{tab:main_struct} shows results across metrics for different approaches. In particular, adapters that focus on \structure{} conditioning tend to use more parameters than ones used for style therefore we trained two versions of \methodname{}. One with 17M (roughly matching the number of parameters used for \style{} conditioning) and 32M parameters, which is still less than T2I-Adapter \cite{mou2023t2i}.  \methodname{} obtains state-of-the-art performance across the board, even when reducing parameters by half compared to T2I-Adapter. These results show that \methodname{} is a very efficient approach for \structure{} conditioning which can be scaled to obtain high-performant models. We refer the reader to the Appendix for additional results and comparisons.

\subsection{Qualitative Comparison}

To show qualitative results for both \structure{} and \style{} conditioning we take images used for conditioning by previous approaches while also adding some conditioning images selected by us.  Results are shown in Fig. \ref{fig:style_main} where we can see that \methodname{} better captures the style of the conditioning image compared with previous adapter approaches. We show additional comparison with more models on an extended set of samples in the Appendix.

We also show samples for structure conditioning on depth maps predicted by MiDaS \cite{ranftl2020towards} in Fig \ref{fig:struct_main}. \methodname{} shows nice sample quality and superior adherence to the fine-grained depth structure compared to ControlNet or T2I-Adapter. We attribute this improvement to our choice of directly adapting the convolutions in the U-Net as opposed to an indirect adaption via skip connections as previous approaches have proposed \cite{mou2023t2i, zavadski2023controlnetxs}.

\subsection{Ablation experiments}
To showcase the modularity of our approach, we train multiple configurations per condition and assess their performance. We study the influence of the LoRAs rank as well as the specific layer choice, which is further discussed for each conditioning type.

\setlength{\mycw}{0.1\textwidth}
\newcolumntype{P}[1]{>{\centering\arraybackslash}p{#1}}

\npdecimalsign{.}
\nprounddigits{3}
\begin{figure}[tbp]
    \begin{minipage}[b]{0.48\linewidth}
        \centering
        \scalebox{0.8}{
        \begin{tabular}{ P{\mycw} | P{\mycw} P{\mycw} P{\mycw} P{\mycw} }

        & \multicolumn{4}{c}{LoRA Rank} \\

       Prompt & 16 &  32 &  64 &   128 \\
    
        \midrule

        \includegraphics[width=\linewidth]{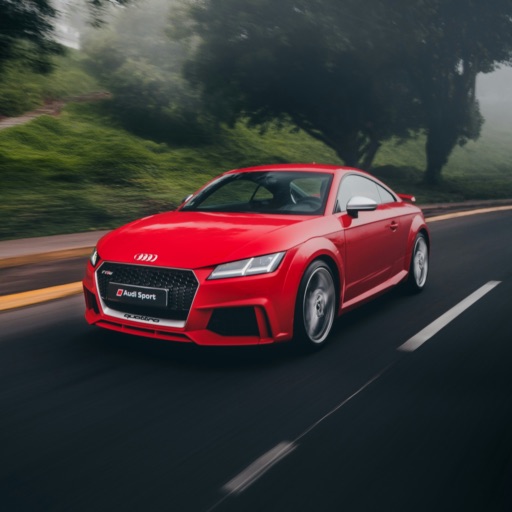} &
        \includegraphics[width=\linewidth]{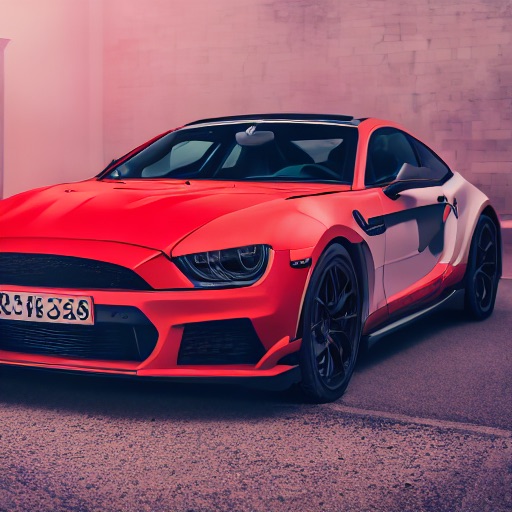} &
        \includegraphics[width=\linewidth]{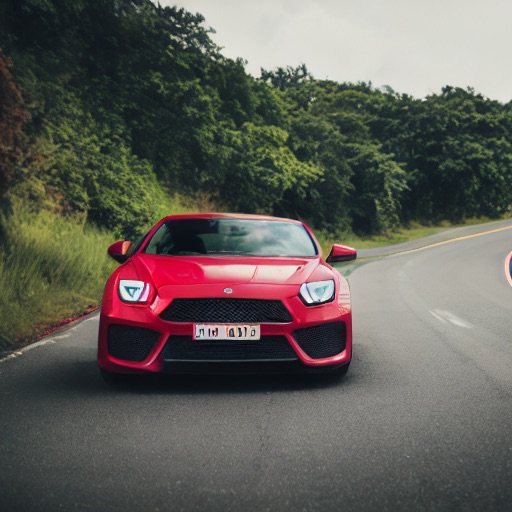} &
        \includegraphics[width=\linewidth]{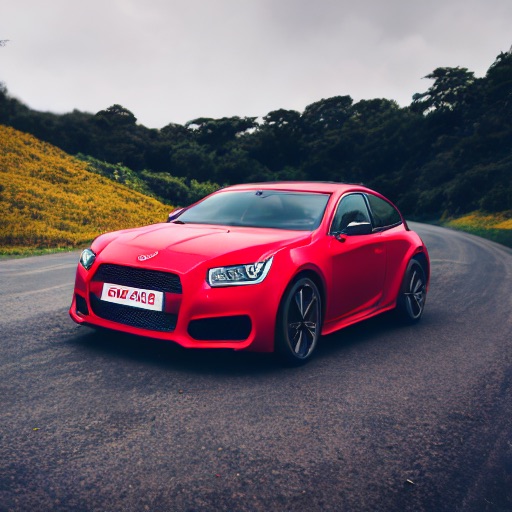} &
        \includegraphics[width=\linewidth]{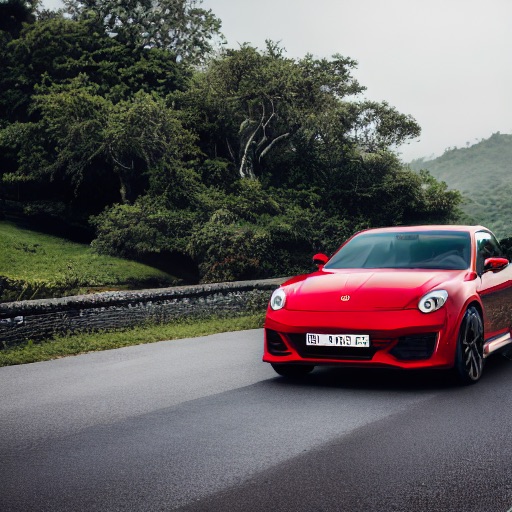} \\
    
        \includegraphics[width=\linewidth]{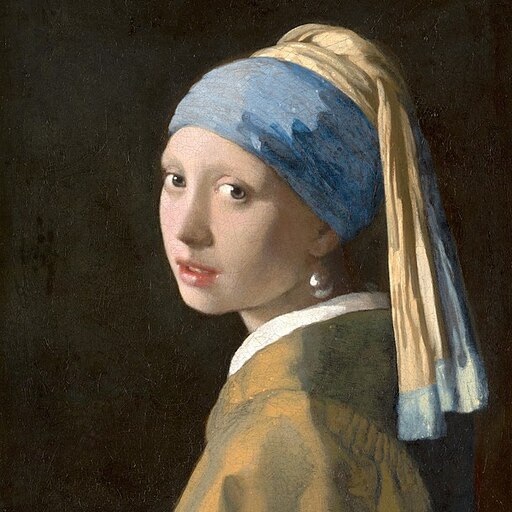} &
        \includegraphics[width=\linewidth]{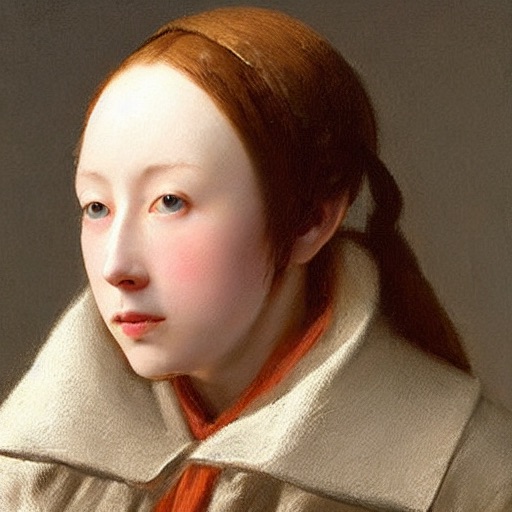} &
        \includegraphics[width=\linewidth]{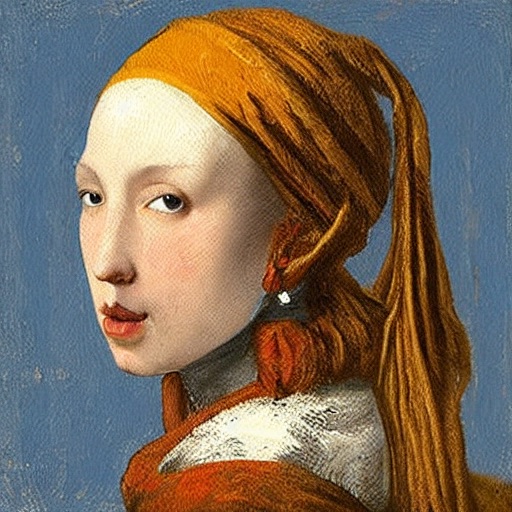} &
        \includegraphics[width=\linewidth]{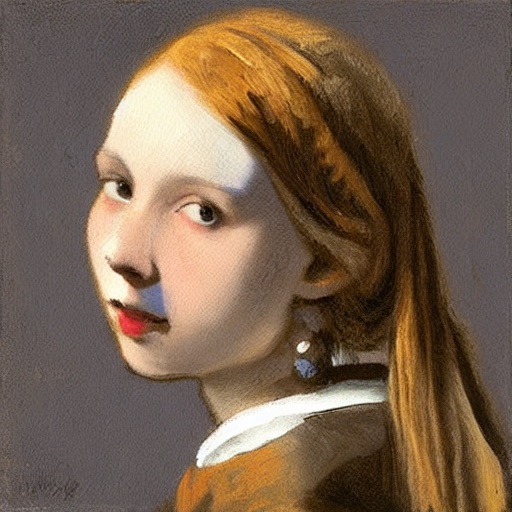} &
        \includegraphics[width=\linewidth]{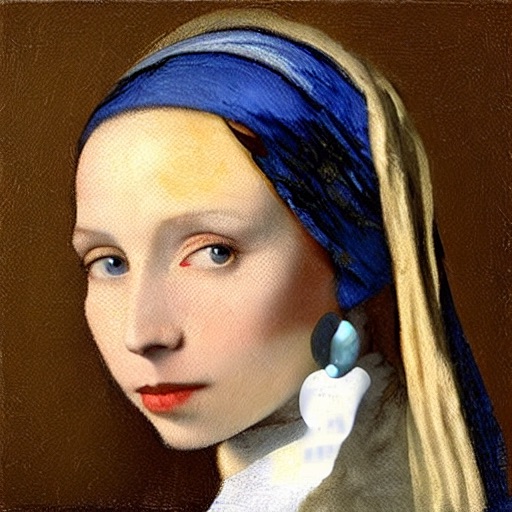} \\
       \includegraphics[width=\linewidth]{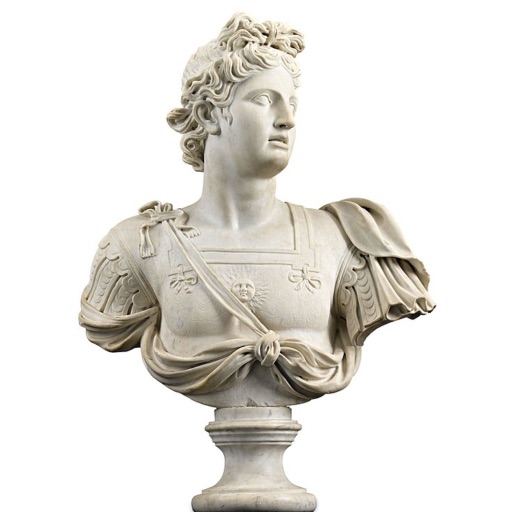} &
       \includegraphics[width=\linewidth]{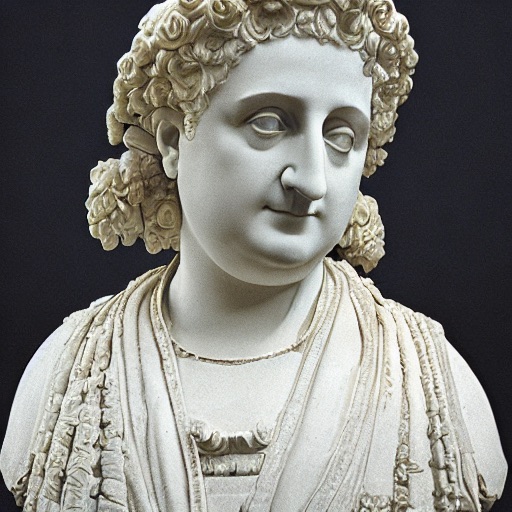} &
        \includegraphics[width=\linewidth]{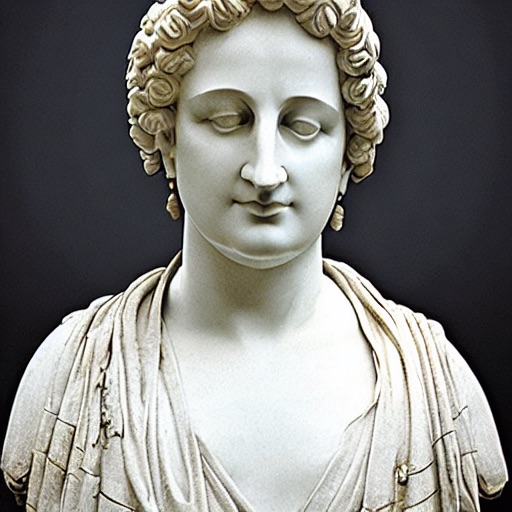} &
        \includegraphics[width=\linewidth]{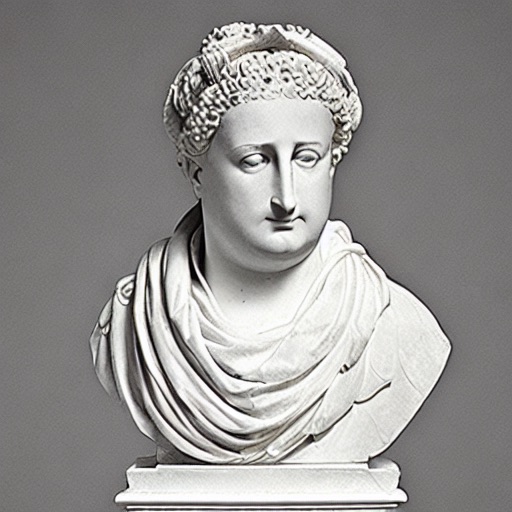} &
        \includegraphics[width=\linewidth]{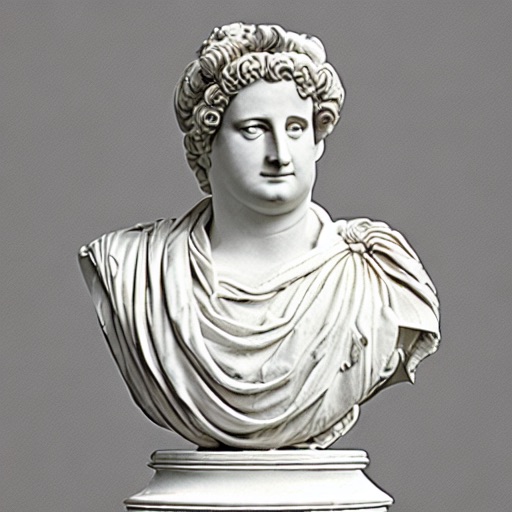} \\

    \end{tabular}
    }
    \label{fig:style_rank}
    \captionof{figure}{We show how the choice of the LoRA rank affects the generation quality. The rank refers to the cross-attention models in Table \ref{tab:style_size_abl}. Even the smallest model with only 2M parameters can meaningfully capture the content of the image albeit with minor losses in quality.  }
    \end{minipage}
    \hspace{0.03\linewidth} %
    \begin{minipage}[b]{0.48\linewidth}
        \centering
        \vspace{0pt}
        \captionof{table}{Evaluation of different global conditional LoRA configurations. All models were trained using CLIP VIT-L/14 as an image encoder and evaluated on 5000 samples. Adapting the Cross-Attention layers yields the best performance. Even very low-rank LoRA conditioning can still produce good performance.}
        \scalebox{.65}{
        \begin{tabular}{c c c n{2}{3} n{2}{3} }
        \toprule
            \textbf{Layer} & \textbf{Rank} & \textbf{Parameters} & \textbf{CLIP-I} & \textbf{CLIP-T}   \\
            \midrule
             Cross Attention & 208 & 21M & 0.8367517590522766 & 0.6307616829872131 \\
             Cross Attention & 128 & 13M & 0.8346480131149292 & 0.6338164210319519 \\
             Cross Attention & {\color{white}0}64 & {\color{white}0}7M & 0.8289241790771484 & 0.637165904045105 \\
             Cross Attention & {\color{white}0}32 & {\color{white}0}4M & 0.8156231045722961 & 0.6417122483253479 \\
             Cross Attention & {\color{white}0}16 & {\color{white}0}2M & 0.7992510795593262 & 0.6494192779064178 \\
    
             \cdashlinelr{1-5}
    
             Self Attention & 128 & 13M & 0.8282693028450012 & 0.638936385512352 \\
    
             \cdashlinelr{1-5}
    
             Cross \& Self Attention & 128 & 26M & 0.8422175645828247 & 0.6286923587322235 \\
             Cross \& Self Attention & {\color{white}0}64 & 13M & 0.8365278840065002 & 0.6355270743370056 \\
    
             \cdashlinelr{1-5}

             Convolutional & 128 & 20M & 0.8063789010047913 & 0.6470479816198349 \\

             \bottomrule\\\\\\
        \end{tabular}
        }
        \label{tab:style_size_abl}
    \end{minipage}

\end{figure}

\npnoround

\subsubsection{Rank and Layer choice}
As the modularity of our approach is a key differentiator from other methods, we perform an analysis of how the choice of adapted layers and the overall rank influence the performance in the context of style conditioning (Table \ref{tab:style_size_abl}). Overall the cross-attention layers perform best and show very good performance even with a very small rank. Convolutional and self-attention layers perform worse but may be better suited for other tasks such as conditioning on the style of an image instead of its semantics.

\setlength{\mycw}{0.15\textwidth}

\newcolumntype{P}[1]{>{\centering\arraybackslash}p{#1}}

\begin{table}[tb]
\centering
\scalebox{.8}{
\begin{tabular}{ c  P{\mycw} P{\mycw} P{\mycw}  P{\mycw} P{\mycw} P{\mycw} }

    \rotatebox{90}{\parbox{\mycw}{\centering Style}} &
    \includegraphics[width=\linewidth]{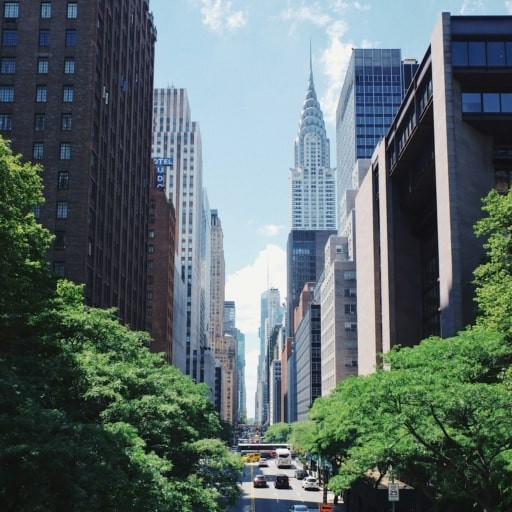} &
    \includegraphics[width=\linewidth]{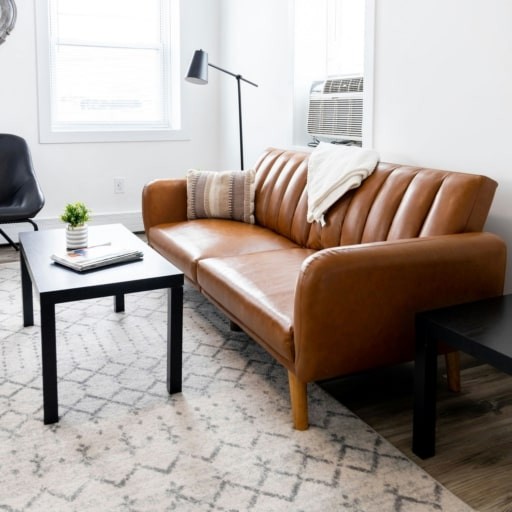} &
    \includegraphics[width=\linewidth]{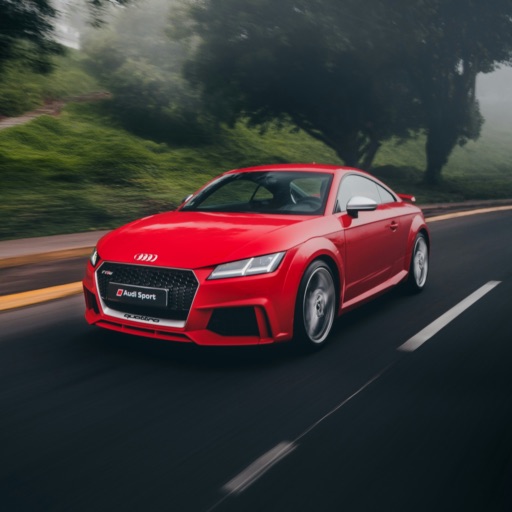} &
    \includegraphics[width=\linewidth]{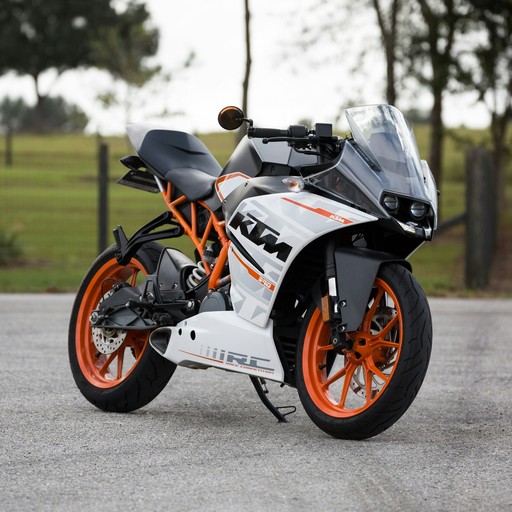} &
    \includegraphics[width=\linewidth]{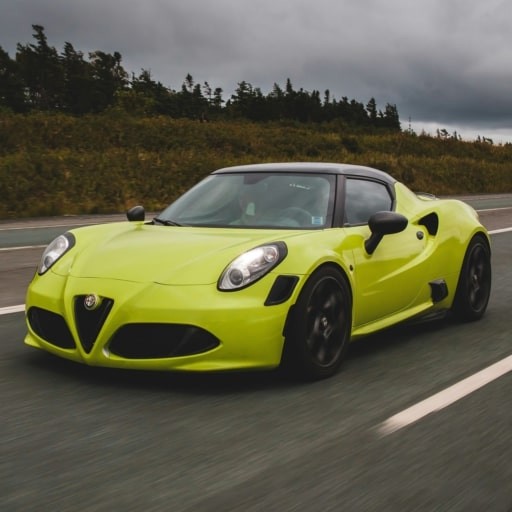} &
    \includegraphics[width=\linewidth]{images/joint/car3prompt.jpg} \\

     \rotatebox{90}{\parbox{\mycw}{\centering Structure}} &
    \includegraphics[width=\linewidth]{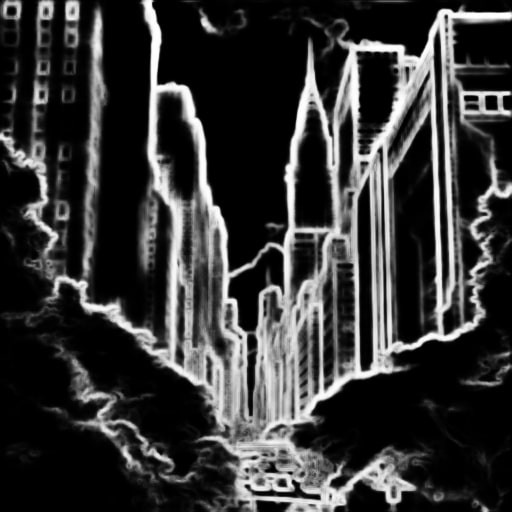} &
    \includegraphics[width=\linewidth]{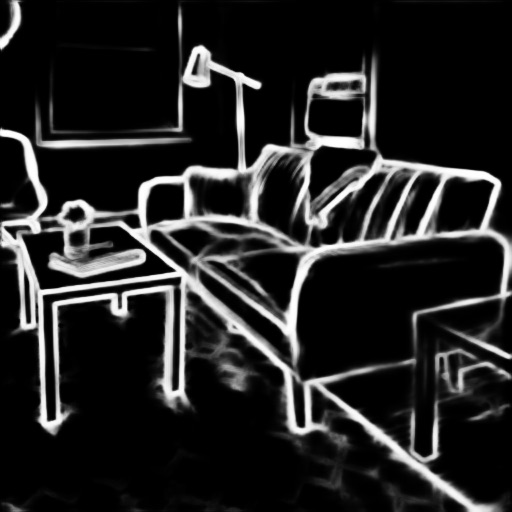} &
    \includegraphics[width=\linewidth]{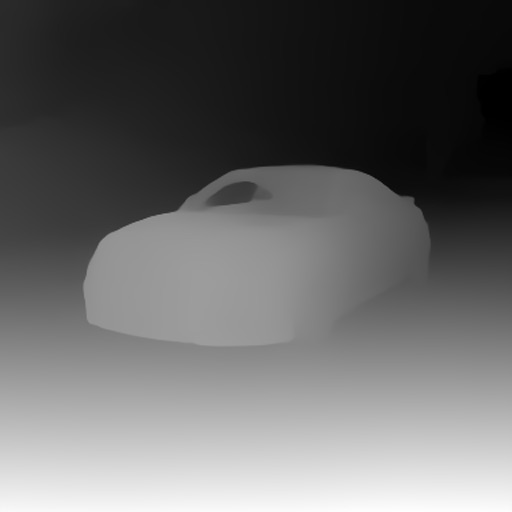} &
    \includegraphics[width=\linewidth]{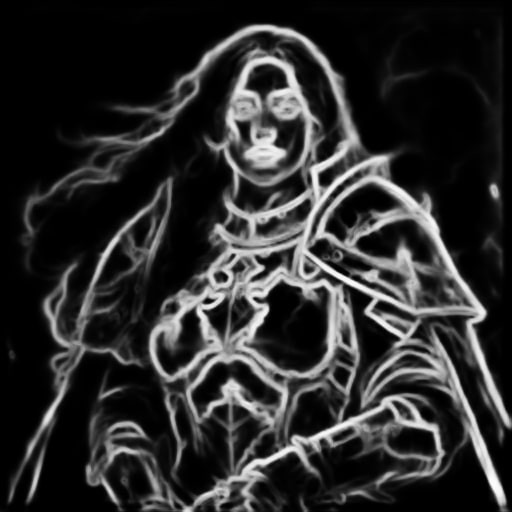} &
    \includegraphics[width=\linewidth]{images/joint/carcorrectprompt.jpeg} &
    \includegraphics[width=\linewidth]{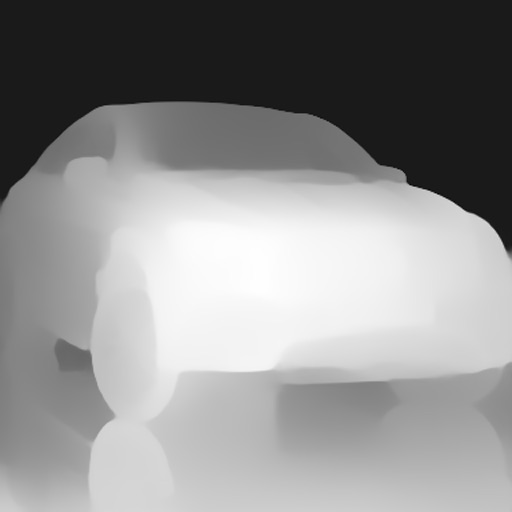} \\

    \rotatebox{90}{\parbox{\mycw}{\centering Sample}} &
    \includegraphics[width=\linewidth]{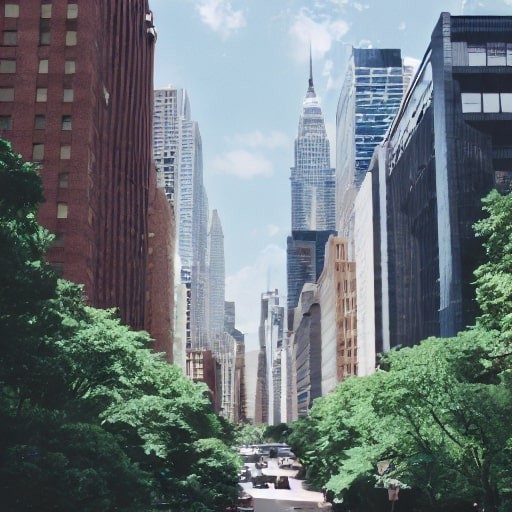} &
    \includegraphics[width=\linewidth]{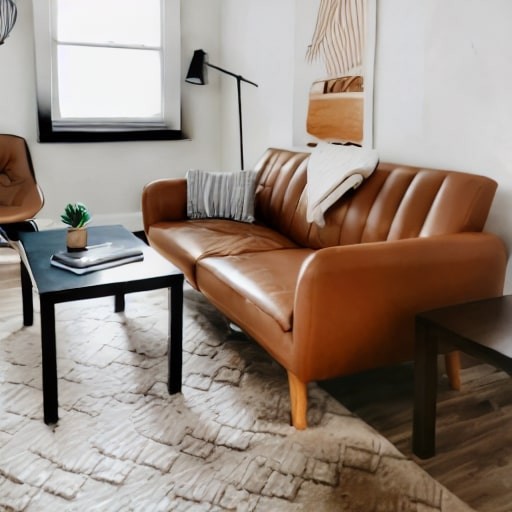} &
    \includegraphics[width=\linewidth]{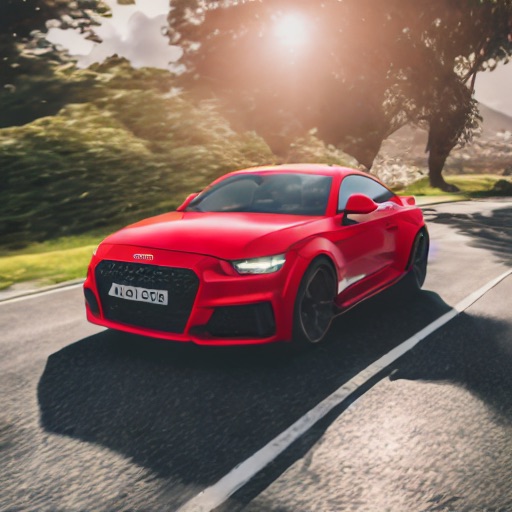} &
    \includegraphics[width=\linewidth]{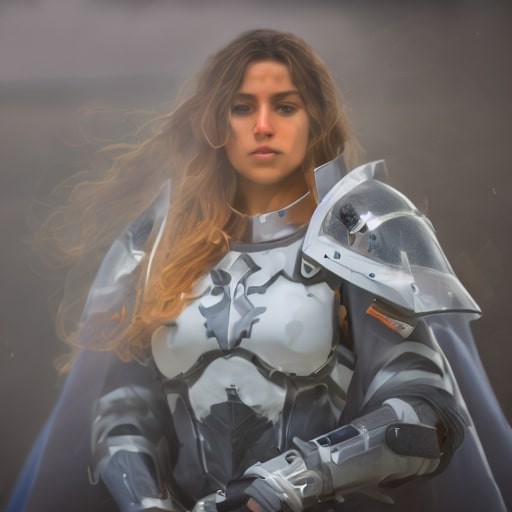} &
    \includegraphics[width=\linewidth]{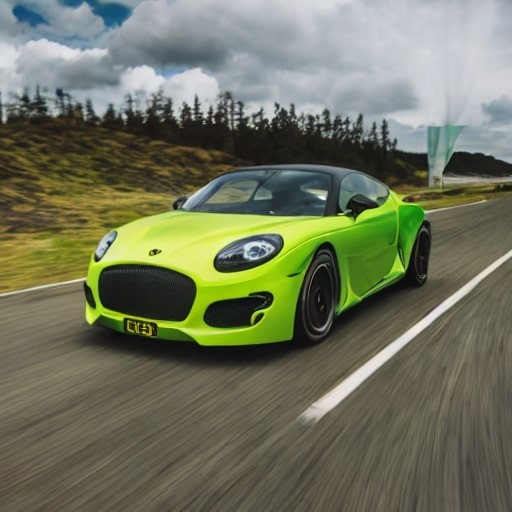} &
    \includegraphics[width=\linewidth]{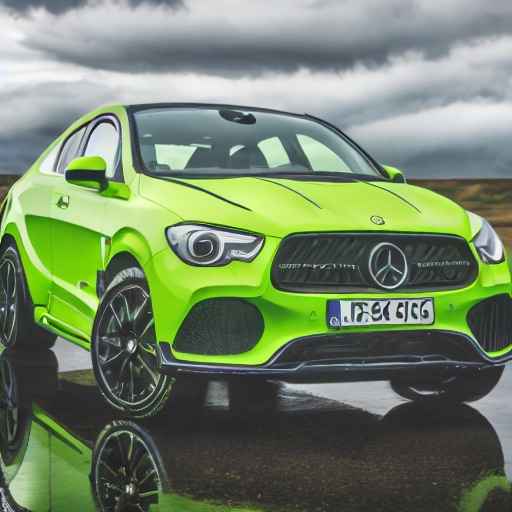} \\

\end{tabular}

}

\captionof{figure}{Examples of using a style and structure \methodname{} jointly. We show both reconstruction tasks (first, second, and third columns) and cases where the style and structure image do not match perfectly (fourth, fifth, and sixth columns).}

\label{fig:joint}
\end{table}

\subsubsection{Structure} 
\label{sec:ref-structure}
We adapt the convolutional layers and vary in what blocks we adapt them. Stable Diffusion 1.5 contains four upsampling and downsampling blocks and one middle block. Each block itself contains two ResNet \cite{resnet} blocks with two convolutional layers each. We analyze two different configurations: Configuration A only adapts the first layer in every block, totaling nine adapted convolutional layers. Configuration B adapts the first convolutional layer in every ResNet block, i.e. 17 adapted layers in total (Table \ref{tab:main_struct}). Finally, we show examples of jointly using a style and structure conditioning on Fig. \ref{fig:joint}. We refer the reader to the Appendix for additional results on prompt combinations.

\section{Conclusion}

In this paper we have introduced \methodname{}, an approach to condition the image generation process of foundational text-to-image models. \methodname{} is an powerful and efficient approach for conditioning that unifies \structure{} and \style{} control using a conditional LoRA architecture that enables zero-shot generalization. Our approach enables fine-grained control over both \structure{} and \style{} obtaining state-of-the-art results and outperforming recent approaches while also optimizing a more compact number of parameters. \methodname{} presents progress towards acquiring fine-grained control over text-to-image models in an effective manner.

\section*{Acknowledgments}
This project has been supported by the German Federal Ministry for Economic Affairs and Climate Action within the project ``NXT GEN AI METHODS – Generative Methoden für Perzeption, Prädiktion und Planung'', the German Research Foundation (DFG) project 421703927, and the bidt project KLIMA-MEMES. The authors gratefully acknowledge the Gauss Center for Supercomputing for providing compute through the NIC on JUWELS at JSC and the HPC resources supplied by the Erlangen National High Performance Computing Center (NHR@FAU funded by DFG).

Further, we would like to thank Micheal Neumayr for his help on the paper, and Owen Vincent for continuous technical support.

\bibliographystyle{splncs04}
\bibliography{main}

\cleardoublepage
\setcounter{page}{1}
\setcounter{linenumber}{1}
\setcounter{footnote}{1}

\setcounter{section}{6}
\setcounter{figure}{7}
\setcounter{table}{3}

\section{Appendix}
\subsection{Limitations \& Future Work}
While our \methodname{} is a general method to condition LoRAs and can be applied to any type of deep learning model, we have only shown its effectiveness in text-to-image diffusion models based on Stable Diffusion. Applying \methodname{} to fully transformer-based diffusion models such as DiT \cite{Li_2023_CVPR} or large language models remains an interesting future direction to further verify its model-agnostic capabilities.

\subsection{Impact Statement}
This work aims to improve the control over generated images using text-to-image (T2I) diffusion models by introducing an efficient and holistic method of adding additional conditioning to them. While previous methods enabled similar control over either the style or structure of the generated images, they were held back by requiring large auxiliary networks or being limited to only applying to one modality. As this work introduces a universal and more flexible approach to introducing additional conditioning that is not inherently limited to only one particular network architecture, it aids in making highly specific control over the generation process more accessible. This extension of the capabilities of T2I diffusion models, similar to progress in improving base models, carries the risk of further enabling the generation of more believable disinformation or harmful content.

\subsection{Style and Structure Conditioning}
\label{appendix-a}

\paragraph{Joint Conditioning} 
In Figure \ref{fig:joint} of the main paper, we show examples of joint conditioning where we use our \methodname{} for both style and structure conditioning. The first three columns show a reconstruction task with perfectly matching style and structure conditioning. The resulting samples (third row) are very close to the original images (top row), indicating the superb composability of multiple LoRAdapters. The last three columns show samples where the style and structure conditioning do not match. In the fifth column, we show the combination of the depth of the red car with the style of the green car: Not only does the color of the sample match the style image but the design of the green car is successfully applied, e.g. the headlights are round as opposed to the more edgy headlights of the red car. The background is also transferred correctly. 
 
To further evaluate the performance of joint conditioning, we conducted additional experiments with the style and structure conditioning coming from different classes (Figure \ref{fig:joint-more}). Even though the classes are completely different, \methodname{} can still sensibly combine the two conditioning.

Lastly, we investigate the effects of using a much larger adapter such as ControlNet to provide structure guidance (Figure \ref{fig:hed_cn_comp}) instead of our efficient LoRA-based adapter. We either use our method for both style and HED conditioning or just our method for style conditioning and ControlNet for HED conditioning. In either case, the adapters were trained separately. ControlNet introduces more variance to the generated samples in this reconstruction task, due to its large size (361M parameters) and tendency to interpret the structure conditioning, which results in hallucinations. On the other hand, our \methodname{} results in far less variance due to its smaller size and because each LoRA operates in its own subspace.

\setlength{\mycw}{0.18\textwidth}

\newcolumntype{P}[1]{>{\centering\arraybackslash}p{#1}}

\begin{table}[tb]
\centering
\scalebox{1}{
\begin{tabular}{ P{\mycw} P{\mycw} P{\mycw}  P{\mycw} P{\mycw}  }

    \vspace{0pt}\diagbox[width=\mycw, height=\mycw]{\textbf{Structure}}{\textbf{Style}} &
    \vspace{0pt}\includegraphics[width=\linewidth]{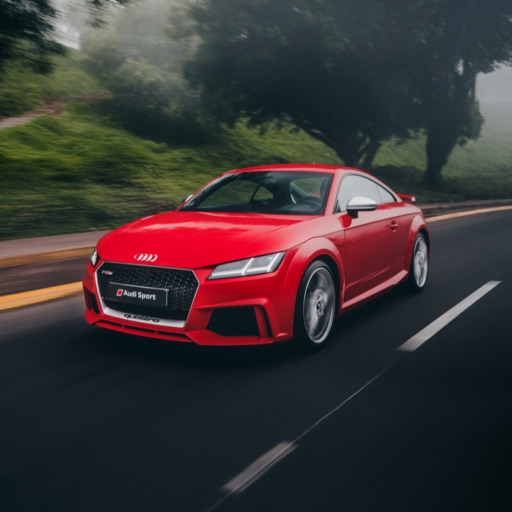} &
    \vspace{0pt}\includegraphics[width=\linewidth]{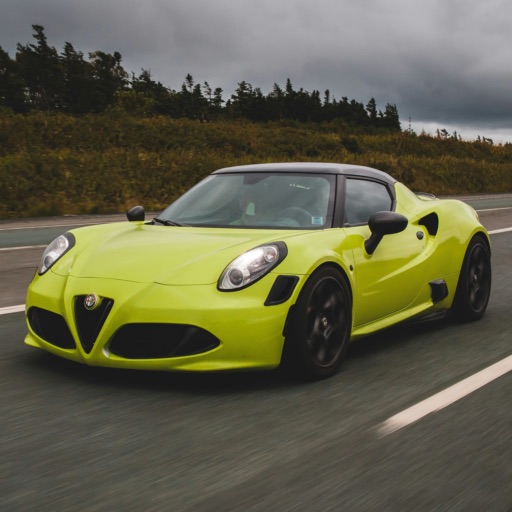} &
    \vspace{0pt}\includegraphics[width=\linewidth]{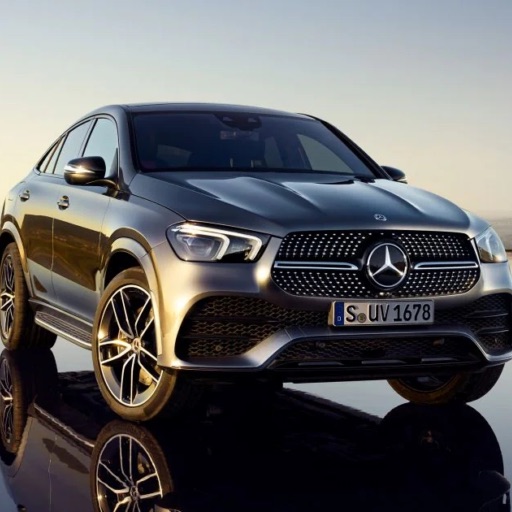} &
    \vspace{0pt}\includegraphics[width=\linewidth]{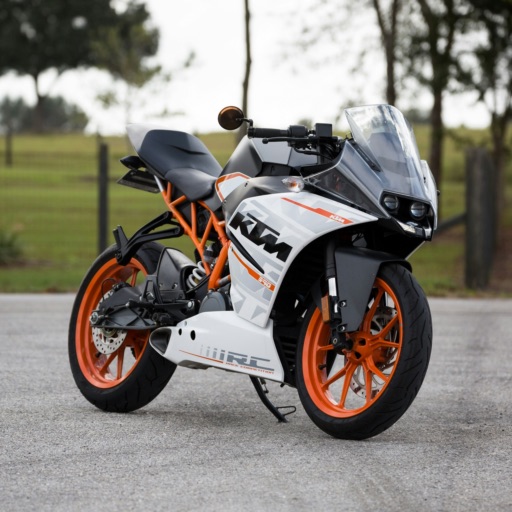} \\

    \includegraphics[width=\linewidth]{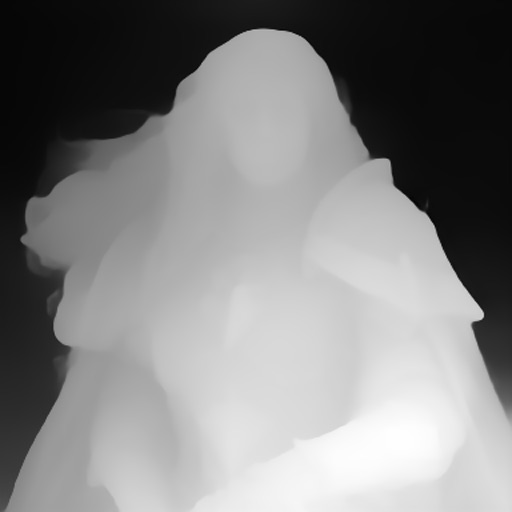} &
    \includegraphics[width=\linewidth]{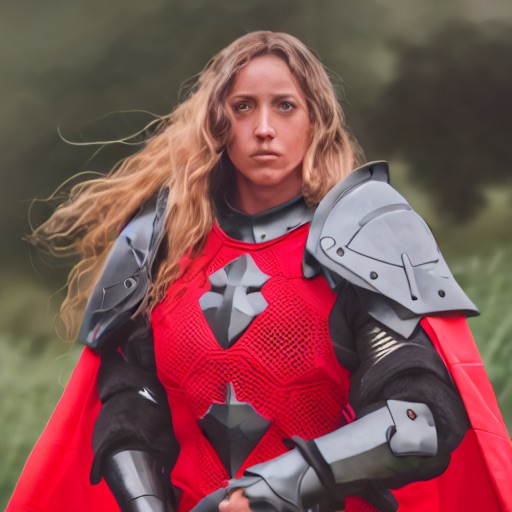} &
    \includegraphics[width=\linewidth]{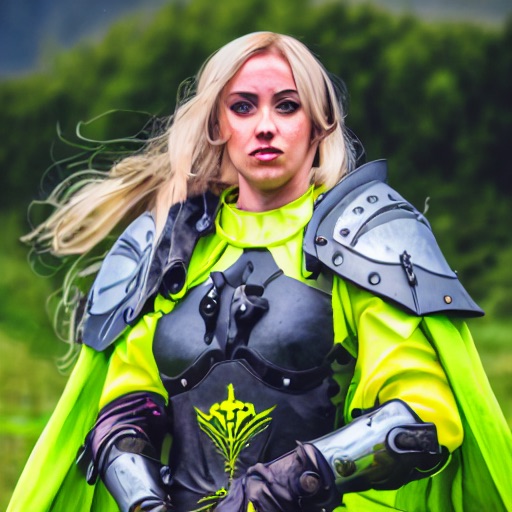} &
    \includegraphics[width=\linewidth]{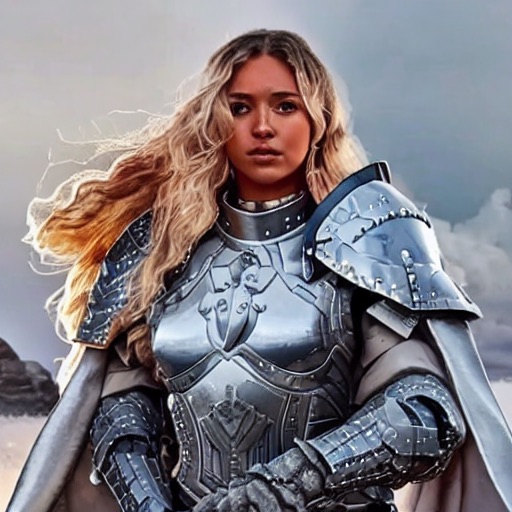} &
    \includegraphics[width=\linewidth]{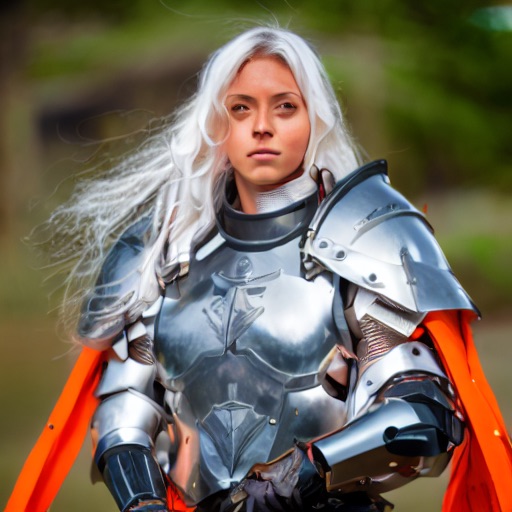} \\

    \includegraphics[width=\linewidth]{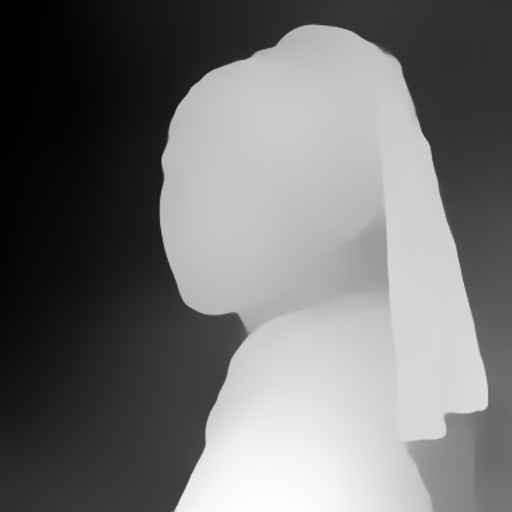} &
    \includegraphics[width=\linewidth]{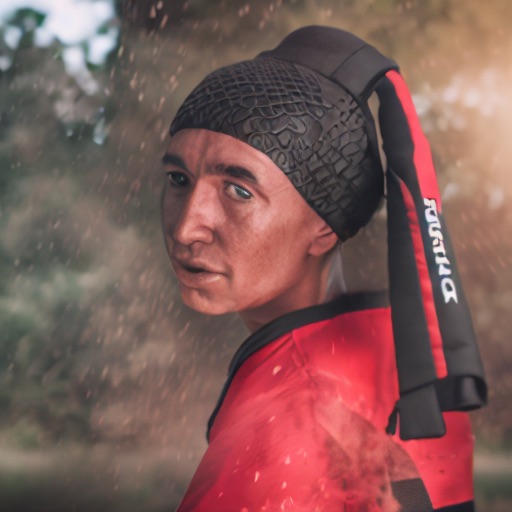} &
    \includegraphics[width=\linewidth]{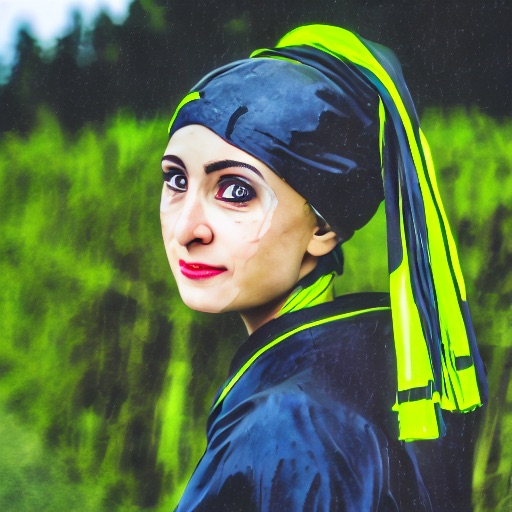} &
    \includegraphics[width=\linewidth]{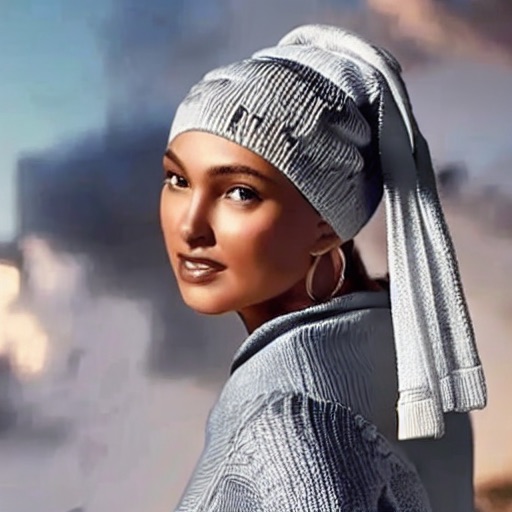} &
    \includegraphics[width=\linewidth]{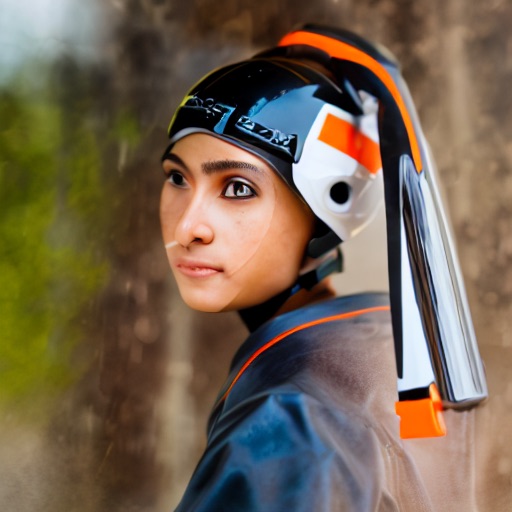} \\

\end{tabular}

}

\captionof{figure}{Samples generated using two LoRAdapters jointly for style and structure conditioning. The structure conditioning is provided as a depth map. Note how style and structure come from two different classes: persons and vehicles. Still \methodname{} can combine the two conditionings and generate coherent samples.}

\label{fig:joint-more}
\end{table}

\paragraph{Style Conditioning}
For style conditioning, we qualitatively compare our method against various other approaches (Figure \ref{fig:style_comp}). This includes models trained from scratch such as Kandinsky \cite{razzhigaev2023kandinsky} and Versatile Diffusion \cite{xu2022versatile}, fully fine-tuned models like Stable Diffusion unCLIP \cite{rombach2022high}, and other adapter approaches like Uni-ControlNet \cite{zhao2023uni} and IP-Adapter \cite{ye2023ip-adapter}. \methodname{} compares favorably even against models that were trained from scratch or fully fine-tuned. Additionally, we keep the original text conditioning, which is an advantage over adapters such as SeeCoder \cite{xu2023prompt} that replace the original text conditioning.

We also ablate over the choice of which layers to adapt using our smaller CLIP ViT-L/14 based model with only 13M parameters and a rank of 128 (Figure \ref{fig:layer_comp}). Quantitative findings are in Table \ref{tab:main_style} of the main paper. Overall, we find that adapting the cross-attention layers yields the highest fidelity and similarity to the prompt image with self-attention layers being a close second. Another advantage of adapting the cross-attention layers is that it enables excellent composability with the text prompt (Figure \ref{fig:prompt}). We can add text prompts referring to the subject in the style image and the model realistically alters the subject according to the semantic context of the image. For instance, note the prompt "wearing a hat" resulting in a crown for the female warrior and a loose summer hat for the girl standing in the garden.

\paragraph{Structure Conditioning}
Next to the quantitative evaluation for depth (Table \ref{tab:main_struct}), we also evaluate two model configurations trained on HED maps. As described in section \ref{sec:ref-structure}, configuration A adapts only the first convolutional layer in every down or upsampling block whereas configuration B adapts the first convolutional layer in every ResNet block. Configuration B yields the better performance but even the much smaller A configuration manages to outperform existing approaches

Lastly, we also condition one model on key poses \cite{cao2017realtime} as a modality that is quite different from HED or depth maps (Figure \ref{fig:poses}). We show multiple poses for different text prompts which \methodname{} can combine successfully, showing that it can also work with more sparse representations.

\begin{table}[tb]
    \centering
    \caption{Comparison of a smaller and a larger version of our \methodname{} for HED conditioning. Even the smaller model (A) outperforms the leading approaches ControlNet and Uni-ControlNet.}
    \begin{tabular}{l c c c c}
        \toprule

        & \multicolumn{4}{c}{HED} \\
        Model & Params & SSIM $\uparrow$ & FID $\downarrow$ & LPIPS $\downarrow$ \\

        \midrule
        
        ControlNet  & 361M & 0.555 & 20.646 & 0.553  \\
        Uni-ControlNet & 361M & 0.601 & 17.530 & 0.530 \\ 
        T2I-Adapter & {\color{white}0}39M & - & - & - \\

         \methodname{}-A (Ours) & {\color{white}0}\textbf{17M}  & \textbf{0.621} & \textbf{14.584}  & \textbf{0.489}\\

        \methodname{}-B (Ours) & {\color{white}0}\textbf{32M}  & \textbf{0.644} & \textbf{14.761}  & \textbf{0.475}\\

         \bottomrule
    \end{tabular}
    \label{tab:struct_appendix}
\end{table}

\setlength{\mycw}{0.2\textwidth}

\newcolumntype{P}[1]{>{\centering\arraybackslash}p{#1}}

\begin{table}[tb]
\centering
\scalebox{1}{
\begin{tabular}{ P{\mycw} P{\mycw} P{\mycw}  P{\mycw} P{\mycw}  }

    \vspace{0pt}\diagbox[width=\mycw, height=\mycw]{\textbf{Prompt}}{\textbf{Structure}} &
    \vspace{0pt}\includegraphics[width=\linewidth]{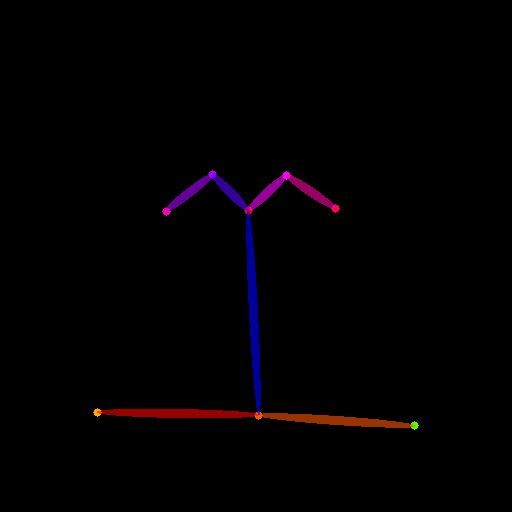} &
    \vspace{0pt}\includegraphics[width=\linewidth]{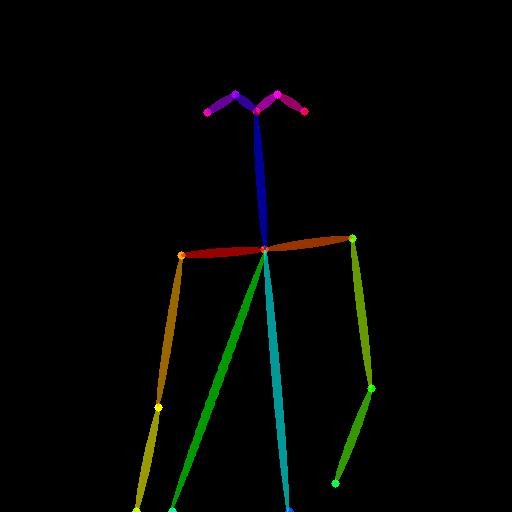} &
    \vspace{0pt}\includegraphics[width=\linewidth]{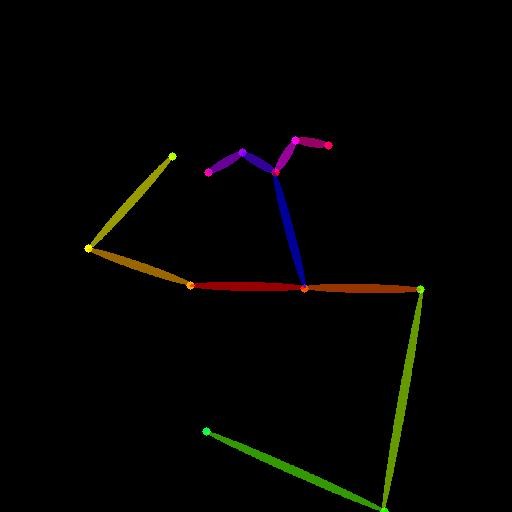} \\

    \vspace{-8pt}
    \begin{minipage}[c][1\mycw][c]{1\mycw}
      \centering
      "A man in black suit"
    \end{minipage} \vspace{0pt} &
    \vspace{-8pt}\includegraphics[width=\linewidth]{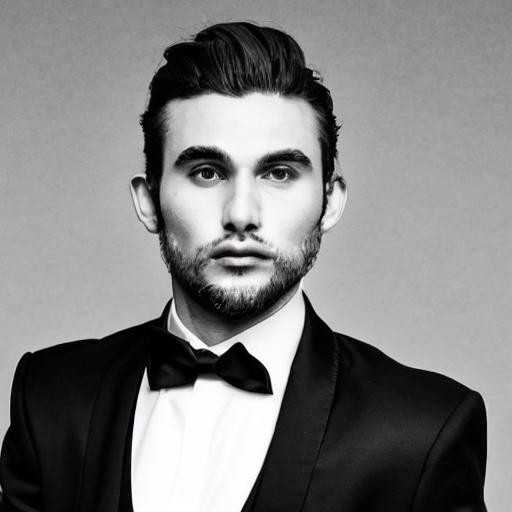} &
    \vspace{-8pt}\includegraphics[width=\linewidth]{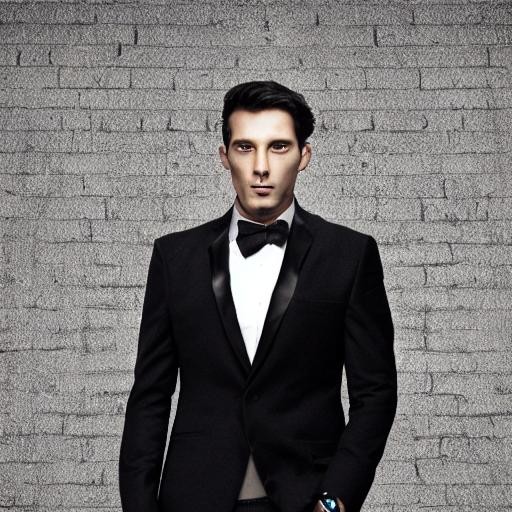} &
    \vspace{-8pt}\includegraphics[width=\linewidth]{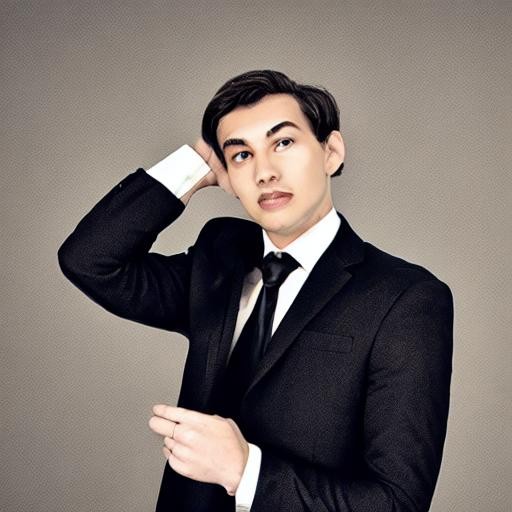} \\

   \vspace{-8pt}
    \begin{minipage}[c][1\mycw][c]{1\mycw}
      \centering
      "A woman in a red shirt"
    \end{minipage} \vspace{0pt} &
    \vspace{-9pt}\includegraphics[width=\linewidth]{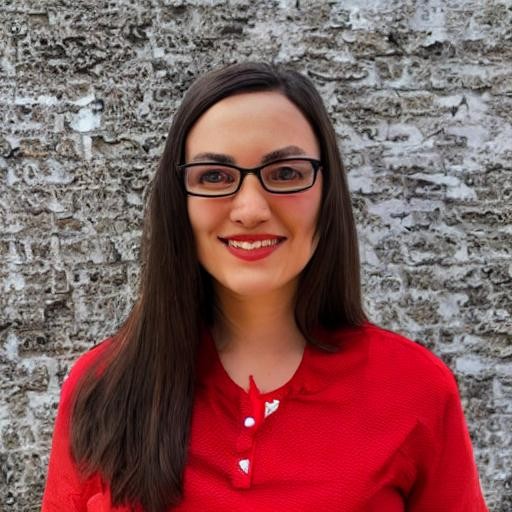} &
    \vspace{-9pt}\includegraphics[width=\linewidth]{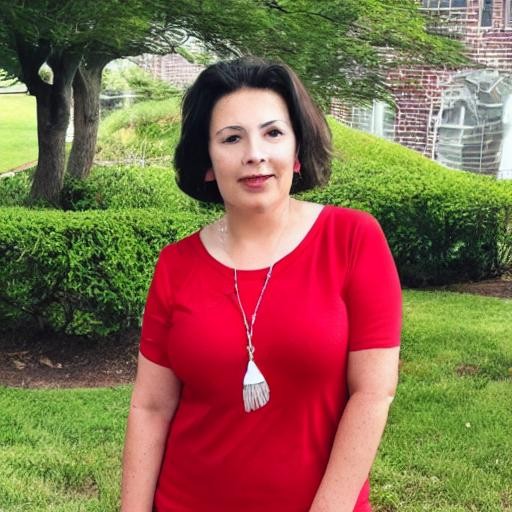} &
    \vspace{-9pt}\includegraphics[width=\linewidth]{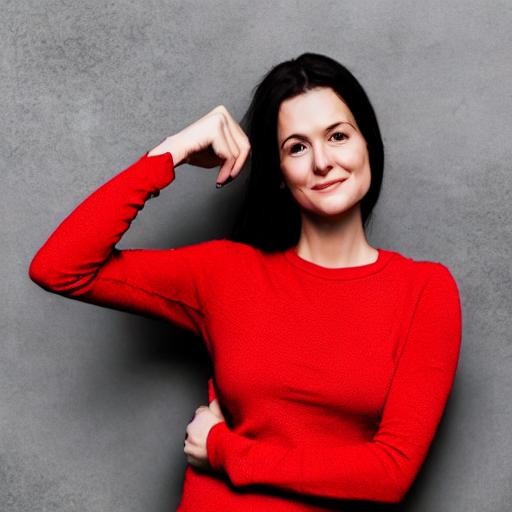} \\

\end{tabular}

}

\captionof{figure}{\methodname{} trained on human key poses using configuration B. Note how the samples adhere to both structure and text conditioning. }

\label{fig:poses}
\end{table}

\setlength{\mycw}{0.1\textwidth}

\begin{table*}[tb]
\centering
\begin{tabular}{  p{\mycw} p{\mycw} p{\mycw} p{\mycw} p{\mycw} p{\mycw} p{\mycw} p{\mycw} p{\mycw} }
    
    \multicolumn{9}{c}{Style Prompt} \\
    \includegraphics[width=\linewidth]{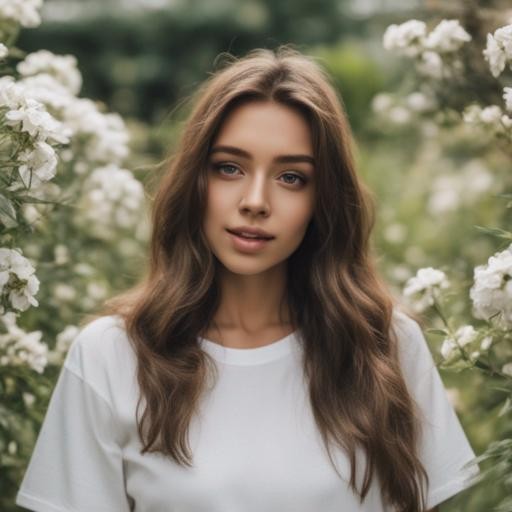}  &
    \includegraphics[width=\linewidth]{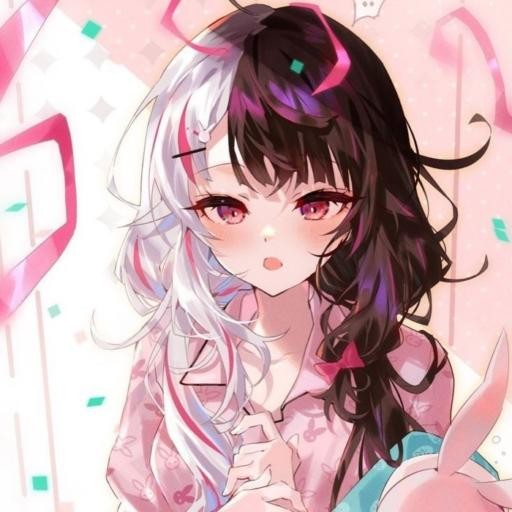}  &
    \includegraphics[width=\linewidth]{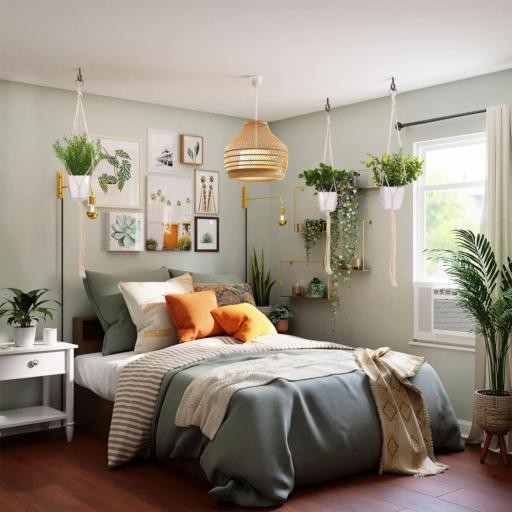}  &
    \includegraphics[width=\linewidth]{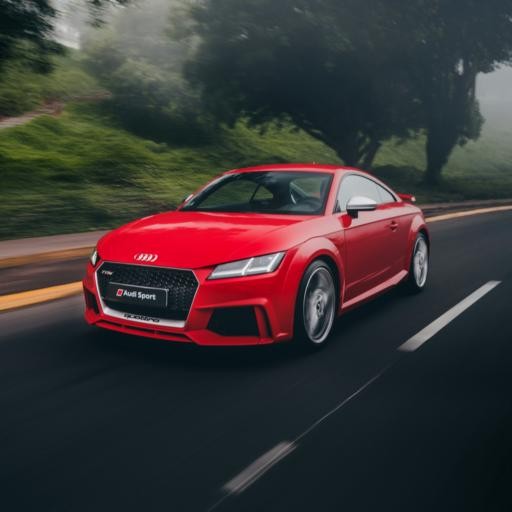}  &
    \includegraphics[width=\linewidth]{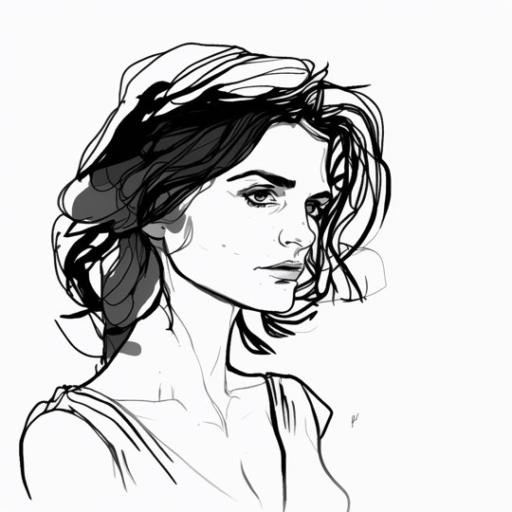}  &
    \includegraphics[width=\linewidth]{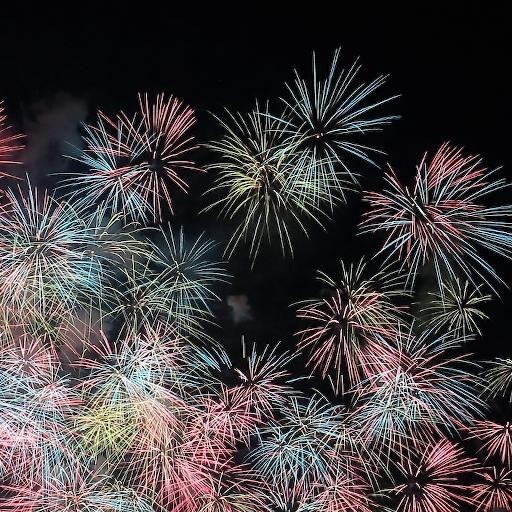}  &
    \includegraphics[width=\linewidth]{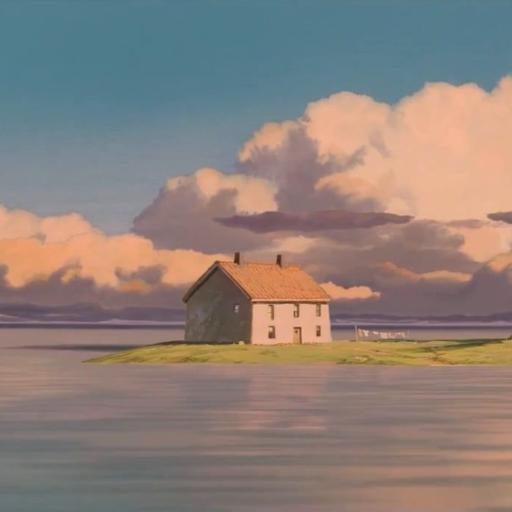}  &
    \includegraphics[width=\linewidth]{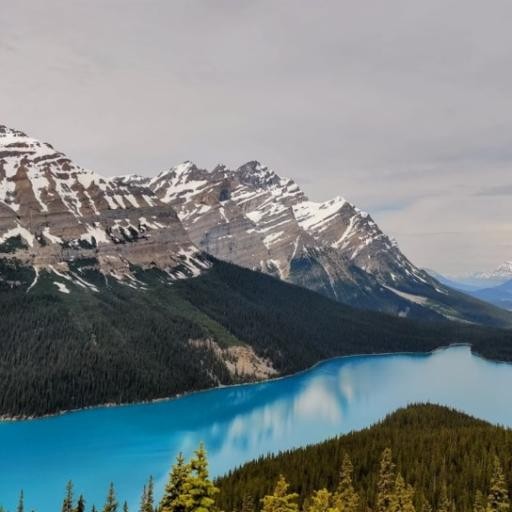}  &
    \includegraphics[width=\linewidth]{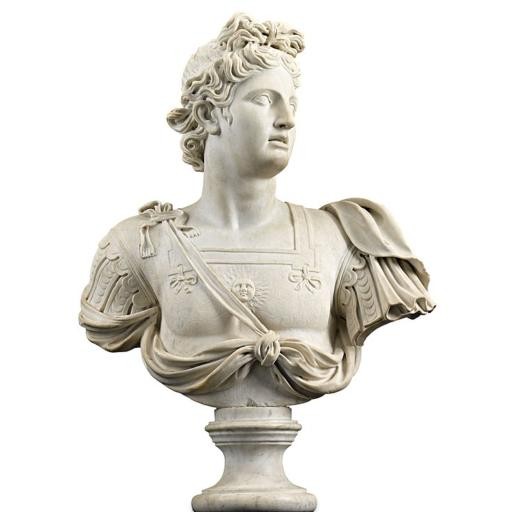}  \\

    \multicolumn{9}{c}{Cross-Attention} \\
    \includegraphics[width=\linewidth]{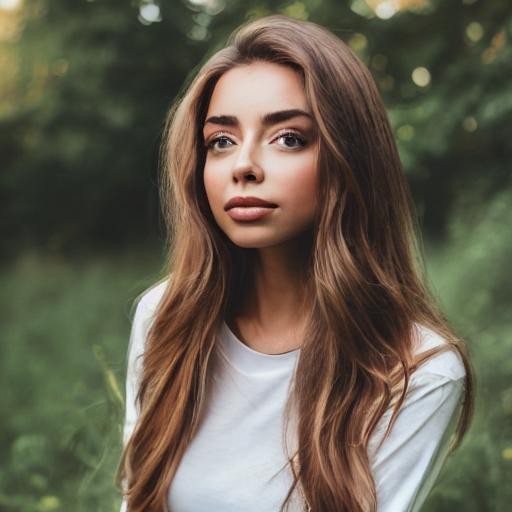}  &
    \includegraphics[width=\linewidth]{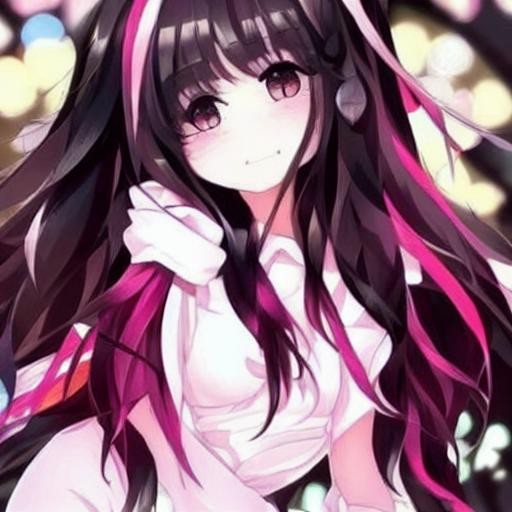}  &
    \includegraphics[width=\linewidth]{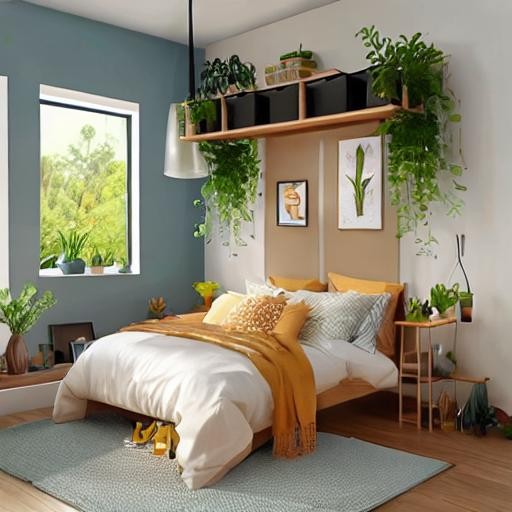}  &
    \includegraphics[width=\linewidth]{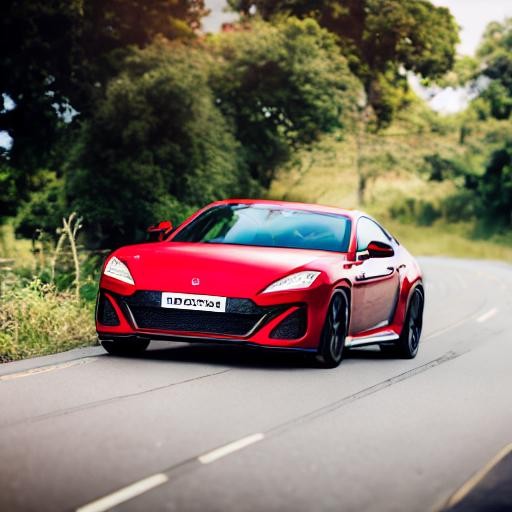}  &
    \includegraphics[width=\linewidth]{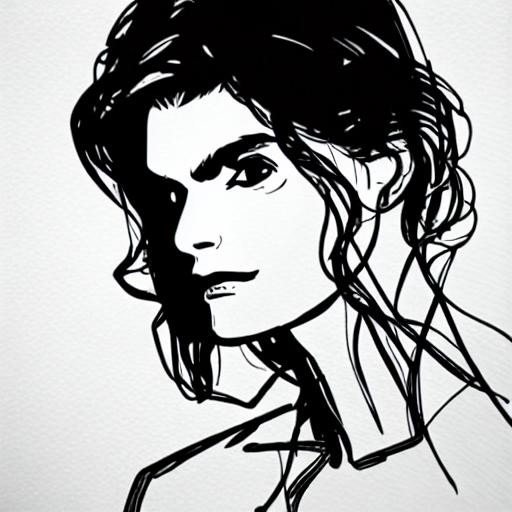}  &
    \includegraphics[width=\linewidth]{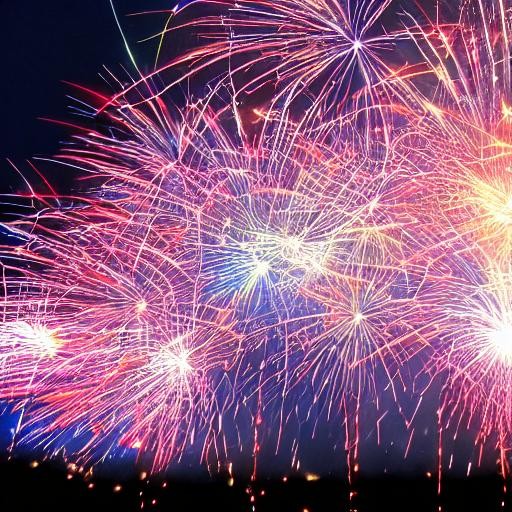}  &
    \includegraphics[width=\linewidth]{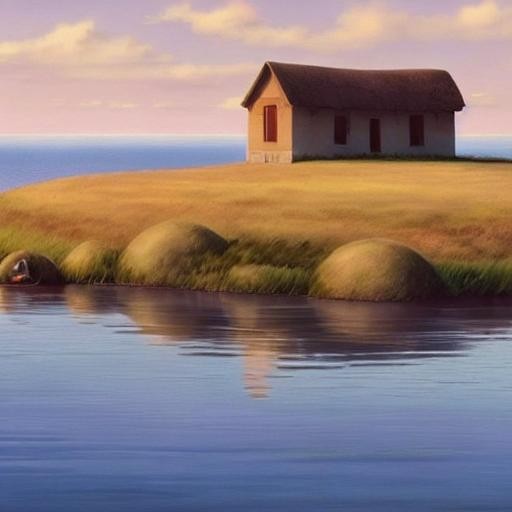}  &
    \includegraphics[width=\linewidth]{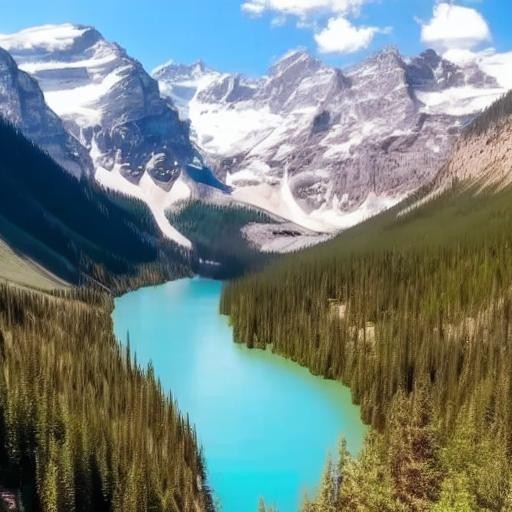}  &
    \includegraphics[width=\linewidth]{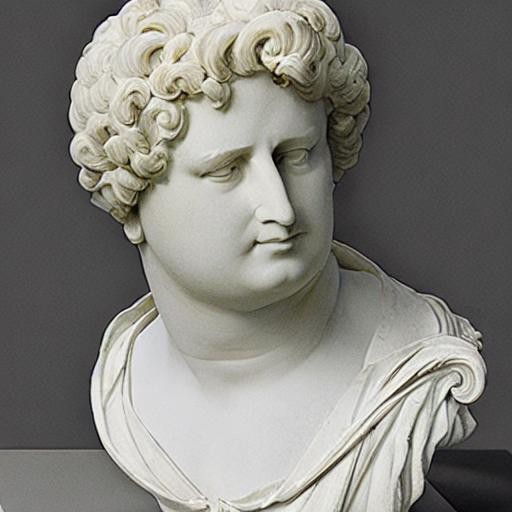}  \\

    \multicolumn{9}{c}{Self-Attention} \\
    \includegraphics[width=\linewidth]{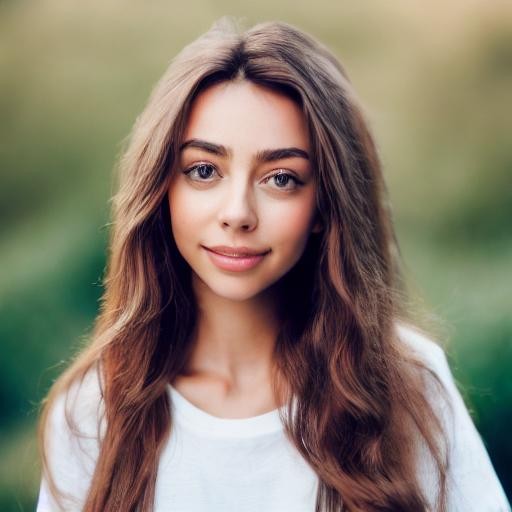}  &
    \includegraphics[width=\linewidth]{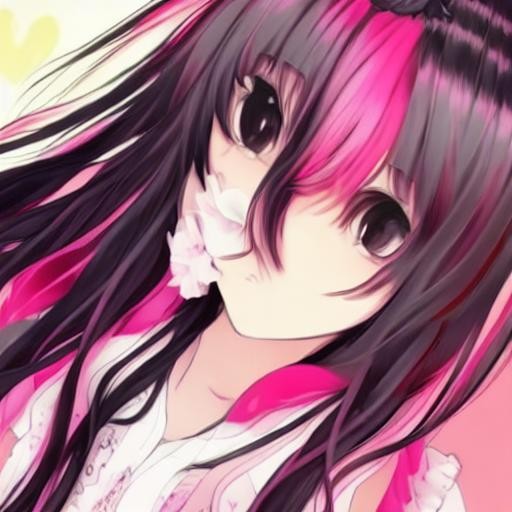}  &
    \includegraphics[width=\linewidth]{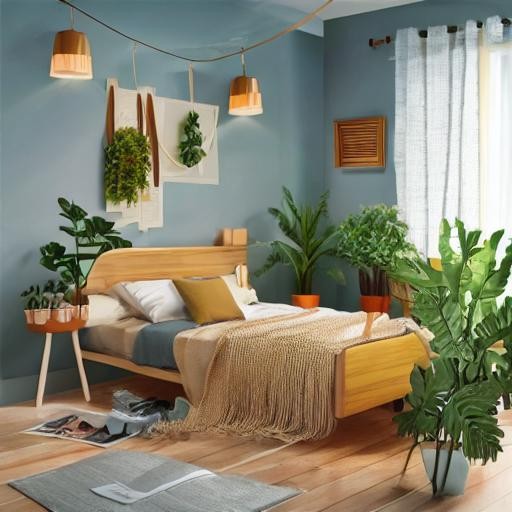}  &
    \includegraphics[width=\linewidth]{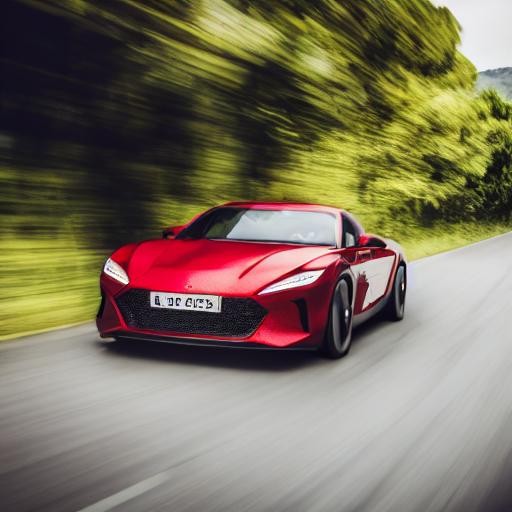}  &
    \includegraphics[width=\linewidth]{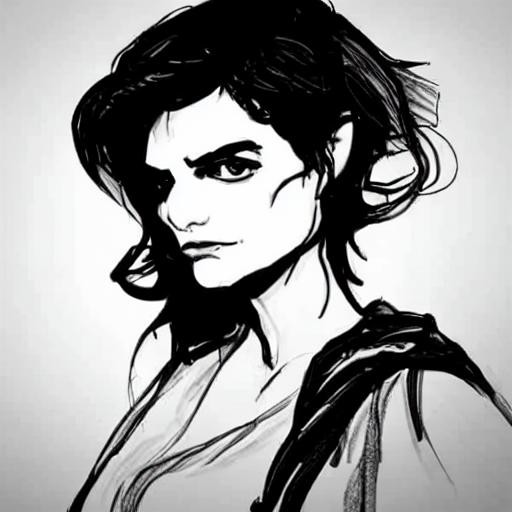}  &
    \includegraphics[width=\linewidth]{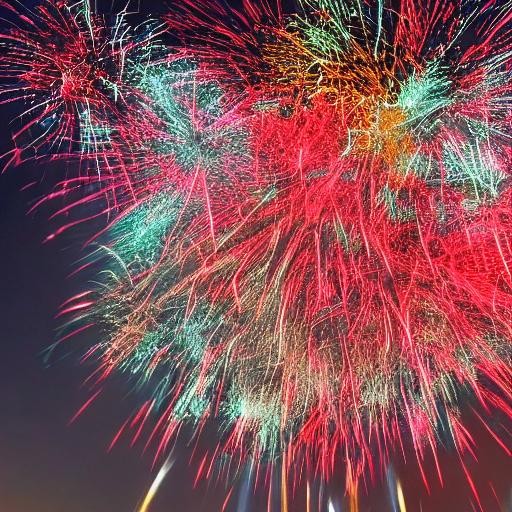}  &
    \includegraphics[width=\linewidth]{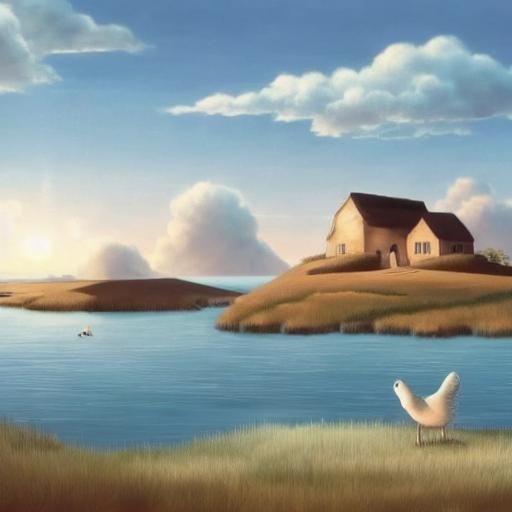}  &
    \includegraphics[width=\linewidth]{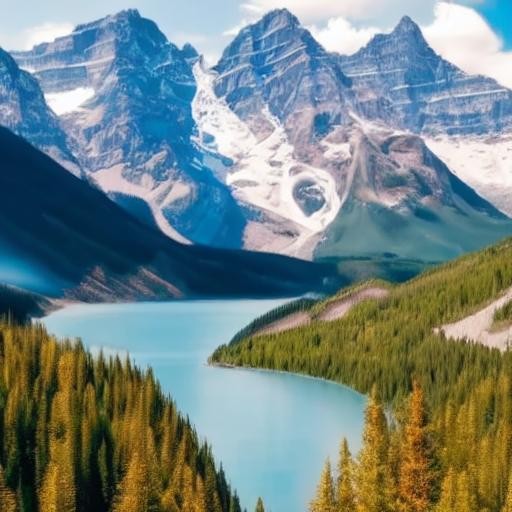}  &
    \includegraphics[width=\linewidth]{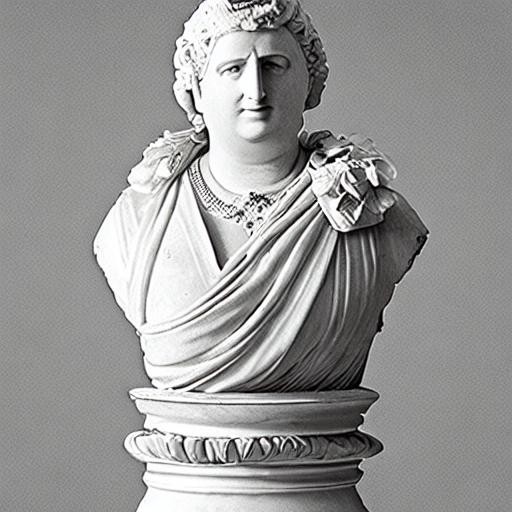}  \\

    \multicolumn{9}{c}{Convolutional} \\
    \includegraphics[width=\linewidth]{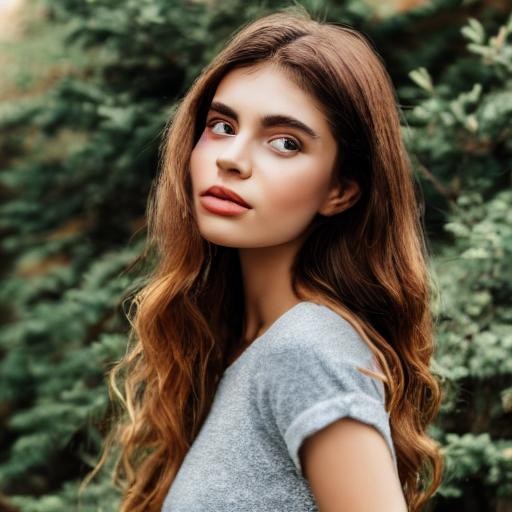}  &
    \includegraphics[width=\linewidth]{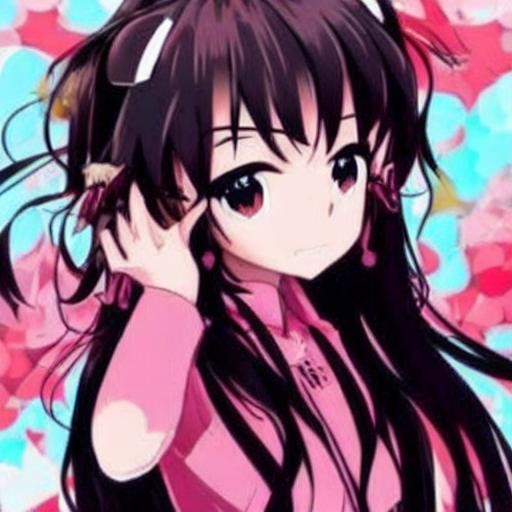}  &
    \includegraphics[width=\linewidth]{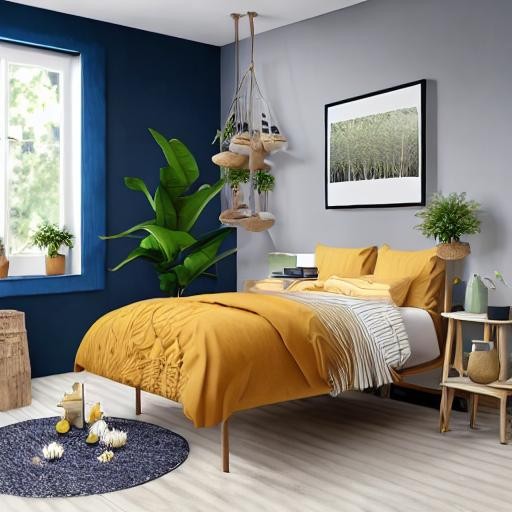}  &
    \includegraphics[width=\linewidth]{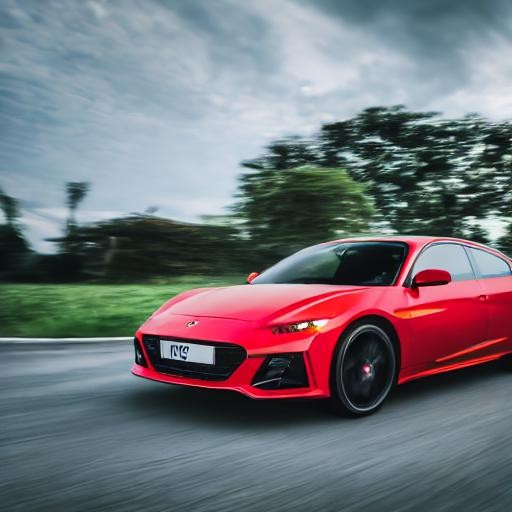}  &
    \includegraphics[width=\linewidth]{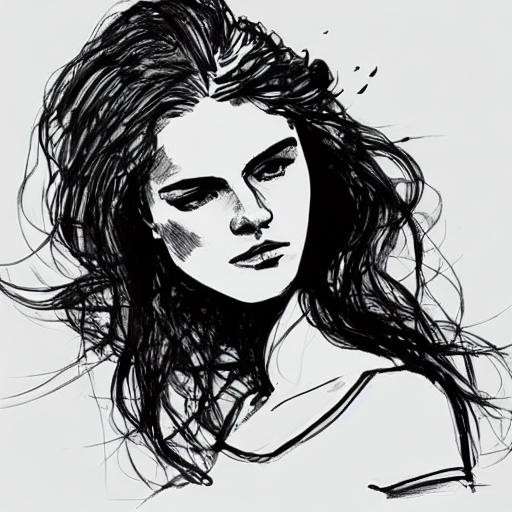}  &
    \includegraphics[width=\linewidth]{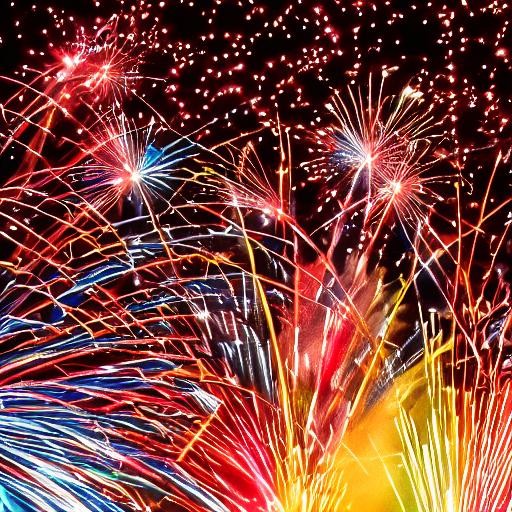}  &
    \includegraphics[width=\linewidth]{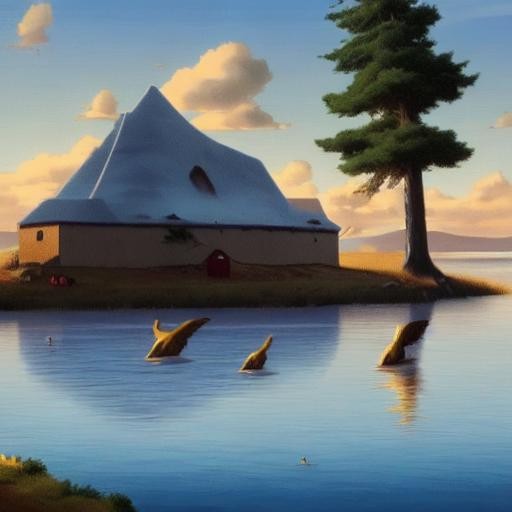}  &
    \includegraphics[width=\linewidth]{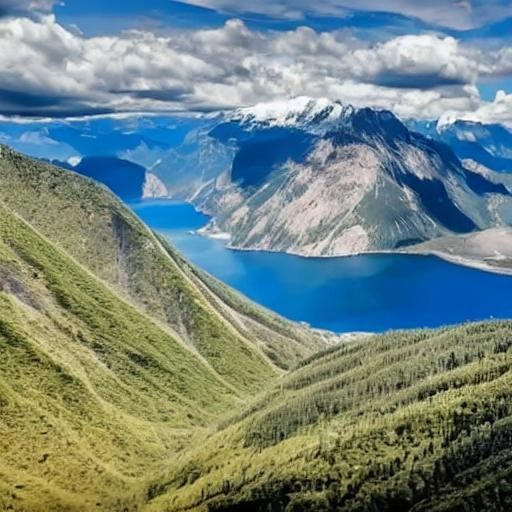}  &
    \includegraphics[width=\linewidth]{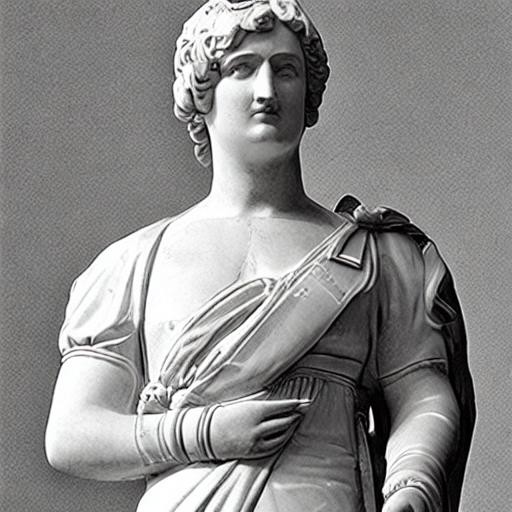}  \\

\end{tabular}

\captionof{figure}{Results of our smaller style \methodname{} for different layers. We only condition on the image and use an empty text prompt. Adapting the cross-attention layers has the highest fidelity and yields the best performance. 
}

\label{fig:layer_comp}
\end{table*}

\setlength{\mycw}{0.18\textwidth}

\begin{table}[tb]
\centering
\begin{tabular}{ c p{\mycw} p{\mycw} p{\mycw} p{\mycw} p{\mycw} }

    & \centering Structure & \centering \textbf{Style (Ours)} & \multicolumn{3}{c}{Samples} \\

    \midrule

    \rotatebox{90}{\parbox{2\mycw}{\centering \textbf{Ours} }} &
    \includegraphics[width=1\mycw]{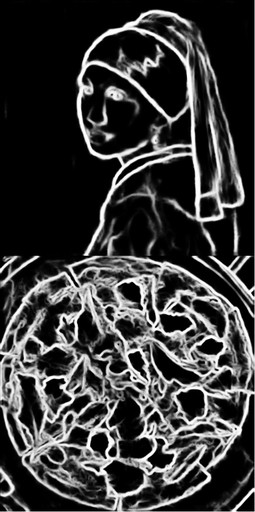} &
     &
    \multicolumn{3}{c}{
        \includegraphics[width=3\mycw]{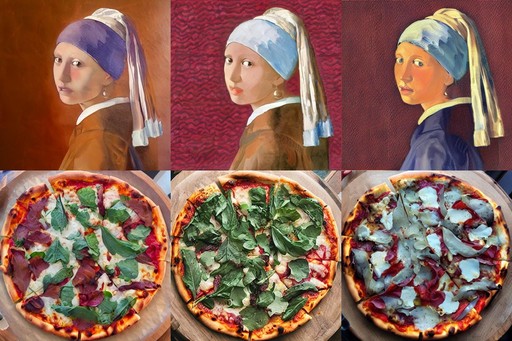}
    } \\

    \rotatebox{90}{\parbox{2\mycw}{\centering \textbf{Ours}}} &

     \multicolumn{2}{c}{
        \includegraphics[width=2\mycw]{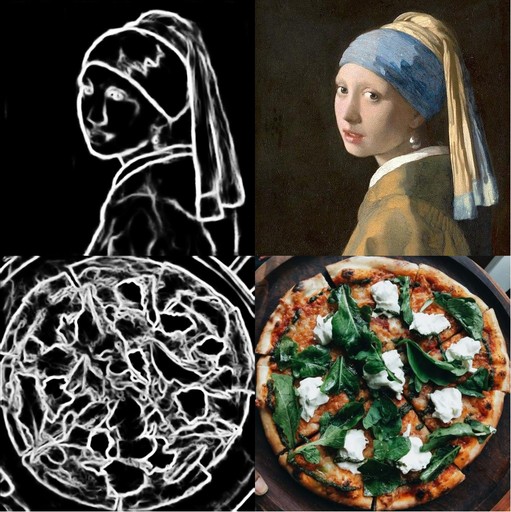}
     }
      &
    \multicolumn{3}{c}{
        \includegraphics[width=3\mycw]{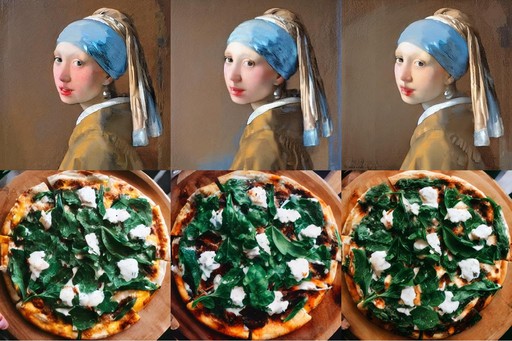}
    } \\

    \rotatebox{90}{\parbox{2\mycw}{\centering ControlNet}} &

     \multicolumn{2}{c}{
        \includegraphics[width=2\mycw]{images/appendix/hed_comp/both_prompt.jpg}
     }
     &
    \multicolumn{3}{c}{
        \includegraphics[width=3\mycw]{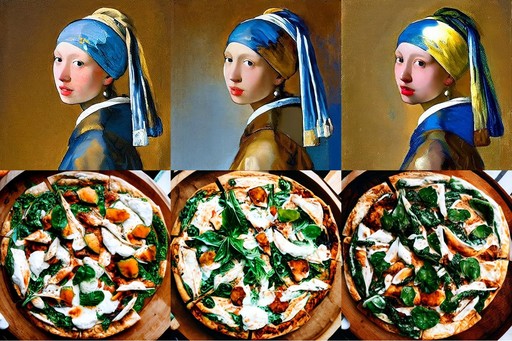}
    }  \\

\end{tabular}

\captionof{figure}{Analyses of interference between style and structure conditioning for \methodname{} and ControlNet. The first row shows HED conditioning using our method. The expressed variance in the sample is solely from the diffusion model (no style conditioning). In the second row, we add style conditioning using our method. Notice the drastic variance reduction between samples, showing how well a combination of multiple modalities using our method works due to each LoRA operating in its own subspace. The third row replaces our structure conditioning with ControlNet. The samples show increased variance, likely due to the large size of ControlNet, even though it uses the same style conditioning.}

\label{fig:hed_cn_comp}
\end{table}

\clearpage

\input{figures/style_qualitative_iptable}

\clearpage

\setlength{\mycw}{0.15\textwidth}

\begin{table*}[tb]
\centering
\begin{tabular}{  p{\mycw} p{\mycw} p{\mycw} p{\mycw} }
    
    \multicolumn{4}{c}{Image Prompt} \\
    \includegraphics[width=\linewidth]{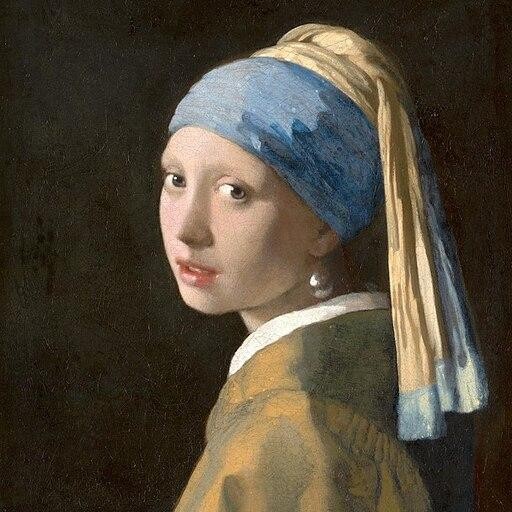}  &
    \includegraphics[width=\linewidth]{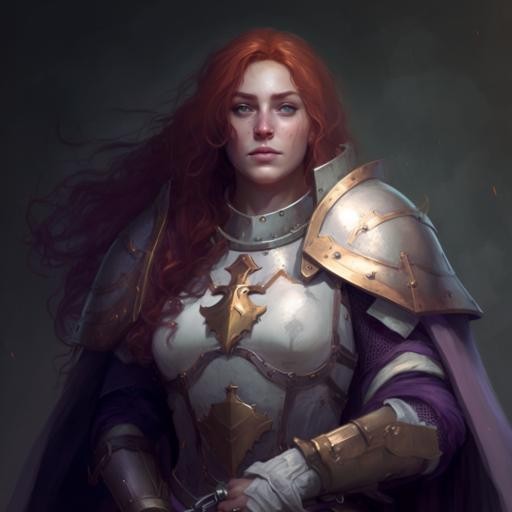}  &
    \includegraphics[width=\linewidth]{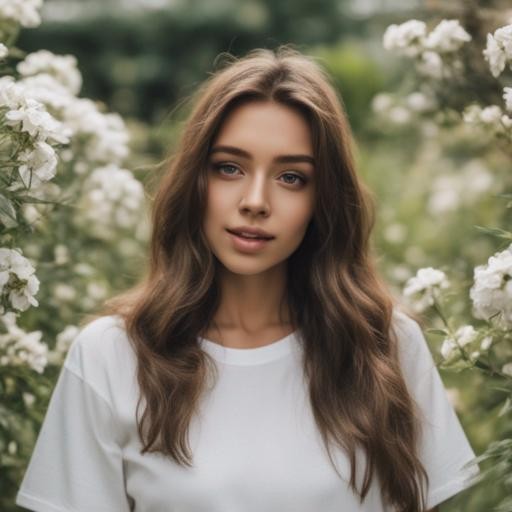}  &
    \includegraphics[width=\linewidth]{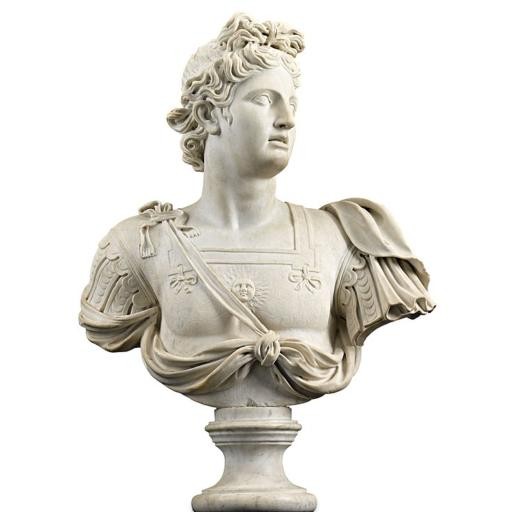}  \\

    \multicolumn{4}{c}{"in heavy plate armor"} \\
    \includegraphics[width=\linewidth]{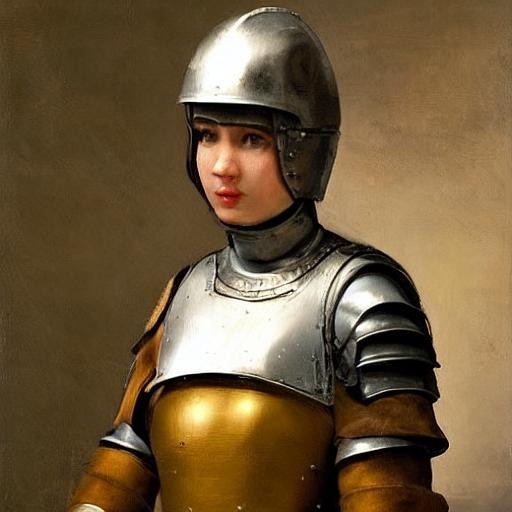}  &
    \includegraphics[width=\linewidth]{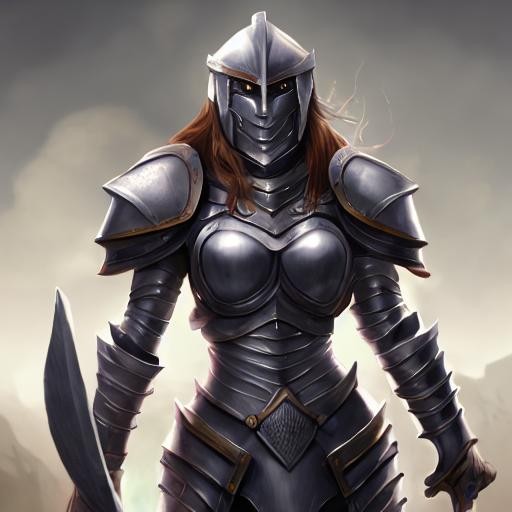}  &
    \includegraphics[width=\linewidth]{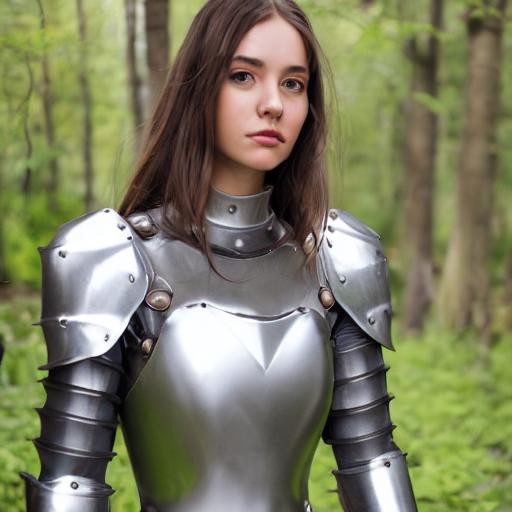}  &
    \includegraphics[width=\linewidth]{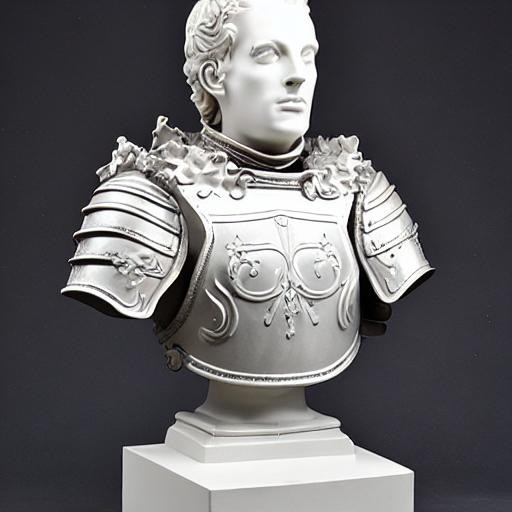}  \\

    \multicolumn{4}{c}{"wearing a yellow shirt"} \\
    \includegraphics[width=\linewidth]{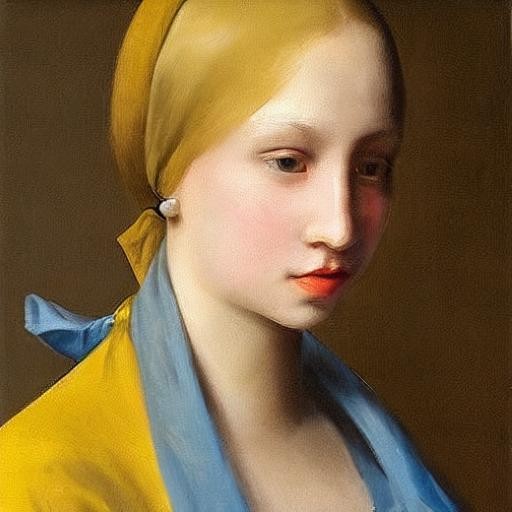}  &
    \includegraphics[width=\linewidth]{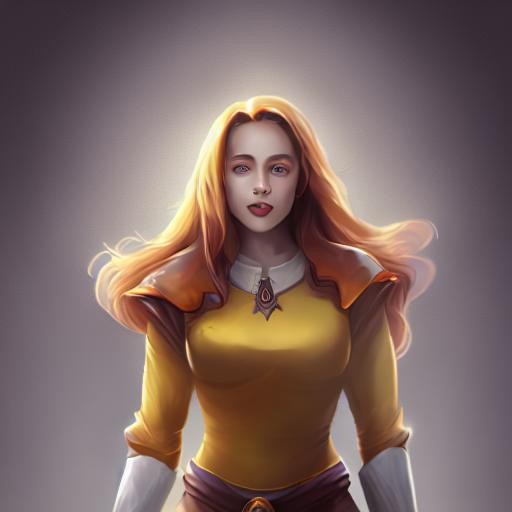}  &
    \includegraphics[width=\linewidth]{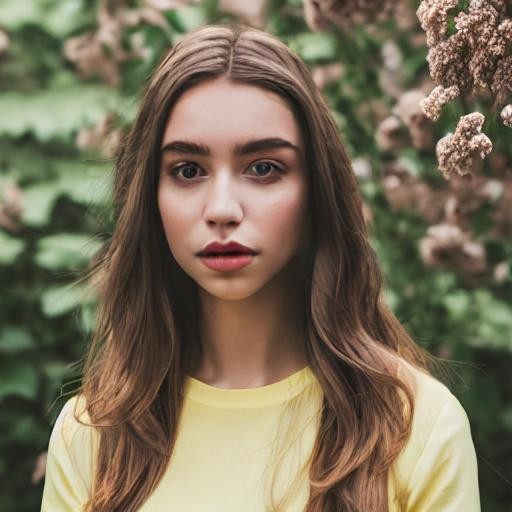}  &
    \includegraphics[width=\linewidth]{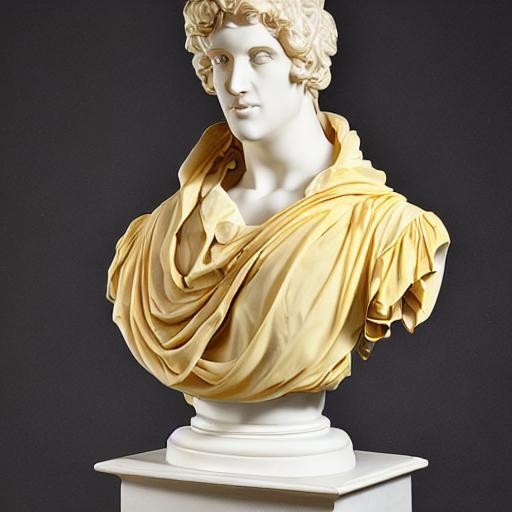}  \\

    \multicolumn{4}{c}{"laughing"} \\
    \includegraphics[width=\linewidth]{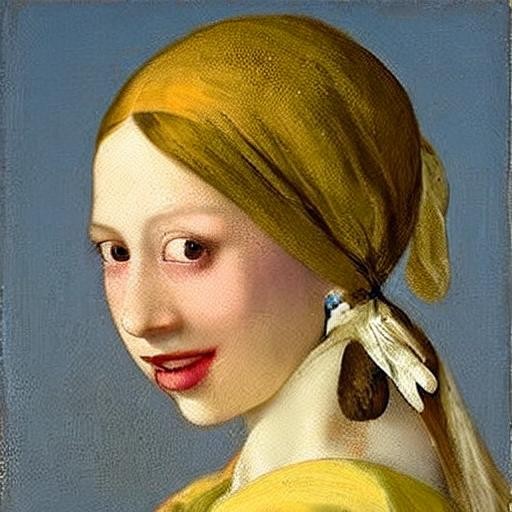}  &
    \includegraphics[width=\linewidth]{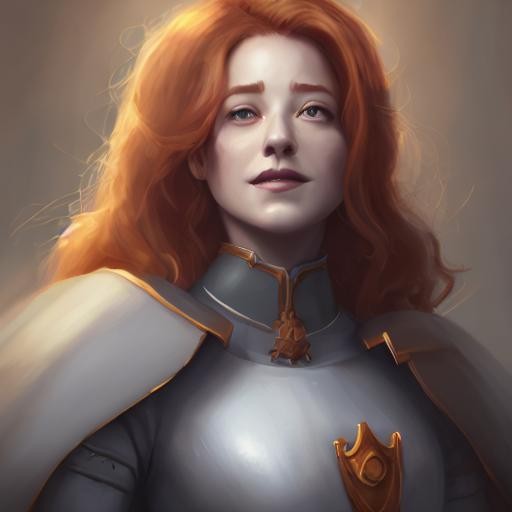}  &
    \includegraphics[width=\linewidth]{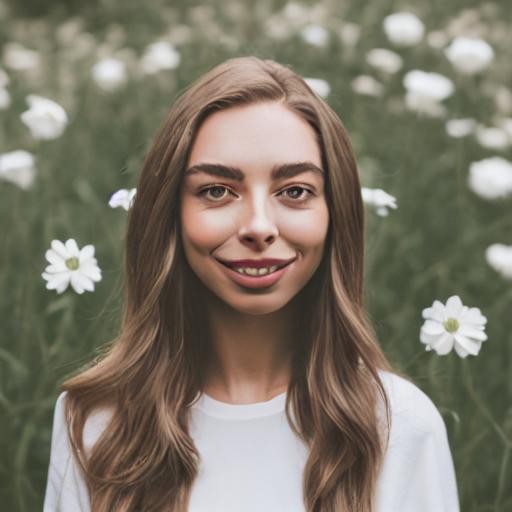}  &
    \includegraphics[width=\linewidth]{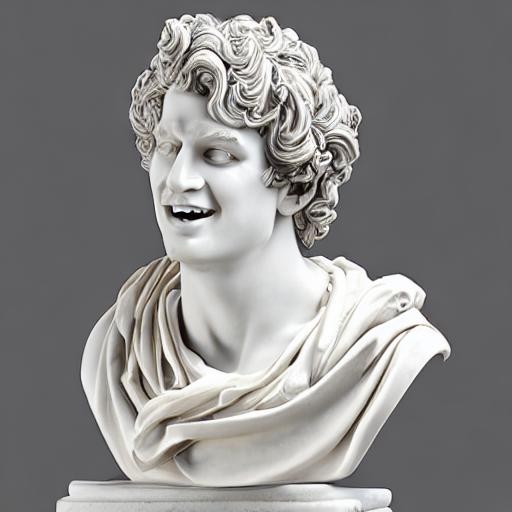}  \\

    \multicolumn{4}{c}{"wearing a hat"} \\
    \includegraphics[width=\linewidth]{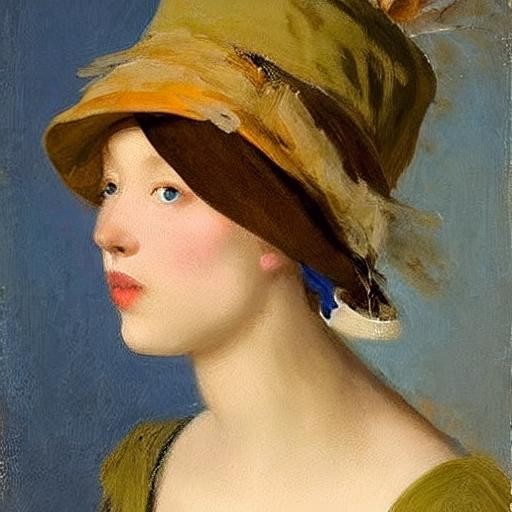}  &
    \includegraphics[width=\linewidth]{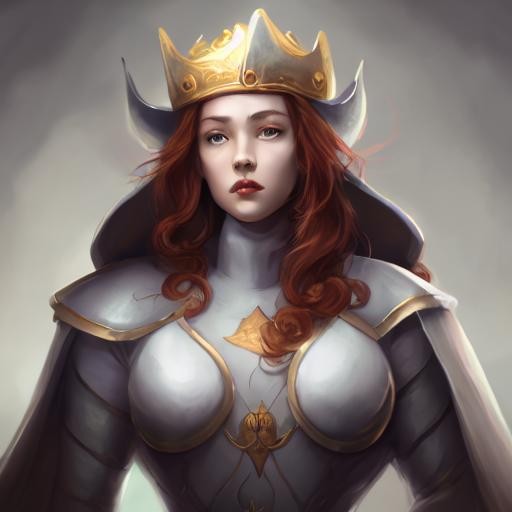}  &
    \includegraphics[width=\linewidth]{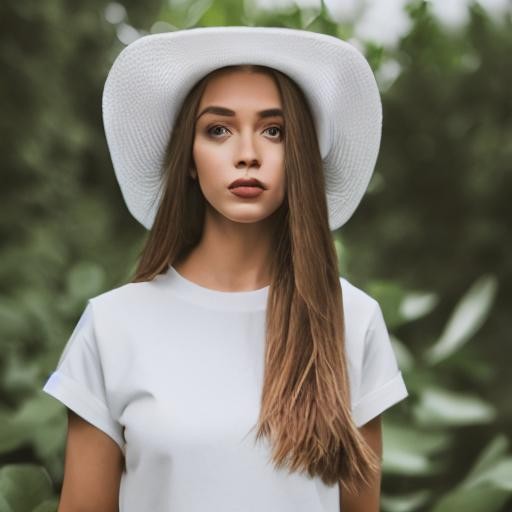}  &
    \includegraphics[width=\linewidth]{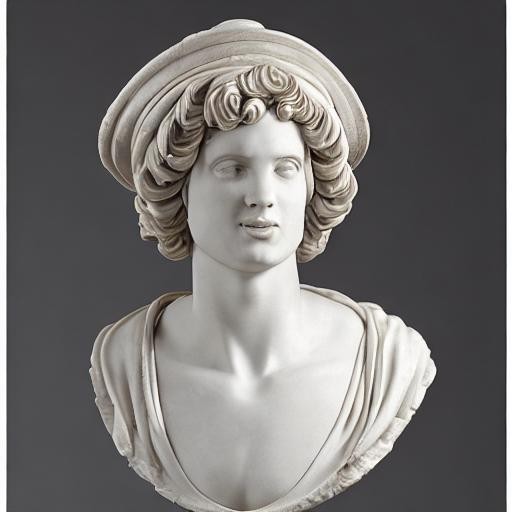}  \\

    \multicolumn{4}{c}{"wearing glasses"} \\
    \includegraphics[width=\linewidth]{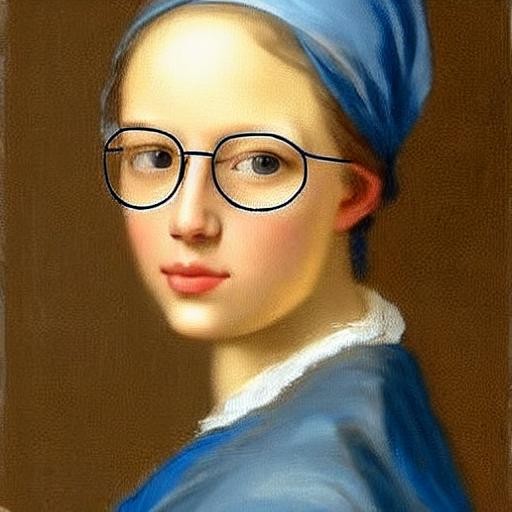}  &
    \includegraphics[width=\linewidth]{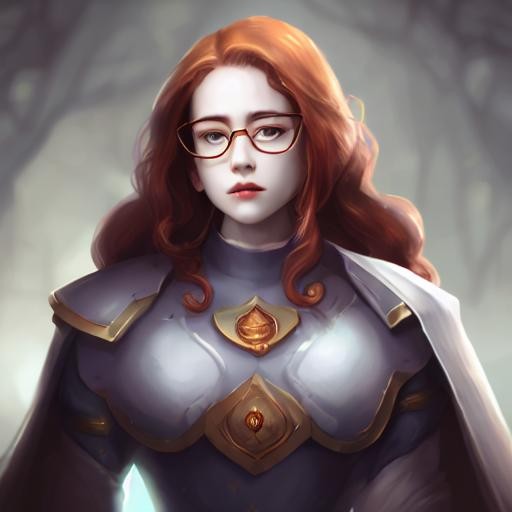}  &
    \includegraphics[width=\linewidth]{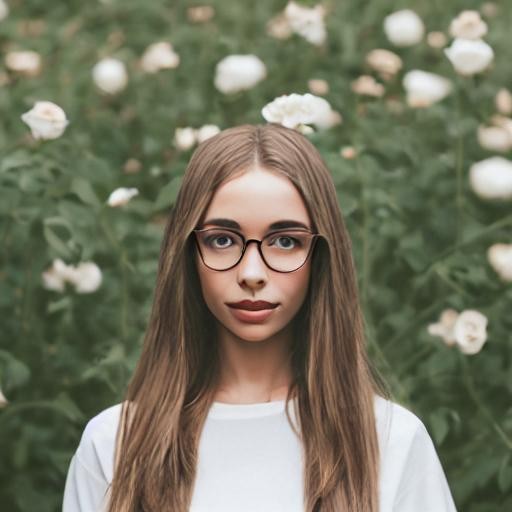}  &
    \includegraphics[width=\linewidth]{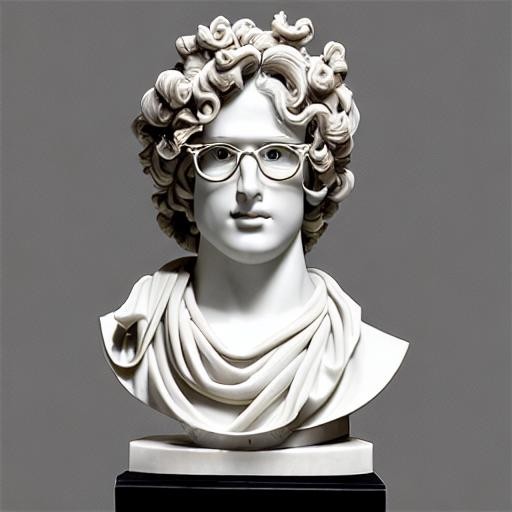}  \\

    \multicolumn{4}{c}{"in anime style"} \\
    \includegraphics[width=\linewidth]{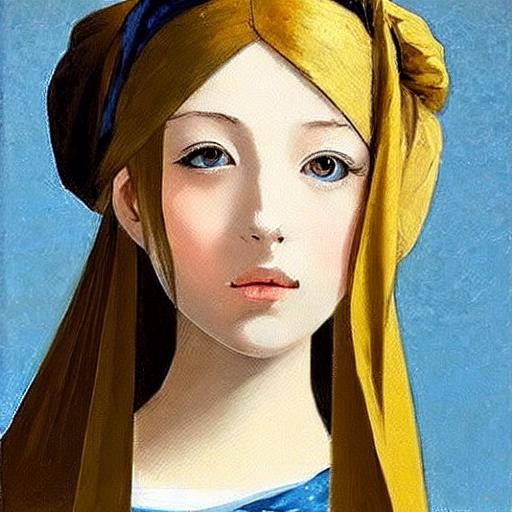}  &
    \includegraphics[width=\linewidth]{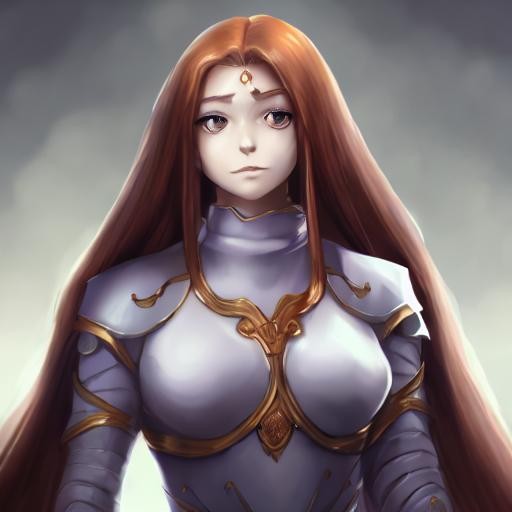}  &
    \includegraphics[width=\linewidth]{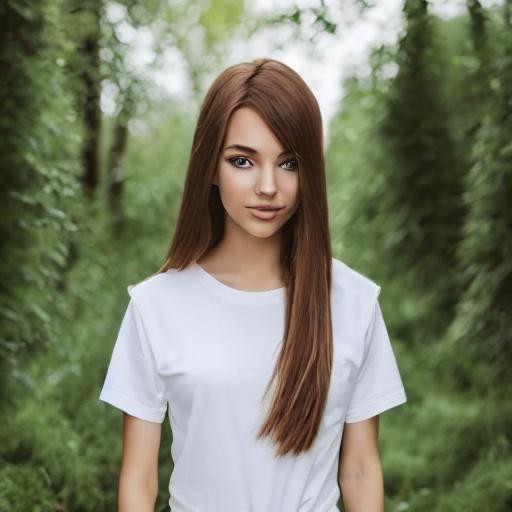}  &
    \includegraphics[width=\linewidth]{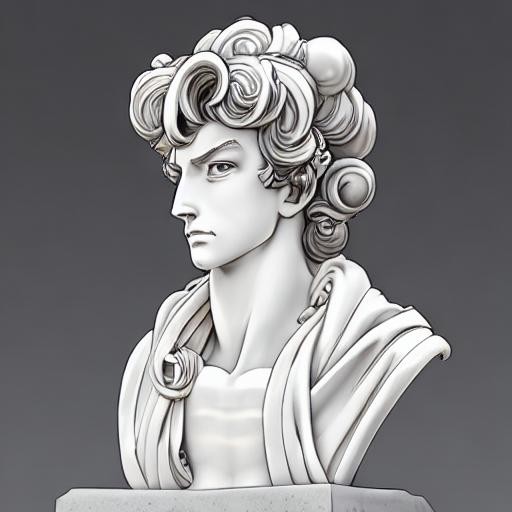}  \\

\end{tabular}

\captionof{figure}{
Unlike methods such as SeeCoder \cite{xu2023prompt}, \methodname{} keeps the original conditioning modality in place. We show that using \methodname{} on cross-attention layer can fuse image and text conditioning information logically and consistently according to the semantics of the image showing the advantage of directly adapting these layers.
}

\label{fig:prompt}
\end{table*}

\end{document}